\documentclass[a4paper,10pt,twocolumn]{article}   % A4, 10pt body text, two columns
\usepackage[margin=2cm]{geometry}                 % 2 cm margins all round

\usepackage[T1]{fontenc}       % 8-bit font encoding; only matters under pdfLaTeX, harmless otherwise
\usepackage{CharisSIL}         % Charis SIL serif for the text (in TeX Live)
\usepackage{newtxmath}         % Times-like maths to sit with Charis
\usepackage[british]{babel}    % British hyphenation and date formats
\usepackage{csquotes}          % quotation marks that follow babel; biblatex uses it for titles
\ifdefined\LaTeXML
  \DeclareUnicodeCharacter{0219}{\c{s}}
  \DeclareUnicodeCharacter{021B}{\c{t}}
  \DeclareUnicodeCharacter{0218}{\c{S}}
  \DeclareUnicodeCharacter{021A}{\c{T}}
\fi

\usepackage{xcolor}                              % named colours
\definecolor{lightblue}{RGB}{33,150,209}         % link colour used below

\usepackage{titlesec}                                        % control heading fonts and spacing
\newcommand{\headsizeI}{\fontsize{10}{12}\selectfont}        % level-1 heading: body size
\newcommand{\headsizeII}{\fontsize{9.5}{11.5}\selectfont}    % levels 2 and 3: half a point smaller
\titleformat{\section}{\normalfont\headsizeI\bfseries}{\thesection.}{0.5em}{}              % 1. Bold
\titleformat{\subsection}{\normalfont\headsizeII\itshape}{\thesubsection.}{0.5em}{}        % 1.1. Italic
\titleformat{\subsubsection}{\normalfont\headsizeII\itshape\bfseries}{\thesubsubsection.}{0.5em}{}  % 1.1.1. Bold italic
\titlespacing{\section}{0pt}{12pt plus 2pt minus 2pt}{5pt plus 1pt minus 1pt}        % space above/below; unstarred form keeps the paragraph indent
\titlespacing{\subsection}{0pt}{11pt plus 2pt minus 2pt}{4pt plus 1pt minus 1pt}
\titlespacing{\subsubsection}{0pt}{11pt plus 2pt minus 2pt}{4pt plus 1pt minus 1pt}

\usepackage{authblk}                                             % author and affiliation block
\usepackage{caption}                    % caption formatting
\usepackage{etoolbox}                                  % hooks used here and in the bibliography setup
\AtBeginEnvironment{tabular}{\footnotesize}            % table bodies in footnote size
\AtBeginEnvironment{tabularx}{\footnotesize}

\usepackage{graphicx}                  % \includegraphics and \resizebox
\graphicspath{{figures/}}              % look for figures here first
\usepackage{array}                     % >{...} column specifications
\usepackage{tabularx}                  % tables that fill a set width
\usepackage{booktabs}                  % \toprule, \midrule, \bottomrule
\usepackage{multirow}                  % cells spanning several rows
\usepackage{threeparttable}            % table notes under a table
\usepackage{longtable}                 % supplement: tables spanning pages
\usepackage{float}                     % supplement: [H] placement
\usepackage{pdflscape}                 % supplement: landscape pages
\newcolumntype{L}[1]{>{\raggedright\arraybackslash}p{#1}}  % supplement: wrapped left column
\usepackage{dblfloatfix}               % lets figure*/table* sit at the bottom of a two-column page

\usepackage{xurl}                      % allow URLs to break anywhere
\usepackage[colorlinks=true, linkcolor=lightblue,     % coloured link text rather than boxes
            citecolor=lightblue, urlcolor=lightblue]{hyperref}
\newcommand{\supref}[1]{\ref{#1}}     % supplementary reference: prints the number, no dead link

\usepackage[
  backend=biber,          % biber processes the .bib
  style=authoryear,       % (Author Year) citations, author-year list
  sorting=nyt,            % list sorted by name, year, title
  sortcites=true,         % citation groups sorted the same way
  language=british,       % British strings and dates in the list
  maxcitenames=2,         % 1 author: name; 2: both; 3+: first et al.
  mincitenames=1,
  maxbibnames=99,         % never truncate author lists in the reference list
  minbibnames=99,
  giveninits=true,        % initials only: Brown, C.F.
  uniquelist=false,       % do not expand "et al." to tell authors apart
  uniquename=false,
  useprefix=true,         % Van der Geer, J. (not Geer, J. van der)
  date=year,              % year only in citations and list
  urldate=long,           % "visited on 18 February 2026"
  dateabbrev=false,       % full month names
  doi=true,               % print DOIs
  url=true,               % print URLs (cleared below when a DOI exists)
  isbn=false,             % no ISBNs
  eprint=false            % no arXiv identifiers
]{biblatex}
\DeclareSourcemap{\maps[datatype=bibtex]{         % use Zotero's Journal Abbr (BBT: shortjournal)
  \map[overwrite=true]{                           % in place of the full journal title
    \step[fieldsource=shortjournal]
    \step[fieldset=journaltitle, origfieldval]
  }
}}
\DeclareSourcemap{\maps[datatype=bibtex]{         % iDiv dataset: show the repository as organisation,
  \map{                                           % not as publisher
    \step[fieldsource=url, match=\regexp{idata\.idiv\.de}, final]
    \step[fieldsource=publisher, fieldset=organization, origfieldval]
    \step[fieldset=publisher, null]
  }
}}
\AtEveryBibitem{%
  \iffieldundef{url}{}{\clearfield{howpublished}}%       % drop howpublished when a URL is present
  \iffieldundef{doi}{}{\clearfield{url}%                 % when a DOI exists, drop the URL and visited date
    \clearfield{urlyear}\clearfield{urlmonth}\clearfield{urlday}}%
}
\appto{\bibsetup}{\raggedright}                   % ragged-right reference list (no stretched lines)

\usepackage{fancyhdr}                                            % custom headers and footers
\renewcommand{\subsectionmark}[1]{}                              % subsections do not change it

\begin{document}

\title{Geospatial embeddings detect old-growth forests but buffered spatial validation narrows their advantage over Sentinel features}

\author[a,*]{Thomas Ratsakatika}
\author[b]{Mihai Zotta}
\author[c]{Srinivasan Keshav}
\author[a]{Emily R. Lines}

\affil[a]{Department of Geography, University of Cambridge, Downing Place, Cambridge, \mbox{CB2 3EN}, United Kingdom}
\affil[b]{Fundația Conservation Carpathia, Calea Feldioarei nr. 27A, Brașov 500450, Romania}
\affil[c]{Department of Computer Science and Technology, University of Cambridge, 15 JJ Thomson Avenue, Cambridge, \mbox{CB3 0FD}, United Kingdom}
\affil[*]{Corresponding author. Email: trr26@cam.ac.uk}

\date{23 September 2026}

\twocolumn[%
  \begin{@twocolumnfalse}
    \maketitle
    \begin{abstract}
        Old-growth forests develop over centuries under minimal anthropogenic disturbance, producing structurally complex and biodiverse stands. In Europe, protecting them requires mapping that is accurate for individual forest parcels yet deployable continent-wide. Geospatial foundation model (GFM) embeddings enable label-scarce land classification, but their value for old-growth detection remains unknown. Here, we map old-growth forests across 211,893\,ha of Romania's Southern Carpathians, a beech--spruce landscape typical of the Alpine Biogeographic Region. We construct high-confidence, expert-informed reference labels for old-growth and non-old-growth parcels. We add AlphaEarth, TESSERA v2 and \mbox{Sentinel-1/2} features to a common baseline of topographic and human-access predictors, then compare them under spatially blocked validation with and without \mbox{10\,km} train-test buffers to limit residual autocorrelation. With buffering, GFM and \mbox{Sentinel-1/2} predictors increase precision-recall AUC by \mbox{0.21--0.25 [95\% CIs: 0.15--0.34]} relative to baseline, indicating spectral data contain a spatially robust old-growth signal. With a PR-AUC of \mbox{0.84 [0.79--0.88]}, TESSERA outperforms \mbox{Sentinel-1/2} (\mbox{$+$0.08 [$+$0.05 to $+$0.11]}) and AlphaEarth (\mbox{$+$0.08 [$+$0.04 to $+$0.12]}) under unbuffered spatial validation. At a \mbox{10\,km} buffer, however, this advantage narrows to \mbox{$+$0.04 [$-$0.01 to $+$0.11]} and \mbox{$+$0.03 [$-$0.04 to $+$0.10]}, intervals consistent with no difference. At \mbox{10\,m} resolution, convolutional neural networks add no benefit over pixel-based XGBoost. Comparisons with four national- and continental-scale products show the importance of non-old-growth labels, and reveal 81\% agreement between our predictions and a field-calibrated map. We conclude that buffered spatial validation is vital when transferring old-growth detection models to unseen landscapes, and provide our labels and predictions for future work.
    \end{abstract}
    \vspace{0.5\baselineskip}
    \vspace{0.5\baselineskip}
    
    \bigskip
  \end{@twocolumnfalse}
]

\section{Introduction}

Temperate old-growth forests are valuable ecosystems characterised by large old trees, structural complexity and high biodiversity \parencite{spies_ecological_2004, wirth_old-growth_2009}. They provide critical ecosystem services, from carbon storage \parencite{keith_re-evaluation_2009} and sequestration \parencite{luyssaert_old-growth_2008} to microclimate buffering \parencite{frey_spatial_2016} and water-flow regulation \parencite{jones_forest_2022}. While `primary' forests must have no clearly visible indications of human activity \parencite{fao_global_2020}, the term `old-growth' encompasses forests that show minimal signs of past anthropogenic disturbance, provided their natural structures and processes are not significantly altered \parencite{european_commission_commission_2023}. This makes old-growth forests of considerable interest for European forest conservation, and the EU Biodiversity Strategy for 2030 prioritises old-growth alongside primary forests for mapping, monitoring and protection \parencite{european_commission_eu_2021}. However, these forests are small and fragmented \parencite{barredo_mapping_2021}, and detecting them requires a method that is accurate for individual parcels yet deployable continent-wide.

Remote sensing approaches for mapping old-growth forests generally fall into two groups (adapted from \cite{hirschmugl_review_2023}). The first uses interpretable forest metrics, such as tree height and structural complexity, typically from airborne laser scanning (ALS) data. Old-growth conditions are then identified using expert-defined thresholds (e.g. \cite{trouve_identifying_2024}) or supervised models trained on these metrics alongside Earth observation (EO)-derived features (e.g. \cite{lalechere_assessing_2024, adiningrat_mapping_2024, hirschmugl_classification_2026}). The second approach uses supervised models to detect old-growth forests directly from EO data, without extracting interpretable forest metrics (e.g. \cite{spracklen_identifying_2019, munteanu_using_2022}). Both approaches often incorporate non-EO geospatial predictors of old-growth forest such as topography and transport access \parencite{mikolas_primary_2019}, and these frequently rank as important features \parencite{bubnicki_conservation_2024, lalechere_assessing_2024}. Input data are predominantly organised at the pixel, plot or parcel level, with features passed to a model such as a decision tree. While some studies use texture features (e.g. \cite{hirschmugl_classification_2026, lalechere_assessing_2024, spracklen_identifying_2019}), models that use spatial context, such as convolutional neural networks (CNNs), are not routinely used for old-growth detection (\cite{hirschmugl_review_2023}; Web of Science search, see Section~\supref{sec:sup_lit_search} of the supplementary material), despite horizontal structural complexity being indicative of old-growth conditions \parencite{spies_ecological_2004}.

Existing studies differ in their prediction objectives and spatial scales, making intercomparison challenging. Prediction objectives include binary old-growth/non-old-growth forest classification (e.g. \cite{adiningrat_mapping_2024,adiningrat_investigating_2025, spracklen_identifying_2019}), categorical classification of succession or developmental stages (e.g. \cite{falkowski_characterizing_2009, martin_old_2022}), as well as continuous indices of forest maturity, primary forest likelihood, and forest structural complexity (e.g. \cite{fuhr_detecting_2022, lalechere_assessing_2024, sabatini_spatially-explicit_2020, munteanu_using_2022, bubnicki_conservation_2024}). Spatial scales vary from individual forest plots (e.g. \cite{adiningrat_investigating_2025, fuhr_detecting_2022, lalechere_assessing_2024}) to national (e.g. \cite{munteanu_using_2022, bubnicki_conservation_2024}) and continental (\cite{sabatini_spatially-explicit_2020}, although this study uses non-EO geospatial predictors only).

Many studies rely on ALS-derived features as a structural signature of old-growth forests \parencite{hirschmugl_review_2023}. Although ALS features are effective predictors \parencite{adiningrat_mapping_2024, lalechere_assessing_2024}, many countries lack national surveys and obtaining ALS data can be resource-intensive, limiting widespread adoption. In contrast, free and open satellite data, including from the Sentinel and Landsat missions \parencite{torres_gmes_2012,drusch_sentinel-2_2012,wulder_current_2019}, offer global coverage and support monitoring and mapping with lower ongoing costs. Geospatial foundation model (GFM) embeddings offer an alternative to hand-crafted EO features. GFMs are trained on large multi-sensor, multi-temporal remote sensing datasets \parencite{xiao_foundation_2025} and models such as AlphaEarth \parencite{brown_alphaearth_2025} and TESSERA \parencite{feng_tessera_2026, feng_tessera_v2_2026} provide analysis-ready embeddings designed for label-scarce land classification. However, the value of GFM embeddings over established approaches for old-growth detection is not yet known.

Like many environmental features, old-growth forests are highly spatially clustered \parencite{mikolas_primary_2019}, and their environmental predictors follow smooth gradients \parencite{legendre_spatial_1993}. Spatial autocorrelation between the data on which a model is fitted and evaluated violates the independence assumption of most statistical methods, inflating reported performance and weakening the transfer of conclusions beyond the study area \parencite{legendre_spatial_1993, ploton_spatial_2020, kattenborn_spatially_2022}. However, few existing old-growth mapping studies explicitly account for spatial autocorrelation, a gap that extends across GFM research more broadly \parencite{zhou_measuring_2026}.

\begin{figure*}
    \centering
    \includegraphics[width=1\linewidth]{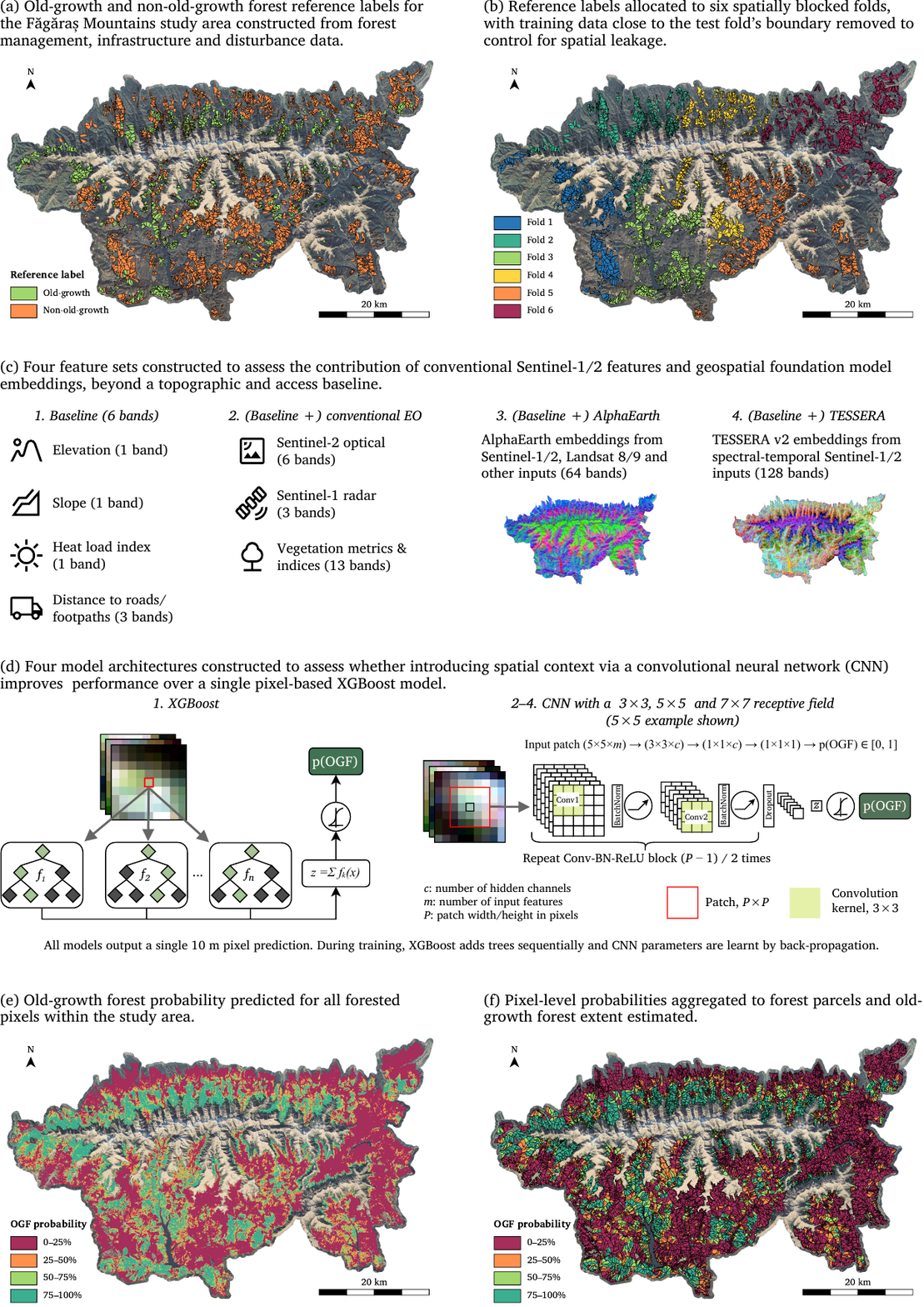}
    \caption{Graphical overview of the materials and methods: (a) reference label construction; (b) spatially blocked fold allocation; (c) feature set composition (icons from Google Material Symbols, Apache License 2.0); (d) pixel-based XGBoost and patch-based convolutional neural network (CNN) architectures; (e) pixel-level predictions across the study area; (f) aggregation of pixel-level predictions to forest parcels. Făgăraș Mountains area of interest bounding box centred on 24.742° E, 45.528° N. Aerial photograph \textcopyright{} Fundația Conservation Carpathia, reproduced with permission. BN, batch normalisation; Conv, convolution; OGF, old-growth forest; ReLU, rectified linear unit.}
    \label{fig:methods}
\end{figure*}

Here, we detect old-growth forests using only freely available EO data, with spatial validation and buffering based on measured autocorrelation (Fig.~\ref{fig:methods}). We use detailed forest parcel data to construct high-confidence reference labels for both old-growth and non-old-growth forests in our study area. We compare the performance of Sentinel-1/2-derived features (hereafter, conventional EO) with AlphaEarth and TESSERA v2 embeddings, controlling for topographic and access (e.g. distance to road) covariates. To our knowledge, this is the first evaluation of GFM embeddings for old-growth forest detection (Web of Science search, Section~\supref{sec:sup_lit_search} of the supplementary material).

Our study area is the Făgăraș Mountains in Romania, one of the largest primary and old-growth forested landscapes in temperate Europe \parencite{kameniar_synchronised_2023}. The forests are dominated by European beech (\textit{Fagus sylvatica}) and Norway spruce (\textit{Picea abies}), with some intermixed silver fir (\textit{Abies alba}) \parencite{kameniar_synchronised_2023}, typical of the wider Alpine Biogeographic Region \parencite{sundseth_natura_2009}. The landscape's rugged topography and mixed land use make it a challenging and representative test case for old-growth detection in Europe \parencite{sabatini_where_2018}. Four existing national and continental studies map old-growth or similar forests in the study area, two derived from machine learning \parencite{sabatini_spatially-explicit_2020, munteanu_using_2022} and two from expert inspection and thresholding \parencite{kathmann_potential_2017, schickhofer_inventory_2019}; however, their estimates of old-growth forest extent vary threefold from 20,865\,ha to 64,146\,ha (Section~\ref{sec:existing_studies}). In this paper, we (1) evaluate the agreement of these maps with our locally verified labels, motivating a revised modelling approach; (2) investigate to what extent EO data, including GFM embeddings, can detect old-growth forests beyond topographic and access covariates; and (3) test whether introducing spatial context through a CNN model architecture improves performance. Finally, we map old-growth forests across the entire study area and provide vector and raster layers to support future research.

\begin{table*}[b]
    \centering
    \caption{Vector and raster datasets used for old-growth forest and non-old-growth forest reference label construction in the Făgăraș Mountains area of interest (AOI).}
    \label{tab:datasets_groundtruth}
    \begin{tabularx}{\textwidth}{@{}c >{\raggedright\arraybackslash}p{3.5cm} X@{}}
        \toprule
         & \textbf{Dataset} & \textbf{Description} \\
        \midrule
        (a) & Natura 2000 boundaries
            & Vector of Natura 2000 protected area boundaries as of end-2024. Source: \textcite{eea_natura_2025}. \\[3pt]
 
        (b) & Forest parcels map
            & Vector of 15,515 (non-overlapping, post-cleaning) forest parcel boundaries within the AOI as of 2018. Source: Fundația Conservation Carpathia. \\[3pt]
 
        (c) & Forest management plans
            & Vector layers of forest parcel management attributes, for seven planning periods between 2013 and 2025. Source: Fundația Conservation Carpathia. \\[3pt]
 
        (d) & Virgin and quasi-virgin forests
            & Vector layer of virgin and quasi-virgin forests expanding on Romania's national catalogue \parencite{mmediu_catalogul_2020} and verified by Fundația Conservation Carpathia (FCC) forestry experts. Source: Fundația Conservation Carpathia. \\[3pt]
 
        (e) & Roads and footpaths
            & Vector layer of paved roads, unpaved roads and footpaths. Source: \textcite[extracted on 28 May 2026]{openstreetmap_contributors_openstreetmap_2026}. \\[3pt]
 
        (f) & Forest disturbances
            & Forest disturbance 30\,m raster layer from the European Forest Disturbance Atlas 1985--2023 (1985--2020 bands used). Source: \textcite{viana-soto_european_2025}. \\
 
        \bottomrule
    \end{tabularx}
\end{table*}

\section{Materials and methods}

\subsection{Study area}

The area of interest (AOI) is the Făgăraș Mountains in Romania's Southern Carpathians (Fig.~\ref{fig:methods}), operationally delineated by the union of the \textit{Munții Făgăraș} and \textit{Râul Târgului--Argeșel--Râușor} Natura~2000 sites \parencite{eea_natura_2025}. Elevations range from 343\,m to 2,523\,m above sea level \parencite[own analysis of][]{hawker_30_2022}, peaking at \textit{Vârful Moldoveanu}, the highest mountain in Romania. Forests cover 152,366\,ha (71.9\%) of the 211,893\,ha AOI, comprising broadleaf (14.2\% of the AOI, median elevation 921\,m), mixed (30.9\%, 1,136\,m) and coniferous (26.8\%, 1,498\,m) \parencite[own analysis of][]{clms_corine_2020, hawker_30_2022}. The landscape follows a vegetation gradient typical of the Southern Carpathians, with broadleaf forests dominated by European beech (\textit{Fagus sylvatica}) at lower elevations, transitioning at higher elevations to coniferous forests, dominated by Norway spruce (\textit{Picea abies}) and, to a lesser extent, silver fir (\textit{Abies alba}) \parencite{kameniar_synchronised_2023}. Heathland and natural grassland account for a further 22.9\% of the AOI and occur predominantly above 1,900\,m. The ecosystem hosts high levels of biodiversity, with preliminary surveys identifying over 2,000 animal, plant, fungal and lichen species \parencite{linnell_biodiversity_2016}, including a dense community of large carnivores such as brown bears (\textit{Ursus arctos}), Eurasian lynx (\textit{Lynx lynx}) and grey wolves (\textit{Canis lupus}) \parencite{dyck_draculas_2022,iosif_eurasian_2022}.

\begin{table*}[t]
    \centering
    \caption{Input data for the four feature sets (excluding the coordinate-only control) used during model training: \textit{baseline} = a--d (6 bands); \textit{baseline + conventional Earth observation (EO)} = a--f (28 bands); \textit{baseline + AlphaEarth} = a--d,\,g (70 bands); \textit{baseline + TESSERA} = a--d,\,h (134 bands). All data were resampled onto the 10\,m EPSG:3035 grid.}
    \label{tab:datasets_predictors}
    \begin{tabularx}{\textwidth}{@{}c >{\raggedright\arraybackslash}p{2.5cm} c X@{}}
        \toprule
         & \textbf{Dataset} & \textbf{Bands} & \textbf{Description} \\
        \midrule
        \multicolumn{4}{@{}l}{\textit{Baseline predictors (topography and access)}} \\[3pt]
        (a) & Elevation (m) & 1
            & Forest and Buildings Removed Copernicus Digital Elevation
              Model (FABDEM), at 1 arc second (${\sim}30$\,m). Source: \textcite{hawker_30_2022}. \\[3pt]

        (b) & Slope (degrees) & 1
            & Slope at 1 arc second (${\sim}30$\,m), computed from (a) using the Horn method \parencite{horn_hill_1981}. \\[3pt]

        (c) & Heat load index & 1
            & Unitless topographic index of potential heat load at 1 arc second (${\sim}30$\,m), calculated using \textcite[Eq. 1]{mccune_equations_2002}, with values lowest on north-east-facing slopes. Input variables: slope (b), aspect from (a) using the Horn method
              \parencite{horn_hill_1981} and latitude (cell midpoint). \\[3pt]

        (d) & Distance to roads and footpaths (m) & 3
            & Euclidean distance to paved roads, unpaved roads and footpaths, derived from roads and
              footpaths vectors (e) in Table~\ref{tab:datasets_groundtruth}, calculated directly on the 10\,m target grid. \\[3pt]
        \midrule
        \multicolumn{4}{@{}l}{\textit{Conventional Earth observation predictors}} \\[3pt]

        (e) & ESA WorldCover Sentinel-1/2 composites & 12
            & European Space Agency (ESA) WorldCover 2020 annual composites, comprising Sentinel-2 median red, green, blue (RGB) and near-infrared (NIR), normalised difference vegetation index (NDVI) percentile (p90, p50, p10), and Sentinel-1 median synthetic aperture radar (SAR) VV, VH and VH/VV ratio at 0.3 arc seconds (${\sim}10$\,m); and Sentinel-2 median short-wave infrared (SWIR, bands 11/12) at 0.6 arc seconds (${\sim}20$\,m). Source: \textcite{zanaga_esa_2021}. \\[3pt]

        (f) & CLMS Vegetation Phenology and Productivity & 10
            & Selected Vegetation Phenology and Productivity (VPP) time-series metrics for 2020 (season 1) at 10\,m, provided by the Copernicus Land Monitoring Service (CLMS), capturing start- and end-of-season (SOS/EOS) dates and values, minimum/maximum seasonal values, seasonal amplitude and productivity, and green-up and green-down slopes. Source: \textcite{clms_high_2020}. \\[3pt]
        \midrule
        \multicolumn{4}{@{}l}{\textit{Geospatial foundation model predictors}} \\[3pt]

        (g) & AlphaEarth embeddings & 64
            & Annual 10\,m embeddings for 2020, derived from spectral-temporal Sentinel-1/2 and Landsat 8/9 inputs, with additional multi-source data used during training, including topography and land cover. Source: \textcite{brown_alphaearth_2025}. \\[3pt]

        (h) & TESSERA v2 embeddings & 128
            & Annual 10\,m embeddings for 2020, derived from spectral-temporal Sentinel-1 and Sentinel-2 data. Source: \textcite{feng_tessera_v2_2026}. \\
        \bottomrule
    \end{tabularx}
\end{table*}

\subsection{Reference label construction}

We constructed reference labels of old-growth forest (OGF) and non-old-growth forest (non-OGF) parcels within the AOI by applying a rule-based data fusion procedure to forest management, infrastructure and disturbance datasets (Table~\ref{tab:datasets_groundtruth}). Unlike presence-only approaches that rely on pseudo-absences or background points, we constructed explicit non-OGF labels to reduce label noise and enable false-positive estimation, addressing a key limitation identified by \textcite{bubnicki_conservation_2024}. A map of 15,515 forest parcels covering 92.1\% of the forested area within the AOI served as a base geometry (Table~\ref{tab:datasets_groundtruth}b), and the parcel attributes were augmented with data from seven forest management plan vector layers dated from 2013 to 2025 (Table~\ref{tab:datasets_groundtruth}c). Where conflicts occurred, the forest management plan for the year closest to 2020, our reference year, was prioritised. The resulting parcel map contained average stand-age attributes for 9,402 parcels (61\%).

Conservative filters were applied to the parcels to assign high-confidence reference labels. We labelled parcels as OGF if at least 90\% of their area overlapped with a vector layer of `virgin' or `quasi-virgin' forests confirmed by forestry experts at Fundația Conservation Carpathia (FCC), a conservation organisation operating in the AOI (Table~\ref{tab:datasets_groundtruth}d). Under Romanian law, virgin and quasi-virgin forests are defined more strictly than old-growth \parencite{mmediu_ghid_2020} and therefore serve as a conservative reference for the OGF class. Overlapping parcels with a recorded stand age below 80 years or whose attribute comments included disturbance flags were excluded. Roads and footpaths (with a 10\,m buffer) and European Forest Disturbance Atlas pixels (1985--2020) were clipped from the OGF parcel geometries to reduce label noise (Table~\ref{tab:datasets_groundtruth}e--f).

We labelled the remaining parcels with a recorded average stand age of 80 years or less as non-OGF. Under Romanian silvicultural regulations, beech stands below this age are typically eligible for active thinning \parencite{bouriaud_effects_2019}, and evidence from protected Romanian mountain forests indicates that old-growth characteristics do not fully recover within 80 years of the last harvest \parencite{barbu_variation_2023}. The 80-year threshold was used to identify high-confidence negatives, not to define old-growth forests. Therefore, forest parcels older than 80 years received no age-defined label and could be classified by the model as either OGF or non-OGF. Roads, footpaths and disturbances, each independently identifiable from open datasets (Table~\ref{tab:datasets_groundtruth}), were retained as non-OGF rather than clipped.

The resulting reference labels comprised 1,016 OGF and 3,837 non-OGF parcels, covering 12,775\,ha (8.4\%) and 34,574\,ha (22.7\%) of the forested AOI, respectively (Fig.~\ref{fig:methods}a). Compared with non-OGF parcels, OGF parcels were typically located at higher altitudes, on steeper slopes and further from roads (Fig.~\supref{fig:reference_label_statistics}). This distribution held for coniferous, broadleaf and mixed parcels, and aligned with expected patterns \parencite{mikolas_primary_2019}. Further details on the reference label construction are provided in Section~\supref{sec:supp_ref_labels} of the supplementary material.

\subsection{Performance metrics}\label{sec:methods_metrics}

The study used a common 10\,m pixel grid in the coordinate reference system EPSG:3035. Forest parcels, not pixels, are the units of management within the AOI and are more relevant for evaluating a model's usefulness. We therefore calculated parcel-level metrics by comparing each reference parcel's label (OGF = 1, non-OGF = 0) with the unweighted mean of the predictions (0, 1 or 0--1) for all pixels fully enclosed by that parcel. This process excluded 28 parcels in which no pixel was fully enclosed. Parcel-level results were reported for the remaining 4,825 parcels.

We reported multiple performance metrics as no single measure is universally suited for all evaluation objectives \parencite{stehman_selecting_1997}. Our primary metric was the precision-recall area under the curve (PR-AUC), complemented by the area under the receiver operating characteristic curve (ROC-AUC), the overlap coefficient, precision, recall and F1. PR-AUC, calculated as the average precision, summarises the trade-off between precision (the proportion of OGF predictions that were true OGF parcels) and recall (the proportion of true OGF parcels predicted as OGF). It ranges from the OGF prevalence rate (0.21) for a no-skill classifier to 1 for a perfect classifier. We selected PR-AUC as our primary metric because it is threshold-free, not inflated by the non-OGF majority class, and sensitive to the performance of the minority OGF class.

ROC-AUC quantifies the probability that a randomly selected OGF parcel scores higher than a randomly selected non-OGF parcel and ranges from 0.5 for a no-skill classifier to 1 for a perfect classifier. Its advantage is that it is independent of prevalence, improving comparability across studies; however, it can appear optimistic under class imbalance \parencite{saito_precision-recall_2015}, such as that present in our study.

The overlap coefficient quantifies the area shared by two normalised histograms of the predicted scores for OGF and non-OGF reference parcels, ranging from 0 for perfectly separated classes to 1 for identical class distributions \parencite{pastore_measuring_2019}. It is threshold-free and prevalence-insensitive, but does not capture ranking performance.

Precision, recall and F1 (the harmonic mean of precision and recall) were calculated at a 0.5 majority-area threshold for the existing studies (Fig.~\supref{fig:threshold_sensitivity}). For our models, each parcel was classified as OGF or non-OGF using the threshold that maximised F1 on its own fold's pooled inner-fold (see Section~\ref{sec:methods_spatial_cv}) predictions; precision, recall and F1 were then calculated from the pooled held-out binary predictions. This ensured that the threshold used to assess each parcel was never fitted to that same parcel.

\subsection{Evaluation against existing studies}

We evaluated four existing studies against our reference labels: two derived from machine learning models (\cite{sabatini_spatially-explicit_2020,munteanu_using_2022}, hereafter `model-based' studies) and two derived from expert inspection and thresholding (\cite{kathmann_potential_2017,schickhofer_inventory_2019}, hereafter `rule-based' studies). They map different combinations of primary, old-growth and high-conservation-value forests; however, our OGF labels (virgin and quasi-virgin forests) and non-OGF labels (stands $\leq$80 years) represent parcels that these definitions should reasonably be expected to include and exclude, respectively. Nonetheless, each study was designed for a different objective, and our evaluation was limited to their parcel-level predictions within the AOI.

\textcite{sabatini_spatially-explicit_2020} provides a Europe-wide binary (0/1) 250\,m primary forest raster combining documented forests from the European Primary Forest Database (EPFD) v1.0 \parencite{sabatini_where_2018} with additional forests modelled using a boosted regression tree trained on geospatial proxies such as climate and topography, and thresholded to fill the shortfall between national estimates of undisturbed forests and the EPFD \parencite{sabatini_protection_2020}. \textcite{munteanu_using_2022} provide a 30\,m four-class high-conservation-value forest (HCVF) raster from maximum entropy models trained on optical, radar and geospatial proxy data, from which we assigned their two structurally complex classes, prime and at-risk HCVF, as OGF. \textcite{kathmann_potential_2017} provide a vector layer of primary and old-growth forest polygons for the Romanian Carpathians from vegetation index thresholding, visual interpretation and rule-based exclusions. \textcite{schickhofer_inventory_2019} provide a vector layer of potential primary and old-growth forest polygons for Romania from visual interpretation of very-high-resolution optical imagery against naturalness criteria calibrated with field observations; of the four, it is the most comprehensive expert inventory of primary and old-growth forests covering our AOI. The two model-based studies, \textcite{sabatini_spatially-explicit_2020, munteanu_using_2022}, also publish continuous probability surfaces (0--1) of primary and forest structural complexity, respectively.

We applied a common evaluation method across studies. The original rasters and rasterised vectors were reprojected onto the common reference grid, and nearest-neighbour resampling was applied to the raster outputs to preserve the authors' predictions. For both the binary and continuous outputs, we calculated per-study parcel-level OGF scores from the unweighted mean of the pixel-level predictions (0, 1 or 0--1) fully enclosed by that parcel (see Section~\ref{sec:methods_metrics}). We calculated PR-AUC, ROC-AUC, precision, recall and F1 for the binary outputs, and we additionally calculated the overlap coefficient and associated histograms for the continuous outputs. 

\subsection{Feature sets and model architectures}\label{sec:methods_predictors}

We constructed five feature sets to evaluate the extent to which freely available EO data contain information predictive of old-growth forests, beyond a topographic and access baseline (Fig.~\ref{fig:methods}c). Following \textcite{ploton_spatial_2020}, we defined a \textit{coordinate-only control} feature set, using the easting and northing of pixels as sole predictors. A \textit{baseline} feature set was constructed from six geospatial layers: elevation, slope, heat load index, and distance to paved roads, unpaved roads and footpaths (Table~\ref{tab:datasets_predictors}a--d). We then constructed three EO-based feature sets to evaluate their incremental and relative contribution: \textit{baseline + conventional EO}, a 28-band stack comprising \textit{baseline} plus annual median Sentinel-1/2 composites and vegetation index temporal dynamics (Table~\ref{tab:datasets_predictors}e--f); \textit{baseline + AlphaEarth}, a 70-band stack comprising \textit{baseline} plus AlphaEarth embeddings (Table~\ref{tab:datasets_predictors}g); and \textit{baseline + TESSERA}, a 134-band stack comprising \textit{baseline} plus TESSERA v2 embeddings (hereafter TESSERA; Table~\ref{tab:datasets_predictors}h). We intentionally used analysis-ready products for the conventional EO feature set to ensure that the acquisition effort was comparable to the GFM embeddings. A shared no-data pixel mask was applied to all layers to ensure each feature set covered all pixels included in the analysis.

To determine whether introducing spatial context affected model performance, we defined four model architectures: XGBoost \parencite{chen_xgboost_2016} with a single-pixel input and CNNs with $3\times3$-, $5\times5$- and $7\times7$-pixel receptive fields (Fig.~\ref{fig:methods}d). All models output single-pixel old-growth scores (0--1), which were aggregated to parcel-level scores as described in Section~\ref{sec:methods_metrics}. We evaluated the performance of the four architectures against each of the five feature sets, resulting in 20 configurations.

\subsection{Spatial autocorrelation}\label{sec:spatial_autocorrelation}

We quantified the extent and magnitude of spatial autocorrelation in our reference labels and predictors to inform the fold separation and train-test buffers applied during cross-validation (Section~\ref{sec:methods_spatial_cv}). Empirical semivariograms \parencite{matheron_principles_1963} were calculated at 1\,km lag bins to a maximum lag of 40\,km (approximately half the maximum extent of the AOI, following \cite{garrigues_quantifying_2006}), and exponential, Gaussian and spherical models were fitted to obtain three estimates of the effective range. Fitted ranges at or above 90\% of the maximum lag, or below the 1\,km lag bin width, were reported as unresolved. Moran's $I$ \parencite{moran_notes_1950} was evaluated at 5, 10, 15 and 20\,km distance bands; we report the 10\,km band here and the others in Table~\supref{tab:sup_autocorrelation_full}.

\begin{figure}[!t]
    \centering
    \includegraphics[width=1\linewidth]{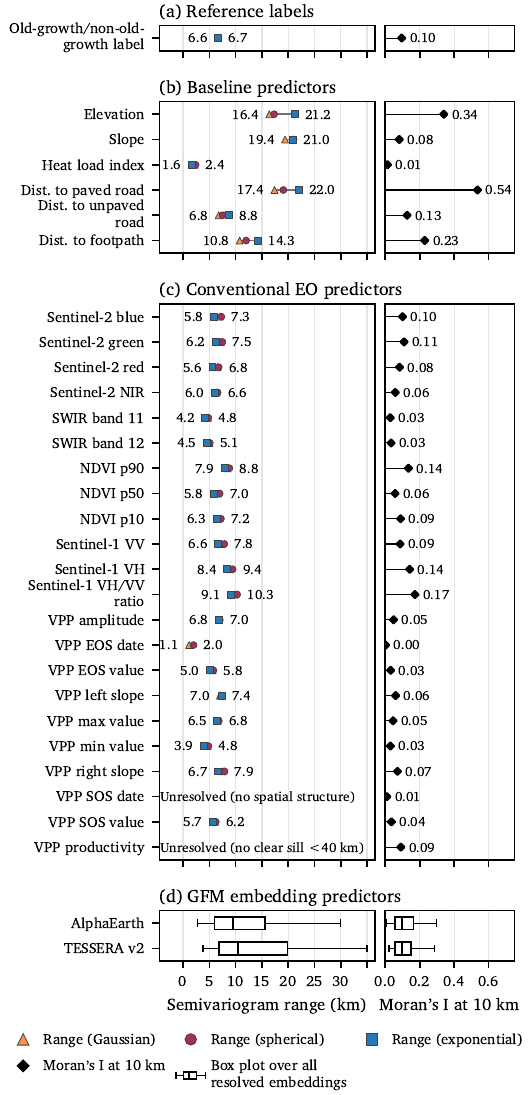}
    \caption{Spatial autocorrelation in: (a) reference labels, (b) baseline predictors, (c) conventional Earth observation (EO) predictors, (d) geospatial foundation model (GFM) embeddings. Left: effective ranges from Gaussian, spherical and exponential models fitted to an empirical semivariogram; right: Moran's $I$ at the 10\,km distance band. Box plots summarise resolved embedding dimensions. Predictors follow Table~\ref{tab:datasets_predictors}: Dist., distance; EOS, end-of-season; NDVI, normalised difference vegetation index; NIR, near-infrared; p10, p50 and p90, 10th, 50th and 90th percentiles, respectively; SAR, synthetic aperture radar; SOS, start-of-season; SWIR, short-wave infrared; VH, vertical-transmit/horizontal-receive polarisation; VPP, Vegetation Phenology and Productivity; VV, vertical-transmit/vertical-receive polarisation.}
    \label{fig:label_predictor_autocorrelation}
\end{figure}

Both the reference labels and predictors showed spatial dependence at kilometre scales (Fig.~\ref{fig:label_predictor_autocorrelation}). The reference labels had a semivariogram range of 6.6--6.7\,km and Moran's $I$ of 0.10 at 10\,km. The predictors with the longest semivariogram range were elevation (16.4--21.2\,km), slope (19.4--21.0\,km) and distance to paved road (17.4--22.0\,km), with Moran's $I$ correspondingly high for elevation and distance to paved road (0.34 and 0.54, respectively). Most conventional EO predictors exhibited ranges up to approximately 10\,km, except for VPP seasonal productivity and start-of-season date, whose spatial ranges were unresolved. The AlphaEarth and TESSERA embeddings were comparable to each other, with median per-dimension ranges (median of the resolved model fits) of 9.4\,km (interquartile range (IQR) 5.9--15.6\,km) and 10.5\,km (IQR 6.8--19.8\,km), and median Moran's $I$ of 0.10 (IQR 0.06--0.16) and 0.10 (IQR 0.06--0.15), respectively.

\subsection{Spatially blocked nested cross-validation}\label{sec:methods_spatial_cv}

We assigned the reference labels to six spatially blocked folds (Fig.~\ref{fig:methods}b) using a rotating $3\times2$ grid-partitioning algorithm that minimised the summed squared deviation of each fold's total and OGF areas from an equal split, with an enforced $\pm$25\% tolerance (Fig.~\supref{fig:sup_fold_balance}). Adjacent parcels were merged into connected components before fold assignment to ensure touching parcels were not split across folds. The resulting folds had an OGF prevalence of 14--29\% and a median cross-fold distance between parcels of 22--34\,km.

Model performance was evaluated for the 20 feature-set/model-architecture configurations using nested cross-validation across the six spatially blocked folds (Algorithm 1). For each configuration, an outer loop held out each fold in turn as a test set, and the remaining five folds were used in an inner cross-validation loop to tune the hyperparameters with Optuna, a Bayesian optimisation framework \parencite{akiba_optuna_2019}, over $T=50$ trials for XGBoost and $T=30$ trials for the more costly CNNs. Hyperparameter tuning had a moderate impact on performance, with the selected trial achieving a median improvement in parcel PR-AUC of 0.02 over the median trial, and 0.13 (XGBoost) and 0.05 (CNN) over the worst trial (Table~\supref{tab:sup_hp_gaps}). This nested approach ensured that the data used to train and tune a model were never used to estimate its performance. We report out-of-fold (OOF) performance as the primary metric, calculated from the aggregated held-out predictions for all six folds. Confidence intervals (CIs) were obtained from the 2.5$^{\text{th}}$--97.5$^{\text{th}}$ percentiles (95\% CIs) of a bootstrap over 20 area-balanced spatial blocks (Fig.~\supref{fig:sup_bootstrap_balance}), resampled with replacement over 5,000 replicates. Contrasts between feature sets and model architectures were calculated as paired differences on these same replicates and reported with their CIs and a two-sided bootstrap $p$-value, which was equal to twice the smaller of the proportions of replicates in which the difference was at or below, or at or above, zero \parencite{davison_bootstrap_1997}.

\begin{table}[htb]
\begin{minipage}{1\linewidth}
\centering
\begin{tabular}{@{}l@{}}
\toprule
\textbf{Algorithm 1}: Six-fold nested cross-validation with Optuna \\
\midrule
\textbf{for} each outer fold $k \in K = \{1, 2, 3, 4, 5, 6\}$: \\
\quad Hold out fold $k$ as a test set \\
\quad Define inner training folds, $\mathcal{J} = \{1, 2, 3, 4, 5, 6\} \setminus \{k\}$ \\
\quad \textbf{for} each Optuna hyperparameter trial, $\theta$ ($T$ trials): \\
\quad\quad \textbf{for} each inner fold, $j \in \mathcal{J}$: \\
\quad\quad\quad Train on data from $\mathcal{J} \setminus \{j\}$ with $\theta$ \\
\quad\quad\quad Evaluate on fold $j$ \\
\quad\quad Return mean PR-AUC to Optuna \\[4pt]
\quad Select $\theta^*$ with highest mean PR-AUC \\
\quad Retrain on all inner folds, $\mathcal{J}$, with $\theta^*$ \\
\quad Predict fold $k$; store out-of-fold predictions \\[4pt]
Pool the out-of-fold predictions from all $K$ folds \\
Report pooled performance metrics with 95\% confidence intervals \\
\bottomrule
\end{tabular}
\end{minipage}
\end{table}

To determine whether residual cross-fold parcel proximity affected performance estimates, we adapted the buffered approach of \textcite{ploton_spatial_2020} to our nested cross-validation design. Training parcels within a buffered distance $d$ of the held-out test fold were excluded, and the model was refitted with the hyperparameters already selected for that fold. The inner-loop hyperparameter search was flat near its optimum and therefore unlikely to bias the reported comparisons (Table~\supref{tab:sup_hp_gaps}). All 20 model configurations were evaluated at \mbox{$d\in\{0,10,20\}\,\mathrm{km}$}. To separate the effects of training data loss from spatial leakage, the XGBoost models were additionally swept at \mbox{$d\in\{0,1,\ldots,30\}\,\mathrm{km}$}, alongside a parcel-matched control at each distance, in which the same number of training parcels was removed randomly rather than by proximity.

Four sensitivity analyses were conducted to assess: robustness to fold composition (leave-one-fold-out, Tables~\supref{tab:sup_lofo_pr_auc}~and~\supref{tab:sup_lofo_contrasts}); training-pixel sampling (unweighted, inverse-prevalence and stratification, Table~\supref{tab:sup_sampling_sensitivity}); model stochasticity (five training seeds, Tables~\supref{tab:sup_seed_sweep}~and~\supref{tab:sup_seed_sweep_contrasts}); and bootstrap block count (6--30 blocks, Table~\supref{tab:sup_block_size_sensitivity}), none of which changed the conclusions.

\subsection{Mapping}

We retrained our best-performing model on all reference labels to produce wall-to-wall 10\,m resolution rasters with calibrated old-growth probability (0--1) and binary classification (0/1) pixels, and a vector with calibrated probability and binary attributes for 15,443 forest parcels (72 parcels contained no pixel centre and were not classified, see Section~\supref{sec:supp_ref_labels}). The probability estimates were obtained using separate pixel- and parcel-level Platt-scaling models, each fitted to the raw out-of-fold old-growth scores and corresponding reference labels \parencite{platt_probabilistic_2000}, on which the calibration errors were also calculated.

A point estimate of the old-growth forest extent within the AOI was calculated as the area of the parcels with an old-growth score exceeding the F1-maximising threshold on the calibrated out-of-fold predictions. Uncertainty in the mapped old-growth area was estimated using a retraining 20-spatial-block bootstrap across 1,000 replicates, resampled with replacement. Wall-to-wall prediction was re-run across the AOI for each replicate, and the old-growth extent was calculated at the parcel level using the same fixed threshold as the point estimate.

\subsection{Determination of area of applicability}

To assess the extent to which our model could be applied beyond the AOI, we calculated its area of applicability (AOA) within the Carpathian Mountain range using the method of \textcite{meyer_predicting_2021}. The AOA delineates the area for which predictor features resemble a model's training data, and therefore where cross-validated performance can be reasonably expected to apply. Using a 100\,m grid in EPSG:3035, we calculated a dissimilarity index (DI) for all pixels within the Carpathian Mountains. The DI is the distance in feature space to the nearest training pixel, with predictors mean-standardised and weighted by importance (XGBoost gain), and normalised by the mean pairwise distance among training pixels. The AOA comprises pixels whose DI does not exceed the largest cross-validated training DI, excluding outliers (i.e. $Q_3+1.5\times\mathrm{IQR}$, the box plot upper whisker). Each training pixel's DI is calculated relative to the nearest training pixel outside its own fold. A polygon delineating the Carpathian Mountains was obtained from the European Environment Agency geospatial data catalogue \parencite{eea_european_2008}, and forest extent was extracted from cells in the ESA WorldCover 2020 raster with a tree-cover fraction $\geq$0.5 \parencite{zanaga_esa_2021}, as CORINE Land Cover polygons do not cover Ukraine.

\section{Results}

\subsection{Evaluation of existing studies}\label{sec:existing_studies}

\begin{table*}
    \centering
    \caption{Comparison of estimated primary, old-growth and high-conservation-value forest extent and old-growth forest (OGF) reference label agreement for existing studies and this study, evaluated on 4,825 OGF and non-OGF parcels (47,348\,ha) across the forested area of interest (forested AOI, 152,366\,ha). OGF prevalence = 21\%. PR-AUC, precision-recall area under the curve; ROC-AUC, receiver operating characteristic area under the curve.}
    \label{tab:existing_studies_aoi_area_performance}
    \setlength{\tabcolsep}{2.5pt}
    \resizebox{\textwidth}{!}{%
    \begin{threeparttable}
        \begin{tabular}{llllcccccccc}
            \toprule
            \multirow{2}{*}[-2.5pt]{Study} & \multirow{2}{*}[-2.5pt]{Extent} & \multirow{2}{*}[-2.5pt]{Method} & \multirow{2}{*}[-2.5pt]{Forest target} & \multicolumn{2}{c}{Predicted area} & \multicolumn{5}{c}{Agreement with Făgăraș reference labels} & \multirow{2}{*}[-2.5pt]{\shortstack{Published\\ROC-AUC$^{\dagger}$}} \\
            \cmidrule(lr){5-6} \cmidrule(lr){7-11}
            & & & & ha & \% forested AOI & PR-AUC & ROC-AUC & F1 & Precision & Recall & \\
            \midrule
            \textcite{sabatini_spatially-explicit_2020} & Europe & Model-based & Primary & 20,865 & 13.7 & 0.35 & 0.68 & 0.43 & 0.44 & 0.42 & 0.65 \\
            \textcite{munteanu_using_2022} & Romania & Model-based & High-conservation-value & 61,837 & 40.6 & 0.54 & 0.86 & 0.61 & 0.51 & 0.76 & 0.86 \\
            \textcite{kathmann_potential_2017} & Romania & Rule-based & Primary \& old-growth & 64,146 & 42.1 & 0.41 & 0.78 & 0.56 & 0.43 & 0.77 & -- \\
            \textcite{schickhofer_inventory_2019} & Romania & Rule-based & Primary \& old-growth & 51,235 & 33.6 & 0.65 & 0.92 & 0.74 & 0.64 & 0.90 & -- \\
            \midrule\midrule
            This study (0\,km buffer)$^{\ddagger}$  & \multirow{2}{*}{Făgăraș} & \multirow{2}{*}{Model-based} & \multirow{2}{*}{Old-growth} & 55{,}025 & 36.1 & 0.84 & 0.95 & 0.77 & 0.74 & 0.80 & -- \\
            This study (10\,km buffer)$^{\ddagger}$ &  &  &  & -- & -- & 0.73 & 0.91 & 0.70 & 0.66 & 0.75 & -- \\
            \bottomrule
        \end{tabular}
        \begin{tablenotes}[flushleft]\footnotesize
            \item $^{\dagger}$ Published ROC-AUC is the authors' own reported value computed on their reference data, which, for \textcite{sabatini_spatially-explicit_2020}, is the estimate corrected for spatial sorting bias and, for \textcite{munteanu_using_2022}, is the structural complexity model. Rule-based products have no equivalent metric.
            \item $^{\ddagger}$ This study shows the point estimates only for the \textit{baseline + TESSERA} XGBoost model. Confidence intervals are shown in Fig.~\ref{fig:performance_matrix} and Table~\supref{tab:sup_performance_matrix}.
        \end{tablenotes}
    \end{threeparttable}%
    }
\end{table*}

\begin{figure*}
    \centering
    \includegraphics[width=1\linewidth]{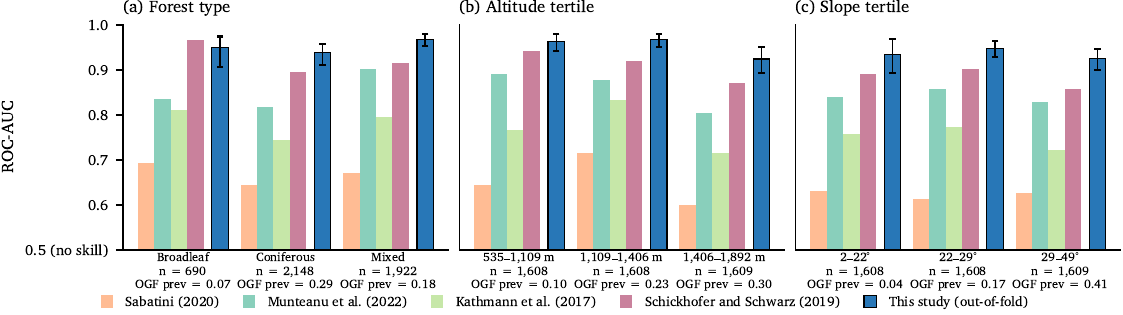}
    \caption{Stratified receiver operating characteristic area under the curve (ROC-AUC) across (a) forest type, (b) altitude tertile and (c) slope tertile for existing studies and this study's XGBoost model trained with \textit{baseline + TESSERA}. Error bars show 95\% confidence intervals for this study's model. Forest type was derived from CORINE Land Cover 2018 \parencite{clms_corine_2020}. $n$, number of parcels; prev., OGF prevalence.}
    \label{fig:stratification}
\end{figure*}

\begin{figure*}
    \centering
    \includegraphics[width=1\linewidth]{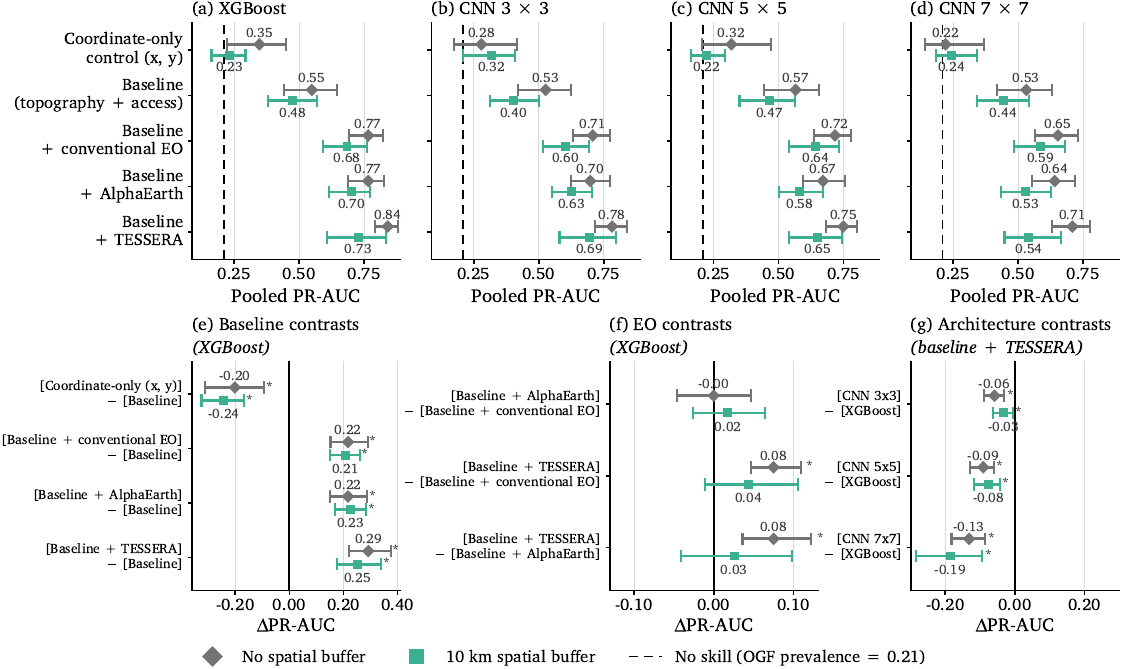}
    \caption{Pooled out-of-fold parcel-level precision-recall area under the curve (PR-AUC) across feature sets and model architectures: (a--d) performance by feature set and model architecture, (e) contrasts against baseline for XGBoost, (f) pairwise contrasts among Earth observation (EO)-derived feature sets for XGBoost, (g) performance contrasts of convolutional neural network (CNN) receptive-field sizes against XGBoost for \textit{baseline + TESSERA}. Points show estimates and bars show 95\% confidence intervals; * denotes a contrast whose two-sided bootstrap $p$-value is below 0.05. Underlying metrics and results, including for the 20\,km spatial buffer arm, are provided in Tables~\supref{tab:sup_performance_matrix}--\supref{tab:sup_architecture_contrasts}. \(\Delta\)PR-AUC, paired difference in PR-AUC; OGF, old-growth forest; TESSERA, TESSERA v2 embeddings.}
    \label{fig:performance_matrix}
\end{figure*}

The four existing studies differed threefold in the mapped extent of primary, old-growth and high-conservation-value forest across the AOI (Table~\ref{tab:existing_studies_aoi_area_performance}). Estimates ranged from 20,865\,ha (13.7\% of the forested AOI) for \textcite{sabatini_spatially-explicit_2020} to 64,146\,ha (42.1\%) for \textcite{kathmann_potential_2017}. Native pixel-level area estimates differed from the parcel-level extents by less than 1\% for all products except \textcite{sabatini_spatially-explicit_2020}, which differed by $+$13.3\% (2,779\,ha, Table~\supref{tab:sup_area_parcel_vs_native}).

Agreement with our OGF/non-OGF reference labels varied substantially across the four studies; however, the ordering of studies was stable whether using PR-AUC, ROC-AUC or F1 (Table~\ref{tab:existing_studies_aoi_area_performance}). For the two model-based studies, the ROC-AUC on our reference labels reproduced the authors' published values almost exactly: 0.68 against 0.65 for \textcite{sabatini_spatially-explicit_2020} (reported in \cite{sabatini_protection_2020}) and 0.86, unchanged, for \textcite{munteanu_using_2022}. Label agreement was highest for \textcite{schickhofer_inventory_2019}, a field-calibrated rule-based map. With a PR-AUC of 0.65, it recovered 90\% of our OGF reference parcels with 64\% precision. Agreement was lowest for \textcite{sabatini_spatially-explicit_2020}, a Europe-wide model-based map derived from geospatial proxies, which achieved a PR-AUC of 0.35 against the 0.21 no-skill floor. This ordering was consistent across different landscapes within the AOI, with \textcite{schickhofer_inventory_2019} ranking highest and \textcite{sabatini_spatially-explicit_2020} lowest across forest type, altitude and slope strata (Fig.~\ref{fig:stratification}, Table~\supref{tab:sup_stratified_roc_auc}). Nominal performance was weakest at the highest altitudes (1,406--1,892\,m) for all studies. Rule-based approaches did not universally achieve higher agreement. The model-based \textcite{munteanu_using_2022} exceeded the rule-based \textcite{kathmann_potential_2017} by 0.13 PR-AUC, with greater precision (51\% versus 43\%) for a comparable recall (76\% versus 77\%).

Analysis of the two model-based studies' continuous probability surfaces showed near-complete overlap in the OGF and non-OGF parcel score distributions for \textcite{sabatini_spatially-explicit_2020} (0.80 overlap coefficient), and partial separation for \textcite{munteanu_using_2022} (0.44), indicating limited discrimination between the two classes. The corresponding histograms showed that the distribution overlap was caused by non-OGF parcels receiving high-confidence OGF scores (Fig.~\supref{fig:histogram}).

\subsection{Contribution of Earth observation feature sets}\label{section:eo_contribution}

All EO-derived feature sets contained information predictive of old-growth forests beyond topographic and access covariates. Adding conventional EO, AlphaEarth embeddings or TESSERA embeddings to the \textit{baseline} feature set increased PR-AUC by 0.22--0.29 without train-test buffering, with all 95\% CIs above zero. This improvement persisted when training parcels within 10\,km and 20\,km of the held-out test folds were excluded, with PR-AUC gains ranging from 0.21 to 0.25 [95\%~CIs 0.15 to 0.34] under a 10\,km buffer (Fig.~\ref{fig:performance_matrix}e, Tables~\supref{tab:sup_performance_matrix}~and~\supref{tab:sup_feature_set_contrasts}).

The best-performing configuration was XGBoost trained on the \textit{baseline + TESSERA} feature set with a 0\,km buffer, which achieved a PR-AUC of 0.84 [95\%~CI 0.79--0.88] (Fig.~\ref{fig:performance_matrix}a). This configuration performed consistently across metrics (ROC-AUC 0.95 [0.94--0.97]; F1 0.77 [0.71--0.81], Table~\supref{tab:sup_performance_matrix}) and forest type, altitude and slope strata (Fig.~\ref{fig:stratification}). Per-fold performance is reported in Tables~\supref{tab:sup_per_fold_pr_auc},~\supref{tab:sup_per_fold_roc_auc}~and~\supref{tab:sup_per_fold_f1}.

TESSERA's advantage over conventional EO and AlphaEarth was statistically significant without a train-test buffer, with respective differences in PR-AUC of $+$0.08 [95\%~CI 0.05--0.11; $p < 0.001$] and $+$0.08 [0.04--0.12; $p < 0.001$] (positive in 6/6 folds). However, TESSERA's advantage was not statistically significant at a 10\,km buffer, with conventional EO and AlphaEarth contrasts of $+$0.04 [$-$0.01 to 0.11; $p = 0.13$] (positive in 5/6 folds) and $+$0.03 [$-$0.04 to 0.10; $p = 0.45$] (positive in 4/6 folds), respectively, although the intervals remain compatible with an advantage as large as the unbuffered one (Fig.~\ref{fig:performance_matrix}f, Table~\supref{tab:sup_embedding_contrasts}).

\begin{figure}[!t]
    \centering
    \includegraphics[width=0.9\linewidth]{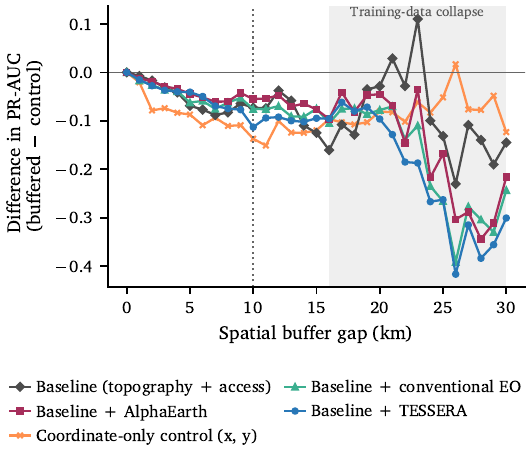}
    \caption{Difference in mean pooled precision-recall area under the curve (PR-AUC) between train-test buffered XGBoost models and parcel-matched controls with the same number of training parcels removed at random. The vertical dotted line marks the selected 10\,km spatial buffer; the grey region indicates training-data collapse ($<$20\% remaining in one or more folds). EO, Earth observation.}
    \label{fig:buffer_performance}
\end{figure}

Residual spatial autocorrelation for the XGBoost configurations ranged from 4.5 to 6.5\,km across the feature sets, with Moran's $I$ $\leq$0.07 at the 10\,km band (Table~\supref{tab:sup_autocorrelation_residuals}). These ranges were at or below those of the labels and EO predictors, and well below those of the baseline predictors, suggesting the model did not learn their long-range ($\geq$10\,km) spatial structure. This residual spatial autocorrelation range was corroborated by a train-test buffer sweep from 0 to 30\,km, which for all feature sets showed a steady fall in the PR-AUC relative to the parcel-matched control before plateauing at a buffer of 8--10\,km (Fig.~\ref{fig:buffer_performance}). This range also corresponded with the coordinate-only control reaching the 0.21 PR-AUC no-skill floor (Figs.~\ref{fig:performance_matrix}a~and~\supref{fig:sup_abs_prauc_spatial_gap}), which, in addition to the collapse in remaining training data at buffer distances $\geq$16\,km, informed our choice of 10\,km for the spatially buffered performance reported in Fig.~\ref{fig:performance_matrix}.

For the parcel-matched control, in which parcels were removed at random rather than by proximity under spatially blocked cross-validation, a 50\% fall in training data reduced PR-AUC by less than 0.04 for the four feature sets (excluding the coordinate-only control), with OGF prevalence unchanged at 21\% (Figs.~\supref{fig:sup_buffer_performance}~and~\supref{fig:sup_abs_prauc_spatial_gap}). This indicated that the quantity of training labels was not a binding constraint.

\begin{figure*}[!t]
    \centering
    \includegraphics[width=1\linewidth]{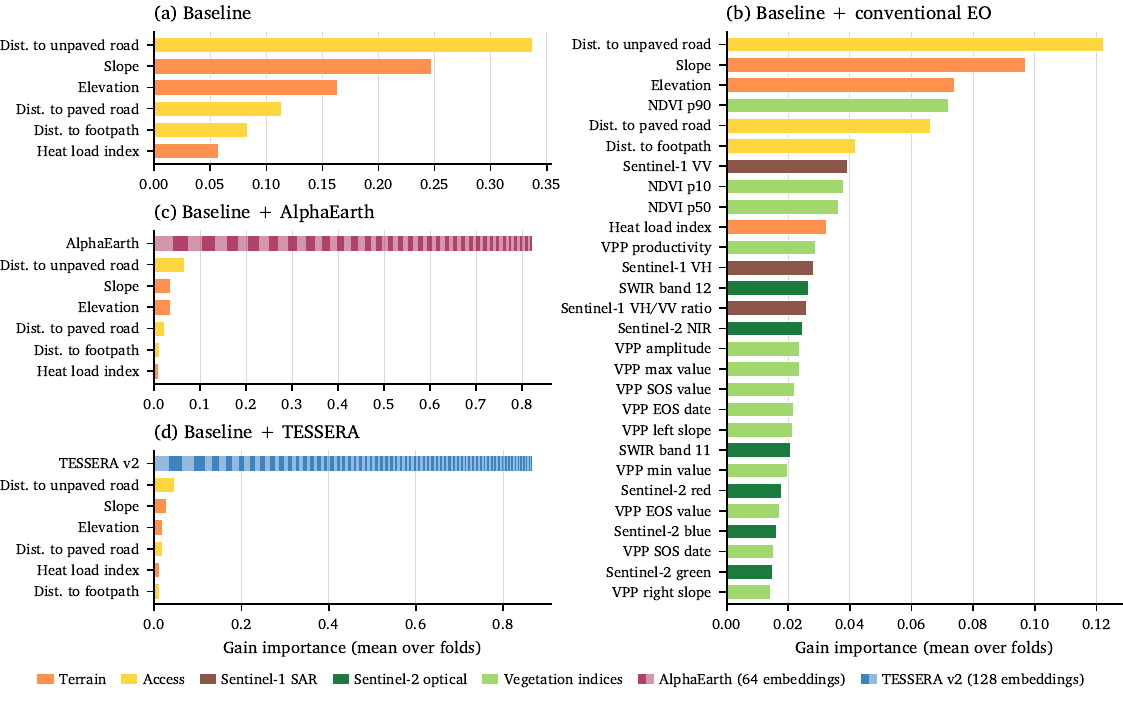}
    \caption{Predictor importance across the four feature sets (excluding the coordinate-only control). Gain-based importance from XGBoost models, averaged across the six spatial cross-validation folds and normalised to sum to one within each feature set: (a) baseline, (b) baseline + conventional Earth observation (EO), (c) baseline + AlphaEarth embeddings, (d) baseline + TESSERA embeddings. The gain values for the 64 AlphaEarth and 128 TESSERA embedding dimensions are aggregated into one bar for each embedding set. Predictors follow Table~\ref{tab:datasets_predictors}: Dist., distance; EOS, end-of-season; NDVI, normalised difference vegetation index; NIR, near-infrared; p10, p50 and p90, 10th, 50th and 90th percentiles, respectively; SAR, synthetic aperture radar; SOS, start-of-season; SWIR, short-wave infrared; VH, vertical-transmit/horizontal-receive polarisation; VPP, Vegetation Phenology and Productivity; VV, vertical-transmit/vertical-receive polarisation.}
    \label{fig:feature_importance}
\end{figure*}

XGBoost gain analysis showed that \textit{baseline} predictors remained important across feature sets, with distance to unpaved road contributing more than any other individual feature in every stack (Fig.~\ref{fig:feature_importance}). Under \textit{baseline + conventional EO}, the vegetation indices were the largest group, with NDVI p10, p50 and p90 contributing more than the six annual median reflectance bands combined. The aggregate AlphaEarth and TESSERA embeddings accounted for 0.82 and 0.87 of the gain, respectively; however, the individual embeddings have no direct physical interpretation. Although AlphaEarth was trained on topographic and land-cover inputs, and therefore \textit{baseline + AlphaEarth} partially double-counted terrain, the baseline features still ranked higher when combined with AlphaEarth than with TESSERA.

\subsection{Contribution of spatial context}\label{section:spatial_context}

Introducing spatial context through a CNN model architecture did not improve parcel-level performance for any feature set (Fig.~\ref{fig:performance_matrix}g, Table~\supref{tab:sup_architecture_contrasts}). At a 0\,km train-test buffer, the performance of the CNN was lower for all feature sets excluding \textit{baseline}, with the best-performing models differing from XGBoost PR-AUC by $-$0.05 [95\%~CI $-$0.09 to $-$0.01; $p = 0.008$] (\textit{baseline + conventional EO}, $5\times5$), $-$0.07 [$-$0.10 to $-$0.03; $p < 0.001$] (\textit{baseline + AlphaEarth}, $3\times3$) and $-$0.06 [$-$0.09 to $-$0.03; $p < 0.001$] (\textit{baseline + TESSERA}, $3\times3$). While the $5\times5$ CNN improved the PR-AUC point estimate by 0.02 on the \textit{baseline} feature set, the 95\% confidence intervals crossed zero. Point PR-AUC performance degraded with receptive-field size for both GFMs. The performance penalty for the CNN models was less conclusive when evaluated at the pixel level, with all \textit{baseline} and two \textit{baseline + AlphaEarth} configurations showing a moderate improvement of up to 0.03 in PR-AUC point estimates (Table~\supref{tab:sup_pixel_performance_matrix}), suggesting the CNN architecture interacted negatively with parcel aggregation.

\subsection{Mapping old-growth in the Făgăraș Mountains}\label{section:mapping}

Using the XGBoost model trained on \textit{baseline + TESSERA}, we classified 55,025\,ha [39,124--62,515\,ha] as candidate old-growth forests within the AOI, covering 36.1\% [25.7--41.0\%] of the forested landscape (Fig.~\supref{fig:sup_area_bootstrap}). This model was used because it had the highest performance with a 0\,km buffer, the regime that best matches interpolation within the AOI where training and target parcels are interspersed rather than spatially separated. Of the 4,294 forest parcels classified as OGF, 2,031 were coniferous (19,195\,ha, 12.6\% of the forested AOI), 1,880 mixed (29,918\,ha, 19.6\% of the forested AOI) and 352 broadleaf (5,710\,ha, 3.7\% of the forested AOI), with the remaining 31 parcels (202\,ha) unclassified by CORINE Land Cover \parencite{clms_corine_2020}. The calibrated pixel-level raster (Fig.~\ref{fig:methods}e) and forest parcels vector (Fig.~\ref{fig:methods}f) had in-sample expected calibration errors (ECE) of 0.5\% and 2.0\%, and maximum calibration errors (MCE) of 1.9\% and 13.6\%, respectively.

Of the four existing studies, our model's predictions most closely aligned with \textcite{schickhofer_inventory_2019}, a field-calibrated expert interpretation of very-high-resolution imagery, which assigns the same OGF/non-OGF class to 83\% of parcels within the AOI (Fig.~\supref{fig:cross_study_agreement}). Excluding parcels with a reference label, which share partial provenance with \textcite{schickhofer_inventory_2019} via a 2001--2005 inventory \parencite{veen_virgin_2010}, agreement averaged 81\%, ranging from 67\% to 87\% across stand-age strata, and was higher for parcels predicted as non-OGF than OGF (Table~\supref{tab:sup_unlabelled_agreement}).

\subsection{Area of applicability within the AOI and Carpathians}

\begin{figure}[!t]
    \centering
    \includegraphics[width=1\linewidth]{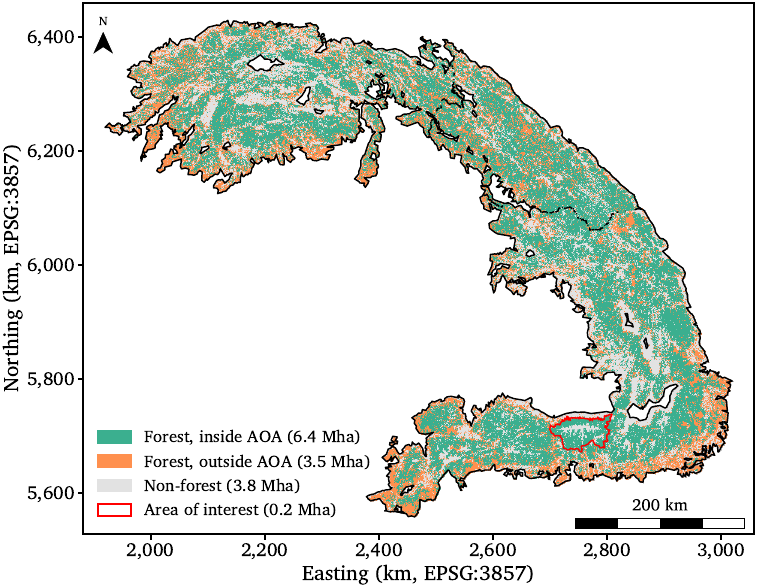}
    \caption{Area of applicability (AOA, \cite{meyer_predicting_2021}) across the Carpathian Mountains for the XGBoost model trained on the \textit{baseline + TESSERA} feature set at 100\,m resolution. Forest is shown inside (green) and outside (orange) the AOA; non-forest land is grey. Of the 9.90 million hectares (Mha) of Carpathian forests, 65.0\% lie within the AOA. The black outline marks the mountain range boundary; the red outline marks the area of interest. Map lines delineate study areas and do not necessarily depict accepted national boundaries.}
    \label{fig:area_of_applicability}
\end{figure}

Within the AOI, 98.4\% of forests were inside the area of applicability (AOA), with a median dissimilarity index (DI) of 0.238, well below the 0.443 threshold. Of the 9.90\,Mha of forest across the entire Carpathian Mountain range, 6.44\,Mha (65.0\%) were inside the AOA (Fig.~\ref{fig:area_of_applicability}). At least 78\% of Carpathian forests between 700 and 2,000\,m in altitude were within the AOA, falling to 18.3\% below 400\,m and 32.0\% above 2,000\,m. Carpathian forests whose nearest training pixel was old-growth totalled 1.02\,Mha, of which 68.8\% were within the AOA; the 0.32\,Mha of such forests outside the AOA indicates where the model is least supported and where additional reference labels would contribute most. 

\section{Discussion}

\subsection{Overview}

Existing maps varied widely both in their agreement with our reference labels and the extent to which they map OGF within the AOI. The two model-based products showed limited discrimination between OGF and non-OGF, resulting in a low PR-AUC, even though ROC-AUC closely matched their published values. The best-performing existing study used a field-calibrated expert interpretation of very-high-resolution imagery and set the benchmark for our locally trained model. Our models show that freely available satellite data contain an old-growth forest signal beyond what topographic and access predictors alone can determine, and that this signal survives spatial separation. Spatial separation did, however, affect the ranking of the EO-based feature sets. TESSERA outperformed conventional EO and AlphaEarth without a train-test buffer, but confidence intervals spanned zero once we imposed a 10\,km buffer. Introducing spatial context through a CNN did not improve parcel-level detection at 10\,m resolution. We provide raster and vector layers of our reference labels and old-growth predictions that cover the forested AOI.

\subsection{Knowing where old-growth forests are not}

Our evaluation of existing studies highlights the importance of true-negative labels for both performance assessment and model training. \textcite{sabatini_spatially-explicit_2020} and \textcite{munteanu_using_2022} derive maps from models trained on presence-only labels, with negatives assumed from pseudo-absences or background points. This limitation is noted by \textcite{bubnicki_conservation_2024} when mapping high-conservation-value forests in Sweden and is likely common across Europe, as labelled datasets, such as the European Primary Forest Database \parencite{sabatini_european_2021}, focus on where the features of interest are rather than where they are not. Although ROC-AUC, the headline metric for the model-based studies, can rank positive-unlabelled models, precision and PR-AUC are confounded by assumed rather than confirmed negatives \parencite{bekker_learning_2020}. This is consequential when, as is the case for old-growth forests, the positive class is rare, because even a low false-positive rate can cause precision to collapse. Both \textcite{sabatini_spatially-explicit_2020} and \textcite{munteanu_using_2022} reproduced their published ROC-AUC on our reference labels; however, precision and PR-AUC were low, and there was high overlap between their OGF and non-OGF prediction distributions. This limits their usefulness for parcel-level decisions within the AOI. True-negative labels are also a plausible contributor to the cleaner separation achieved by our models, likely supported by a heterogeneous mix of non-OGF labels spanning different forest types and terrain (Fig.~\supref{fig:reference_label_statistics}).

\subsection{EO outperforms simple topographic and access features}

Old-growth forest detection has been framed primarily as a structural measurement problem contingent on ALS-derived forest metrics, resulting in a body of work concentrated where ALS surveys exist \parencite{hirschmugl_review_2023, martin_old_2022, fuhr_detecting_2022, adiningrat_mapping_2024, trouve_identifying_2024}. Although previous studies have demonstrated that old-growth forest detection is plausible from satellite EO alone \parencite{spracklen_identifying_2019, munteanu_using_2022}, these studies have not isolated EO's contribution beyond simple topography and access predictors (both of which can be derived from optical and radar data, \cite{farr_shuttle_2007, rolf_generalizable_2021}). Our results show that predictive skill is attributable to optical and radar satellite data and is robust to spatial validation and train-test buffering. This presents a pathway to large-scale mapping of old-growth forests independent of ALS acquisition.

\subsection{Spatial autocorrelation can inflate performance under blocked validation}

Our results agree with previous findings that spatial blocking is necessary but not sufficient to control spatial autocorrelation \parencite{roberts_cross-validation_2017,ploton_spatial_2020,karasiak_spatial_2022}. Our block sizes were informed by the spatial autocorrelation ranges measured in the reference labels and predictors, following guidance that these structures be assessed on the raw data rather than on residuals alone \parencite{roberts_cross-validation_2017}. Larger blocks might have reduced residual spatial autocorrelation, but at the cost of fewer folds and less stable pooled estimates, and without resolving cross-fold parcel proximity at the block borders. Only train-test buffering can remove these pairs directly, and three diagnostics support 10\,km as the buffer for spatially independent performance: the out-of-fold residual semivariogram range (4.5--6.5\,km), the distance at which buffered performance stops falling relative to a parcel-matched control (8--10\,km), and the distance at which the coordinate-only control falls to the no-skill floor (9\,km). However, whether spatial autocorrelation should be controlled or exploited depends on the task. Spatial validation can be overly pessimistic when the modelling objective is interpolation rather than transfer \parencite{wadoux_spatial_2021, roberts_cross-validation_2017}. We therefore compared feature sets and architectures using the 10\,km buffered configuration, and predicted within the AOI using the unbuffered configuration, where unlabelled parcels sit among training parcels and spatial autocorrelation can be exploited justifiably \parencite{roberts_cross-validation_2017, kattenborn_spatially_2022}.

Spatial validation and buffering can change or even reverse a study's conclusions. Here, TESSERA's advantage over conventional EO and AlphaEarth was no longer statistically significant once a 10\,km buffer prevented the model from exploiting training parcels near the fold boundaries. The point estimates still favoured TESSERA, but the intervals were also consistent with no difference. This matters, particularly given the lower interpretability, higher storage requirements, and longer training times associated with high-dimensional inputs. TESSERA's embeddings were calculated per pixel \parencite{feng_tessera_2026, feng_tessera_v2_2026} and exhibited spatial autocorrelation similar to that of AlphaEarth within our AOI, in both the embeddings and model residuals. The unbuffered comparative advantage is therefore unlikely to be attributable to spatial encoding within the embeddings and could instead arise from how our models exploit their spatial structure -- an interaction that warrants further research.

A recent systematic review of 1,122 geospatial AI and GFM studies recommends benchmarks for spatial generalisation, including out-of-region transfer \parencite{zhou_measuring_2026}. Our findings illustrate this need and fall between two emerging strands in the literature. One reports that GFMs outperform hand-engineered optical and radar features \parencite{feng_tessera_2026,feng_tessera_v2_2026, brown_alphaearth_2025, ball_geospatial_2026, yang_evaluating_2026}. The other finds limited spatial transferability for AlphaEarth embeddings compared to conventional EO on agricultural tasks \parencite{ma_harvesting_2026}, and lower performance for biomass estimation in regenerating forests \parencite{rojas-lucero_spectral_2026}. We therefore recommend that comparisons between GFM and conventional EO performance be interpreted with reference to the extent to which spatial independence was enforced.

\subsection{Spatial context introduced by CNNs adds no benefit at 10\,m spatial resolution}

Horizontal structural complexity is a widely accepted indicator of old-growth conditions \parencite{spies_ecological_2004, wirth_old-growth_2009, european_commission_commission_2023}. \textcite{spracklen_identifying_2019} found that adding Sentinel-derived grey-level co-occurrence matrix (GLCM) texture features improved overall accuracy by 3.3 percentage points, and \textcite{hirschmugl_classification_2026} ranked texture features among the most important in some sites. CNNs are therefore an intuitive model architecture for detecting old-growth conditions. This expectation did not hold in our experiments, however, which found that CNNs did not improve parcel-level performance for any feature set. While this may be due to pixel-to-parcel aggregation -- we observed modest CNN gains for some configurations when assessed at the pixel level -- it is more likely driven by spatial resolution. Our largest CNN spans 70\,m, sufficient to contain the structural elements of old-growth stands. However, the textural information that a CNN can exploit diminishes once pixels are coarser than the objects of interest \parencite{woodcock_factor_1987, kattenborn_review_2021}, and the horizontal structures that distinguish old-growth forests may be poorly resolved at 10\,m resolution. In the primeval forests of the Ukrainian Carpathians, for example, 98\% of natural canopy gaps are smaller than 200\,m$^2$ \parencite{hobi_gap_2015}, roughly equal to two 10\,m resolution pixels. \textcite{lalechere_assessing_2024} further found that structural information in optical imagery degrades with coarser pixels. Our results therefore suggest that, at 10\,m resolution, introducing spatial context via a CNN does not contribute more than the denoising that parcel aggregation already provides.

\subsection{Limitations}

Our results come from a single mountain beech--spruce landscape. While the area of applicability extends across 65\% of the Carpathian forests, applicability refers to feature-space support rather than model accuracy, which was not tested outside the AOI or verified in the field. With the exception of \textcite{kathmann_potential_2017}, the existing studies we compared against use reference stands tracing back to a 2001--2005 inventory \parencite{veen_virgin_2010}, from which Romania's National Catalogue of Virgin and Quasi-Virgin Forests underlying our OGF reference labels was partially populated. This may have inflated the existing studies' agreement with our reference labels. Because our reference labels are a non-probability sample, our performance metrics and intervals rely on the fitted predictor--label relationship holding beyond these labels \parencite{stehman_key_2019}. The same constraint applies to our mapped extent, which is not adjusted for omission and commission errors, and whose interval reflects variability from resampling of our reference labels rather than a probability sample \parencite{olofsson_good_2014}.

\subsection{Implications for Europe-wide mapping}

We identify three priorities for future large-scale old-growth forest mapping. First, a greater diversity of reference labels is required, both with respect to new landscapes to expand models' spatial generalisation and area of applicability, and to confirmed negatives to improve class separation and false-positive estimation. Second, old-growth mapping studies require common evaluation metrics \parencite{hirschmugl_review_2023}, and these should be suited to low-prevalence contexts. With primary and old-growth forests representing less than 3\% of Europe's forests \parencite{barredo_mapping_2021}, a continent-wide classifier with no skill can achieve 97\% overall accuracy. Precision-recall metrics are essential for evaluating a model's usefulness, and we recommend their widespread use for this task. Third, testing temporal transfer of models could exploit the advantage of repeat acquisitions offered by EO, including GFMs. This would enable old-growth forests not only to be mapped, but also to be monitored for loss and recovery over time.

\section{Conclusion}

Mapping old-growth forests is a priority for biodiversity conservation in Europe and requires a detection method that is accurate at the forest parcel level yet deployable continent-wide. Here, we show that freely available satellite data can detect old-growth forests beyond topography and access predictors, and our best-performing configuration achieved a PR-AUC of 0.84 [95\%~CI 0.79--0.88] within the AOI. Although TESSERA outperformed conventional EO and AlphaEarth under unbuffered spatially blocked cross-validation, the three EO-derived feature sets' performance was consistent with no difference when a 10\,km buffer was introduced between the training and test labels. Adding spatial context via a CNN showed no advantage over a pixel-based XGBoost model with 10\,m resolution predictors. Our results show that spatially blocked validation is necessary but not sufficient to control spatial autocorrelation, and that train-test buffering is required if the aim is transfer to unseen landscapes. While increased feature and model complexity, including GFMs, may yet yield gains in other contexts, our findings indicate that the limiting factors are the geographic diversity of reference labels and the inclusion of confirmed negatives. For the wider Carpathians, our area of applicability analysis identified the forests where our model was least supported and therefore where new reference labels would contribute most.

\section*{CRediT authorship contribution statement}

\textbf{Thomas Ratsakatika:} Conceptualisation, Methodology, Software, Validation, Formal analysis, Investigation, Data curation, Writing -- original draft, Writing -- review \& editing, Visualisation, Project administration, Funding acquisition. \textbf{Mihai Zotta:} Conceptualisation, Data curation, Investigation, Supervision, Writing -- review \& editing. \textbf{Srinivasan Keshav:} Conceptualisation, Methodology, Supervision, Writing -- review \& editing. \textbf{Emily R. Lines:} Conceptualisation, Methodology, Supervision, Writing -- review \& editing.

\section*{Acknowledgements}

We would like to thank Fundația Conservation Carpathia for facilitating field visits and providing data, contextual insights and expertise. We would also like to thank Ion Holban and Mark Fahey for sharing their deep knowledge and valuable advice.

T.R. was funded by the UKRI Centre for Doctoral Training in the Application of Artificial Intelligence to the Study of Environmental Risks (EP/S022961/1). S.K. was supported by the Robert Sansom Professorship in Computer Science. E.R.L. was funded by a UKRI Future Leaders Fellowship (MR/Y033981/1).

\section*{Data and code availability}

The old-growth forest reference labels and model predictions within the AOI are available on Zenodo \parencite{ratsakatika_old-growth_2026}.

The reference labels and parcel geometries were derived from vector datasets received from Fundația Conservation Carpathia (FCC). FCC compiled these datasets from the approved Natura 2000 management plans for \textit{Munții Făgăraș} and \textit{Râul Târgului--Argeșel--Râușor}, from other forest management plans made available to the public during environmental permitting in Romania, from records published in the \textit{Inspectorul Pădurii} application and from official Ministry of Environment, Waters and Forests applications. Several of these forest management plans have since been superseded, and current data are available via the Ministry of Environment, Waters and Forests GIS portal (\url{https://mmediu.ro/en/portal-gis}, accessed on 15 September 2026).

The prediction data were obtained from public sources listed in Tables~\ref{tab:datasets_groundtruth}~and~\ref{tab:datasets_predictors}. The existing study data are available online for \textcite{munteanu_using_2022} (dataset repository: \cite{munteanu_high_2021}), \textcite{sabatini_spatially-explicit_2020} and \textcite{kathmann_potential_2017} (download link: \url{https://www.greenpeace.org/static/planet4-romania-stateless/2020/11/4f4fa4eb-harta-padurile-virgine-en-kmz-si-shapefile.zip}, accessed 15 September 2026); the \textcite{schickhofer_inventory_2019} vectors must be obtained from the authors directly.

The code used to download these data for the AOI and to reproduce the analysis in this manuscript is available at \url{https://github.com/ratsakatika/detecting-old-growth-forests}.

\section*{Declaration of generative AI and AI-assisted technologies in the manuscript preparation process}

During the preparation of this work, the authors used Claude (Anthropic) to support code development and to polish language. After using this tool, the authors independently reviewed and edited the content as needed and take full responsibility for the content of the published article.

\section*{Declaration of competing interest}

The authors declare that they have no known competing financial interests or personal relationships that could have appeared to influence the work reported in this paper.

\section*{Appendix A. Supplementary material}

Supplementary material to this article can be found after the references.

\printbibliography

\onecolumn
\clearpage
\setcounter{section}{0}\setcounter{figure}{0}\setcounter{table}{0}\setcounter{equation}{0}
\renewcommand{\thesection}{S\arabic{section}}
\renewcommand{\thesubsection}{\thesection.\arabic{subsection}}
\renewcommand{\thesubsubsection}{\thesubsection.\arabic{subsubsection}}
\renewcommand{\thefigure}{S\arabic{figure}}
\renewcommand{\thetable}{S\arabic{table}}
\renewcommand{\theequation}{S\arabic{equation}}
\renewcommand{\theHsection}{supp.\arabic{section}}
\renewcommand{\theHsubsection}{supp.\arabic{section}.\arabic{subsection}}
\renewcommand{\theHsubsubsection}{supp.\arabic{section}.\arabic{subsection}.\arabic{subsubsection}}
\renewcommand{\theHfigure}{supp.\arabic{figure}}
\renewcommand{\theHtable}{supp.\arabic{table}}
\renewcommand{\theHequation}{supp.\arabic{equation}}
\fancyhead[L]{\footnotesize Ratsakatika \textit{et al.} (2026), Supplementary material}

\begin{center}
    {\LARGE Appendix A. Supplementary material: Geospatial embeddings detect old-growth forests but buffered spatial validation narrows their advantage over Sentinel features\par}\label{sec:supp_materials}
    \vspace{1.0em}
    {\large T. Ratsakatika, M. Zotta, S. Keshav, E. R. Lines\par}
    \vspace{0.5em}
    {23 September 2026\par}
\end{center}

\section{Literature search}\label{sec:sup_lit_search}

Web of Science query for existing studies on old-growth forest detection using convolutional neural networks or geospatial foundation models. Search conducted on 18 September 2026, returning 313 results.

\begin{verbatim}
TS=(
  (
    ( "old growth" OR "old-growth" OR primeval OR primaeval OR overmature
    OR "over-mature" OR "late successional" OR "late-successional"
    OR "high conservation value" OR HCVF OR "forest naturalness" )
  OR
    ( (primary OR virgin OR ancient OR intact OR undisturbed OR unmanaged
    OR unlogged OR "never logged" OR naturalness OR "structural complexity")
    NEAR/5 (forest* OR woodland* OR stand OR stands OR tree*) )
  )
AND
  (
    "convolutional neural network*" OR CNN OR "deep learning" OR "neural network*" OR
    "U-Net" OR "semantic segmentation" OR "fully convolutional" OR
    "vision transformer*" OR "self-supervised" OR "foundation model*" OR
    "representation learning" OR "satellite embedding*" OR GeoAI OR AlphaEarth OR
    TESSERA OR Prithvi OR SatMAE OR "Major TOM" OR SkySense OR Satlas OR DOFA OR
    "embedding field*" OR "Google Satellite Embedding*" OR "geospatial embedding*"
  )
)
\end{verbatim}

\begin{table}[H]
\centering
\begin{threeparttable}
\caption{Most relevant results from Web of Science search. Restricted to temperate forest only where old-growth, primary or similar concepts are the explicit prediction target.}
\label{tab:sup_lit_search_records}
\scriptsize
\setlength{\tabcolsep}{3pt}
\renewcommand{\arraystretch}{1.15}
\begin{tabular}{L{1.5cm} L{5cm} L{2cm} L{2cm} L{3.2cm} L{2cm}}
\toprule
Study & Title & Source & DOI & Spatial architecture detects old-growth? & GFM detects old-growth? \\
\midrule

Sohrabi (2026) &
Integrating satellite embeddings and national forest inventory data to map above-ground biomass of temperate deciduous old-growth forests &
Advances in Space Research &
\href{https://doi.org/10.1016/j.asr.2026.04.018}{\nolinkurl{10.1016/j.asr.2026.04.018}} &
No &
Borderline; targets above-ground biomass \\
\midrule

De Assis Barros et al. (2025) &
Mapping the big trees of Vancouver Island with LiDAR, Sentinel-1, Sentinel-2 and deep learning &
IGARSS 2025 &
\href{https://doi.org/10.1109/IGARSS55030.2025.11243688}{\nolinkurl{10.1109/IGARSS55030.2025.11243688}} &
Borderline; targets `big-tree' canopy height &
No \\

\midrule

Li and Li (2023) &
Forest gap extraction based on convolutional neural networks and Sentinel-2 images &
Forests &
\href{https://doi.org/10.3390/f14112146}{\nolinkurl{10.3390/f14112146}} &
Borderline; target is forest canopy gaps &
No \\

\midrule

Hoffmann et al. (2022) &
Predicting species and structural diversity of temperate forests with satellite remote sensing and deep learning &
Remote Sensing &
\href{https://doi.org/10.3390/rs14071631}{\nolinkurl{10.3390/rs14071631}} &
Borderline; targets structural heterogeneity &
No \\

\bottomrule
\end{tabular}
\end{threeparttable}
\end{table}

\section{Materials and methods}

\subsection{Reference label construction: additional details}\label{sec:supp_ref_labels}

A parcel map of 16,199 forest parcel boundaries was used as the base geometry. Although revisions have been made to these boundaries since the layer's compilation (c. 2018), this was the only accessible source providing a contiguous geometry of forest parcels across most of the AOI. The original parcel map contained 1,226 parcels with overlaps (1,129 overlapping pairs), which risked undermining label integrity through feature ambiguity, over-representation and information leakage. A rule-based de-overlap procedure was applied to remove duplicates and slivers. Combined with the removal of 506 parcels outside the AOI and 96 non-productive units (e.g. water bodies, clear-cuts), this resulted in a topologically clean layer of 15,515 parcels for use in the study. Of these 15,515 parcels, 72 contained no pixel centre and were excluded from the parcel-level classification maps, leaving 15,443 classified parcels in total.

The parcel map polygons were enriched with average age and species composition attributes using seven vector layers containing public and private forest management plan data for planning periods between 2013 and 2025. Forest management plan polygons were spatially matched to parcel map polygons where the intersection overlapped at least 90\% of both geometries. This threshold was selected to balance the risk of inaccurate matches against the risk of excluding marginally misaligned parcels. For each matched parcel, species composition (\textit{compoz}, \textit{compoz\_act}), average age (\textit{varsta} or \textit{ta}), and management plan year (\textit{an\_amenaja}, \textit{an}) attributes were transferred to new fields in the parcel map layer. Final average-age attributes were derived by applying data-cleaning and prioritisation rules to the original and enriched parcel map fields. The age and plan-year fields were coerced to numeric data types, with zeros treated as missing data, following expert advice from FCC. For each parcel, the final parcel age was selected from the dated record closest to 2020. Where two records were equally distant from 2020, the earlier one was preferred.

\begin{figure}[H]
    \centering
    \includegraphics[width=0.8\linewidth]{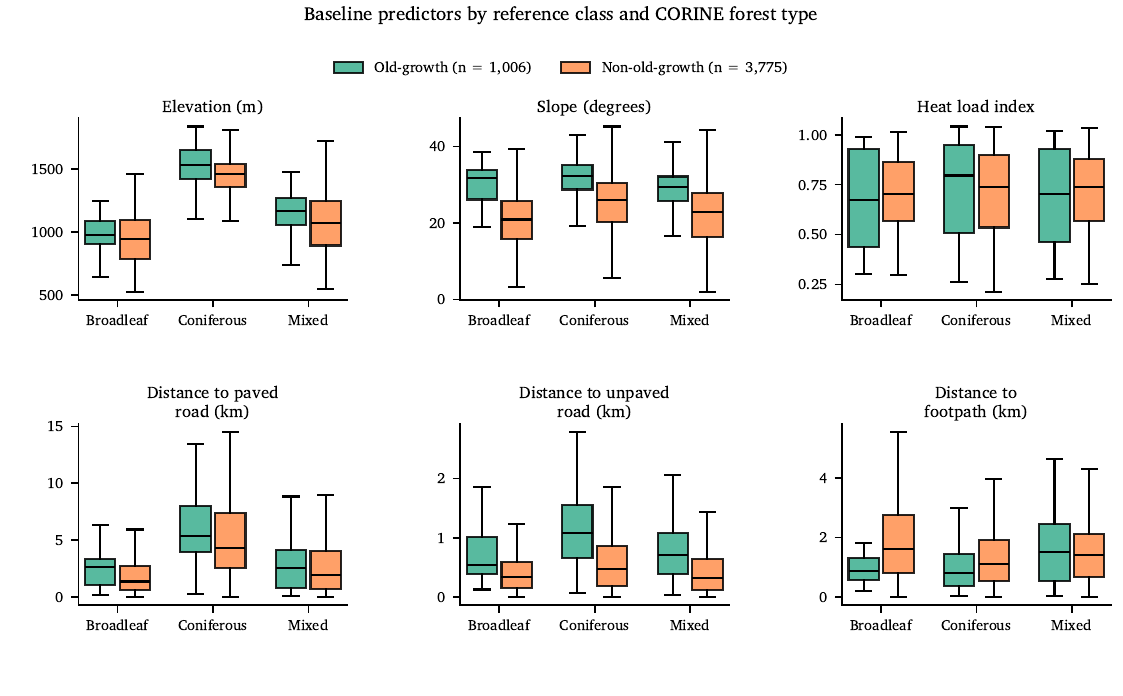}
    \caption{Statistical analysis of reference labels. CORINE Land Cover classifications were unavailable for 66 parcels, and a further six parcels contained no pixels within the predictor mask; both were dropped ($n$ = 1,006 old-growth, 3,775 non-old-growth).}
    \label{fig:reference_label_statistics}
\end{figure}

\subsection{Existing study OGF fraction sensitivity}

\begin{figure}[H]
    \centering
    \includegraphics[width=0.5\linewidth]{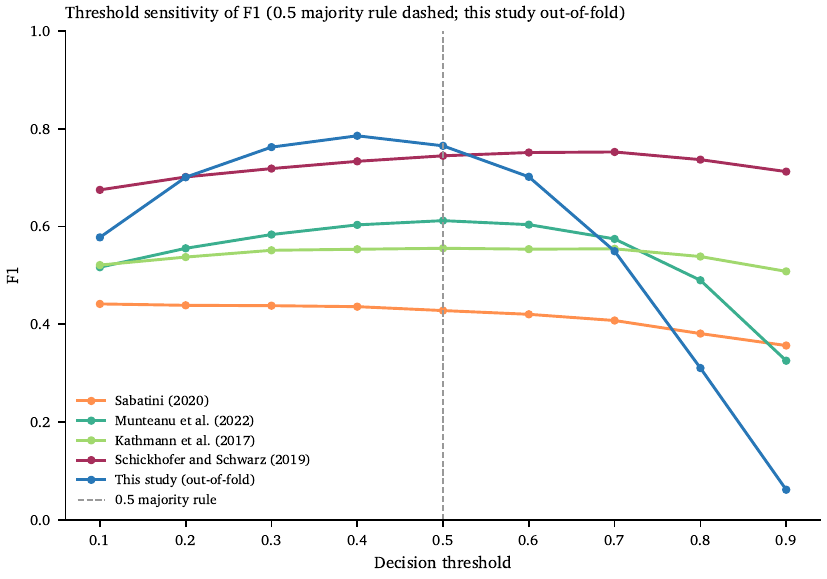}
    \caption{F1 of each product and this study as a function of the decision threshold applied to the parcel OGF fraction. The dashed line marks the selected 0.5 majority rule used for the existing studies.}
    \label{fig:threshold_sensitivity}
\end{figure}

\clearpage

\subsection{Label and predictor autocorrelation analysis}

% ============================================================================
% Supplementary table - Spatial dependence of labels and predictors
% label: tab:sup_autocorrelation_full
% ============================================================================
\begingroup
\footnotesize
\setlength{\tabcolsep}{3pt}
\begin{longtable}{lrrrrlrrrrr}
\caption{Spatial dependence of every variable entering the models and of the reference label, with per-group summary statistics.}\label{tab:sup_autocorrelation_full}\\
\toprule
 & & \multicolumn{3}{c}{Effective range (km)} & & & \multicolumn{4}{c}{Moran's $I$ at distance band} \\
\cmidrule(lr){3-5}\cmidrule(lr){8-11}
Variable & $n$ & Sph. & Exp. & Gau. & Best & Nug.:sill & 5 km & 10 km & 15 km & 20 km \\
\midrule
\endfirsthead
\multicolumn{11}{l}{\footnotesize\textit{Table \ref{tab:sup_autocorrelation_full} continued from the previous page.}}\\
\toprule
 & & \multicolumn{3}{c}{Effective range (km)} & & & \multicolumn{4}{c}{Moran's $I$ at distance band} \\
\cmidrule(lr){3-5}\cmidrule(lr){8-11}
Variable & $n$ & Sph. & Exp. & Gau. & Best & Nug.:sill & 5 km & 10 km & 15 km & 20 km \\
\midrule
\endhead
\midrule
\multicolumn{11}{r}{\footnotesize\textit{Continued on the next page.}}\\
\endfoot
\bottomrule
\endlastfoot
\multicolumn{11}{l}{\textbf{Old-growth reference label} ($k = 1$)}\\[1pt]
\texttt{ogf\_label} & 4853 & 6.72 & 6.63 & 6.67 & Sph. & 0.34 & 0.204 & 0.096 & 0.041 & 0.013 \\
\midrule
\multicolumn{11}{l}{\textbf{Baseline predictors} ($k = 6$)}\\[1pt]
\texttt{dist\_footpath\_m} & 5000 & 12.00 & 14.28 & 10.77 & Exp. & 0.00 & 0.435 & 0.230 & 0.118 & 0.052 \\
\texttt{dist\_paved\_road\_m} & 5000 & 19.08 & 22.01 & 17.40 & Gau. & 0.00 & 0.815 & 0.535 & 0.284 & 0.094 \\
\texttt{dist\_unpaved\_road\_m} & 5000 & 7.50 & 8.79 & 6.77 & Gau. & 0.05 & 0.336 & 0.127 & 0.051 & 0.014 \\
\texttt{elevation\_m} & 5000 & 17.33 & 21.25 & 16.41 & Sph. & 0.05 & 0.605 & 0.340 & 0.181 & 0.090 \\
\texttt{heat\_load\_index} & 5000 & 2.38 & 1.64 & 2.22 & Sph. & 0.59 & 0.027 & 0.013 & 0.008 & 0.006 \\
\texttt{slope\_deg} & 5000 & 21.02 & 20.93 & 19.41 & Exp. & 0.63 & 0.150 & 0.081 & 0.045 & 0.018 \\
\multicolumn{11}{l}{\textit{Group summary (resolved ranges only)}}\\
\quad\textit{Median} &  & 14.66 & 17.60 & 13.59 &  & 0.05 & 0.385 & 0.178 & 0.084 & 0.035 \\
\quad\textit{Min} &  & 2.38 & 1.64 & 2.22 &  & 0.00 & 0.027 & 0.013 & 0.008 & 0.006 \\
\quad\textit{Max} &  & 21.02 & 22.01 & 19.41 &  & 0.63 & 0.815 & 0.535 & 0.284 & 0.094 \\
\midrule
\multicolumn{11}{l}{\textbf{Conventional Earth observation predictors} ($k = 22$)}\\[1pt]
\texttt{ndvi\_p10} & 5000 & 7.20 & 6.34 & 6.70 & Exp. & 0.45 & 0.141 & 0.090 & 0.064 & 0.045 \\
\texttt{ndvi\_p50} & 5000 & 6.96 & 5.78 & 6.66 & Sph. & 0.50 & 0.158 & 0.057 & 0.018 & 0.008 \\
\texttt{ndvi\_p90} & 5000 & 8.82 & 7.88 & 8.40 & Sph. & 0.36 & 0.262 & 0.135 & 0.073 & 0.041 \\
\texttt{s1\_vh} & 5000 & 9.43 & 8.36 & 8.97 & Sph. & 0.34 & 0.283 & 0.142 & 0.072 & 0.033 \\
\texttt{s1\_vh\_vv\_ratio} & 5000 & 10.29 & 9.10 & 9.86 & Sph. & 0.37 & 0.303 & 0.172 & 0.096 & 0.047 \\
\texttt{s1\_vv} & 5000 & 7.84 & 6.65 & 7.35 & Sph. & 0.47 & 0.189 & 0.087 & 0.042 & 0.018 \\
\texttt{s2\_blue} & 5000 & 7.27 & 5.85 & 6.99 & Sph. & 0.43 & 0.203 & 0.100 & 0.055 & 0.031 \\
\texttt{s2\_green} & 5000 & 7.48 & 6.20 & 7.20 & Sph. & 0.36 & 0.233 & 0.110 & 0.059 & 0.033 \\
\texttt{s2\_nir} & 5000 & 6.58 & 5.96 & 6.37 & Exp. & 0.33 & 0.128 & 0.059 & 0.039 & 0.023 \\
\texttt{s2\_red} & 5000 & 6.78 & 5.58 & 6.53 & Sph. & 0.35 & 0.203 & 0.083 & 0.039 & 0.019 \\
\texttt{swir\_b11} & 5000 & 4.80 & 4.23 & 4.74 & Sph. & 0.29 & 0.117 & 0.030 & 0.014 & 0.007 \\
\texttt{swir\_b12} & 5000 & 5.08 & 4.49 & 5.02 & Sph. & 0.28 & 0.131 & 0.034 & 0.014 & 0.007 \\
\texttt{vpp\_ampl} & 5000 & 6.98 & 6.79 & 6.79 & Exp. & 0.54 & 0.107 & 0.048 & 0.025 & 0.013 \\
\texttt{vpp\_eosd} & 5000 & 1.99 & -- & 1.14 & Sph. & 0.85 & 0.006 & 0.002 & 0.001 & -0.000$^\dagger$ \\
\texttt{vpp\_eosv} & 5000 & 5.77 & 5.05 & 5.61 & Exp. & 0.59 & 0.068 & 0.031 & 0.017 & 0.008 \\
\texttt{vpp\_lslope} & 5000 & 7.28 & 7.40 & 7.01 & Exp. & 0.53 & 0.119 & 0.060 & 0.037 & 0.020 \\
\texttt{vpp\_maxv} & 5000 & 6.79 & 6.48 & 6.59 & Exp. & 0.54 & 0.102 & 0.045 & 0.023 & 0.012 \\
\texttt{vpp\_minv} & 5000 & 4.82 & 3.94 & 4.66 & Exp. & 0.72 & 0.049 & 0.030 & 0.021 & 0.013 \\
\texttt{vpp\_rslope} & 5000 & 7.89 & 6.66 & 7.54 & Sph. & 0.62 & 0.142 & 0.071 & 0.037 & 0.020 \\
\texttt{vpp\_sosd} & 5000 & -- & -- & -- & -- & -- & 0.024 & 0.009 & 0.005 & 0.002 \\
\texttt{vpp\_sosv} & 5000 & 6.22 & 5.66 & 6.06 & Exp. & 0.57 & 0.080 & 0.036 & 0.018 & 0.009 \\
\texttt{vpp\_sprod} & 5000 & -- & -- & -- & -- & -- & 0.165 & 0.091 & 0.053 & 0.030 \\
\multicolumn{11}{l}{\textit{Group summary (resolved ranges only)}}\\
\quad\textit{Median} &  & 6.97 & 6.20 & 6.68 &  & 0.46 & 0.136 & 0.060 & 0.037 & 0.019 \\
\quad\textit{Min} &  & 1.99 & 3.94 & 1.14 &  & 0.28 & 0.006 & 0.002 & 0.001 & -0.000 \\
\quad\textit{Max} &  & 10.29 & 9.10 & 9.86 &  & 0.85 & 0.303 & 0.172 & 0.096 & 0.047 \\
\midrule
\multicolumn{11}{l}{\textbf{AlphaEarth embeddings (principal components only)} ($k = 5$)}\\[1pt]
\texttt{pc1} & 5000 & 10.66 & 10.75 & 10.24 & Sph. & 0.16 & 0.404 & 0.205 & 0.106 & 0.053 \\
\texttt{pc2} & 5000 & 29.52 & 33.53 & 27.39 & Exp. & 0.48 & 0.283 & 0.175 & 0.109 & 0.063 \\
\texttt{pc3} & 5000 & 3.25 & 2.76 & 3.29 & Exp. & 0.24 & 0.067 & 0.038 & 0.028 & 0.021 \\
\texttt{pc4} & 5000 & 10.47 & 9.20 & 9.39 & Exp. & 0.62 & 0.115 & 0.070 & 0.037 & 0.015 \\
\texttt{pc5} & 5000 & 19.20 & 18.06 & 17.80 & Exp. & 0.58 & 0.163 & 0.093 & 0.053 & 0.022 \\
\multicolumn{11}{l}{\textit{Summary over all embedding dimensions (resolved ranges only)}}\\
\quad\textit{Median} &  & 9.73 & 8.05 & 8.93 &  & 0.37 & 0.179 & 0.099 & 0.057 & 0.033 \\
\quad\textit{Min} &  & 2.72 & 2.20 & 2.79 &  & 0.06 & 0.038 & 0.007 & 0.002 & -0.001 \\
\quad\textit{Max} &  & 34.49 & 35.61 & 32.50 &  & 0.84 & 0.511 & 0.296 & 0.166 & 0.095 \\
\midrule
\multicolumn{11}{l}{\textbf{TESSERA embeddings  (principal components only)} ($k = 5$)}\\[1pt]
\texttt{pc1} & 5000 & 5.93 & 5.32 & 5.78 & Gau. & 0.38 & 0.186 & 0.065 & 0.028 & 0.015 \\
\texttt{pc2} & 5000 & 27.87 & 35.76 & 26.58 & Exp. & 0.18 & 0.502 & 0.327 & 0.197 & 0.106 \\
\texttt{pc3} & 5000 & 4.73 & 4.64 & 4.53 & Exp. & 0.56 & 0.074 & 0.035 & 0.014 & 0.006 \\
\texttt{pc4} & 5000 & 27.93 & 24.58 & 26.03 & Exp. & 0.82 & 0.057 & 0.035 & 0.027 & 0.019 \\
\texttt{pc5} & 5000 & 12.19 & 10.74 & 11.54 & Exp. & 0.69 & 0.107 & 0.064 & 0.034 & 0.017 \\
\multicolumn{11}{l}{\textit{Summary over all embedding dimensions (resolved ranges only)}}\\
\quad\textit{Median} &  & 10.54 & 10.29 & 10.09 &  & 0.46 & 0.171 & 0.099 & 0.055 & 0.030 \\
\quad\textit{Min} &  & 3.80 & 3.12 & 3.83 &  & 0.02 & 0.054 & 0.021 & 0.008 & 0.002 \\
\quad\textit{Max} &  & 35.39 & 33.40 & 34.98 &  & 0.77 & 0.586 & 0.363 & 0.214 & 0.119 \\
\end{longtable}
\vspace{2pt}
\noindent\footnotesize A range is reported as unresolved (--) when the effective range reaches 90\% of the 40 km maximum lag or falls below the 1 km first lag class. Sph. = spherical, Exp. = exponential, Gau. = Gaussian. \textit{Best} is the lowest-RMSE model with a resolved range. The nugget-to-sill ratio is of the best model. All Moran's $I$ values are significant at $p < 0.05$ except those marked $\dagger$. \par
\endgroup

\subsection{Fold and bootstrap block balance}

% ============================================================================
% Fold and bootstrap balance figures and sensitivity
% ============================================================================

\begin{figure}[H]
    \centering
    \includegraphics[width=0.8\linewidth]{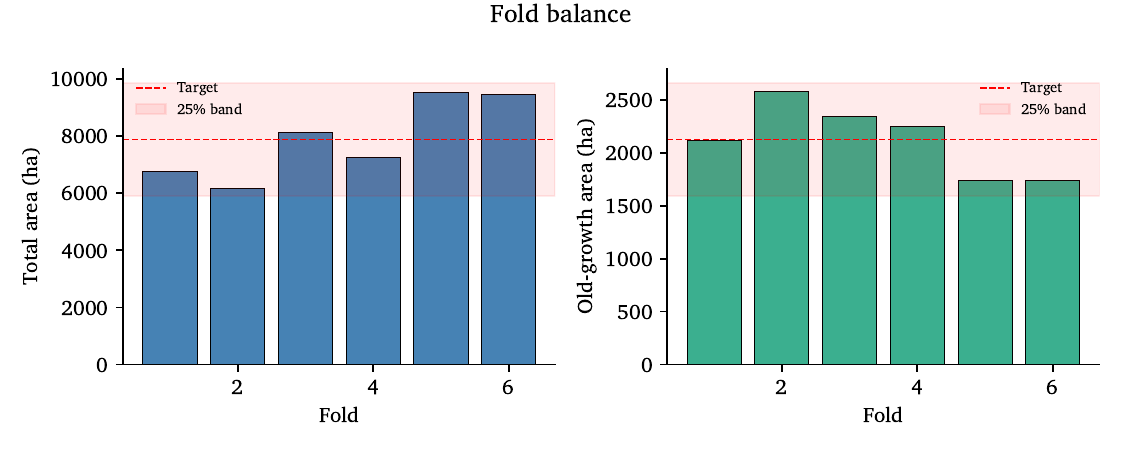}
    \caption{Balance of total and old-growth area across folds.}
    \label{fig:sup_fold_balance}
\end{figure}

\begin{figure}[H]
    \centering
    \includegraphics[width=0.8\linewidth]{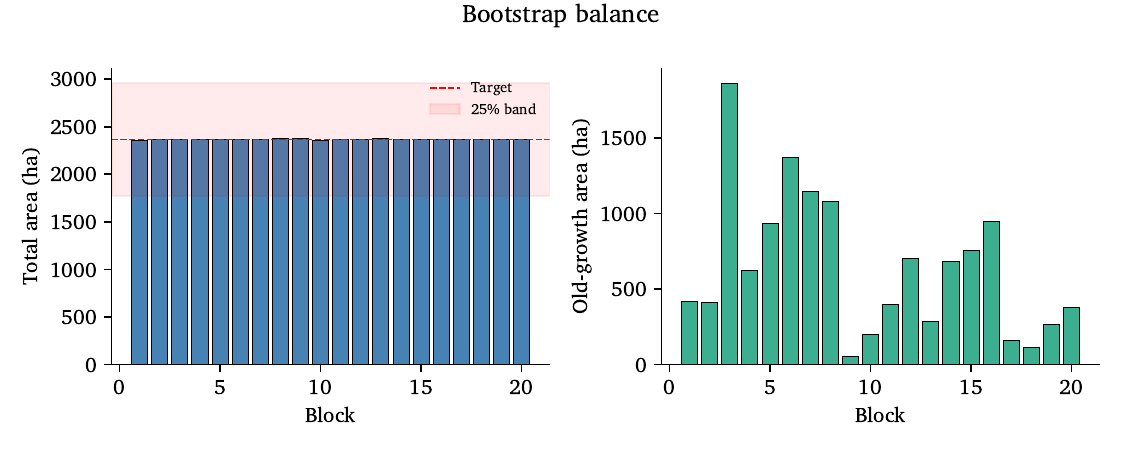}
    \caption{Balance of total and old-growth area across bootstrap blocks (optimisation objective set to area only; OGF balance not optimised).}
    \label{fig:sup_bootstrap_balance}
\end{figure}

\subsection{Sensitivity analysis: block count, hyperparameter search, pixel sampling method}

% ============================================================================
% Supplementary table - Bootstrap block-count sensitivity
% label: tab:sup_block_size_sensitivity
% ============================================================================
\begin{table}[H]
\centering
\caption{Sensitivity of the block-bootstrap interval to the number of spatial blocks, for the XGBoost \textit{baseline + TESSERA} configuration at the 0\,km buffer.}
\label{tab:sup_block_size_sensitivity}
\small
\setlength{\tabcolsep}{6pt}
\begin{threeparttable}
\begin{tabular}{rlcccc}
\toprule
Blocks & Block source & Pooled PR-AUC & Bootstrap mean & 95\% interval & Interval width \\
\midrule
6 & Cross-validation folds & 0.840 & 0.845 & [0.800, 0.889] & 0.088 \\
10 & Spatial partition & 0.840 & 0.841 & [0.786, 0.889] & 0.103 \\
15 & Spatial partition & 0.840 & 0.843 & [0.799, 0.884] & 0.085 \\
\textbf{20} & Saved partition (headline) & 0.840 & 0.840 & [0.793, 0.878] & 0.085 \\
25 & Spatial partition & 0.840 & 0.840 & [0.790, 0.884] & 0.094 \\
30 & Spatial partition & 0.840 & 0.840 & [0.786, 0.888] & 0.102 \\
\bottomrule
\end{tabular}
\end{threeparttable}%
\end{table}

% ============================================================================
% Supplementary table - Hyperparameter-search spread
% label: tab:sup_hp_gaps
% ============================================================================
\begin{table}[H]
\centering
\caption{Spread of inner-fold parcel PR-AUC across hyperparameter search trials, and the flatness of the search near its optimum, summarised as median [min, max] over configuration $\times$ outer-fold combinations.}
\label{tab:sup_hp_gaps}
\small
\setlength{\tabcolsep}{4pt}
\resizebox{\textwidth}{!}{%
\begin{threeparttable}
\begin{tabular}{llrccccc}
\toprule
Family & Feature set & $n$ & Selected $-$ worst & Selected $-$ median & Top-5 range & Top-10 range & Top-20 range \\
\midrule
XGBoost & \textit{All feature sets} & 24 & 0.126 [0.077, 0.207] & 0.015 [0.008, 0.045] & 0.003 [0.001, 0.007] & 0.006 [0.002, 0.022] & 0.010 [0.005, 0.034] \\
 & Baseline & 6 & 0.119 [0.077, 0.170] & 0.012 [0.008, 0.026] & 0.002 [0.001, 0.006] & 0.004 [0.002, 0.009] & 0.008 [0.007, 0.019] \\
 & + Conventional EO & 6 & 0.126 [0.113, 0.167] & 0.026 [0.022, 0.045] & 0.004 [0.001, 0.007] & 0.010 [0.005, 0.022] & 0.021 [0.016, 0.034] \\
 & + AlphaEarth & 6 & 0.117 [0.106, 0.177] & 0.013 [0.008, 0.023] & 0.003 [0.001, 0.005] & 0.005 [0.002, 0.009] & 0.009 [0.005, 0.017] \\
 & + TESSERA & 6 & 0.151 [0.114, 0.207] & 0.015 [0.008, 0.016] & 0.002 [0.001, 0.003] & 0.004 [0.002, 0.009] & 0.009 [0.005, 0.014] \\
\midrule
CNN & \textit{All feature sets} & 72 & 0.050 [0.019, 0.139] & 0.019 [0.006, 0.054] & 0.008 [0.001, 0.026] & 0.013 [0.004, 0.040] & 0.023 [0.007, 0.065] \\
 & Baseline & 18 & 0.062 [0.035, 0.139] & 0.025 [0.016, 0.049] & 0.010 [0.004, 0.026] & 0.016 [0.009, 0.040] & 0.030 [0.019, 0.054] \\
 & + Conventional EO & 18 & 0.040 [0.019, 0.097] & 0.015 [0.008, 0.032] & 0.007 [0.002, 0.021] & 0.010 [0.005, 0.028] & 0.018 [0.009, 0.037] \\
 & + AlphaEarth & 18 & 0.064 [0.029, 0.136] & 0.020 [0.008, 0.054] & 0.010 [0.001, 0.021] & 0.014 [0.005, 0.040] & 0.023 [0.011, 0.065] \\
 & + TESSERA & 18 & 0.042 [0.021, 0.074] & 0.018 [0.006, 0.031] & 0.006 [0.002, 0.011] & 0.011 [0.004, 0.019] & 0.022 [0.007, 0.039] \\
\bottomrule
\end{tabular}
\end{threeparttable}%
}
\end{table}

% % ============================================================================
% Supplementary table - Sampling-mode sensitivity
% label: tab:sup_sampling_sensitivity
% ============================================================================
\begin{table}[H]
\centering
\caption{Cross-validation pixel sampling sensitivity: pooled and per-fold out-of-fold performance of the XGBoost \textit{baseline + TESSERA} model under each sampling mode, at both parcel- and pixel-level. F1 is reported at the F1-maximising threshold on the pooled out-of-fold predictions.}
\label{tab:sup_sampling_sensitivity}
\small
\setlength{\tabcolsep}{4pt}
\begin{threeparttable}
\begin{tabular}{llccccc}
\toprule
 & & \multicolumn{3}{c}{Pooled out-of-fold} & \multicolumn{2}{c}{Across folds: mean [min, max]} \\
\cmidrule(lr){3-5}\cmidrule(lr){6-7}
Sampling mode & Level & PR-AUC & ROC-AUC & $F_1$ & PR-AUC & ROC-AUC \\
\midrule
None (all sampled pixels) & Parcel & 0.840 & 0.954 & 0.789 & 0.865 [0.790, 0.932] & 0.961 [0.939, 0.975] \\
 & Pixel & 0.836 & 0.934 & 0.773 & 0.840 [0.673, 0.939] & 0.943 [0.918, 0.970] \\
\midrule
Inverse prevalence & Parcel & 0.838 & 0.953 & 0.783 & 0.865 [0.794, 0.930] & 0.961 [0.938, 0.974] \\
 & Pixel & 0.835 & 0.933 & 0.771 & 0.840 [0.676, 0.938] & 0.942 [0.919, 0.969] \\
\midrule
Stratified by forest type & Parcel & 0.825 & 0.951 & 0.786 & 0.853 [0.746, 0.919] & 0.957 [0.934, 0.975] \\
 & Pixel & 0.831 & 0.932 & 0.772 & 0.834 [0.635, 0.933] & 0.941 [0.912, 0.967] \\
\bottomrule
\end{tabular}
\end{threeparttable}%
\end{table}

\section{Results}

% ============================================================================
% Supplementary table - Pixel-level performance matrix
% label: tab:sup_pixel_performance_matrix
% ============================================================================
\begin{table}[htbp]
\centering
\caption{Pooled out-of-fold \textbf{pixel-level} performance for every feature set and architecture at the 0\,km training-exclusion buffer, at the pooled outer F1-maximising threshold and at the inner-fold thresholds.}
\label{tab:sup_pixel_performance_matrix}
\small
\setlength{\tabcolsep}{5pt}
\renewcommand{\arraystretch}{0.95}
\resizebox{\textwidth}{!}{%
\begin{threeparttable}
\begin{tabular}{llcccccccccc}
\toprule
 & & \multicolumn{2}{c}{Threshold-free} & \multicolumn{4}{c}{Outer pooled threshold} & \multicolumn{4}{c}{Inner-fold threshold} \\
\cmidrule(lr){3-4}\cmidrule(lr){5-8}\cmidrule(lr){9-12}
Feature set & Architecture & PR-AUC & ROC-AUC & $F_1$ & Precision & Recall & Thresh. & $F_1$ & Precision & Recall & Thresh. \\
\midrule
Baseline & XGBoost & 0.501 & 0.752 & 0.546 & 0.446 & 0.705 & 0.238 & 0.541 & 0.471 & 0.636 & 0.270 \\
 & CNN 3$\times$3 & 0.511 & 0.761 & 0.552 & 0.479 & 0.651 & 0.282 & 0.539 & 0.487 & 0.604 & 0.313 \\
 & CNN 5$\times$5 & 0.533 & 0.749 & 0.539 & 0.459 & 0.653 & 0.264 & 0.530 & 0.515 & 0.546 & 0.359 \\
 & CNN 7$\times$7 & 0.521 & 0.730 & 0.530 & 0.477 & 0.597 & 0.399 & 0.527 & 0.494 & 0.565 & 0.426 \\
\midrule
+ Conventional EO & XGBoost & 0.749 & 0.889 & 0.692 & 0.642 & 0.750 & 0.300 & 0.690 & 0.641 & 0.747 & 0.299 \\
 & CNN 3$\times$3 & 0.722 & 0.894 & 0.709 & 0.646 & 0.785 & 0.301 & 0.705 & 0.652 & 0.767 & 0.316 \\
 & CNN 5$\times$5 & 0.732 & 0.870 & 0.705 & 0.642 & 0.783 & 0.280 & 0.692 & 0.675 & 0.710 & 0.347 \\
 & CNN 7$\times$7 & 0.667 & 0.821 & 0.681 & 0.659 & 0.705 & 0.368 & 0.681 & 0.657 & 0.706 & 0.365 \\
\midrule
+ AlphaEarth & XGBoost & 0.767 & 0.912 & 0.737 & 0.686 & 0.797 & 0.329 & 0.739 & 0.690 & 0.796 & 0.331 \\
 & CNN 3$\times$3 & 0.787 & 0.915 & 0.739 & 0.687 & 0.799 & 0.333 & 0.741 & 0.706 & 0.781 & 0.353 \\
 & CNN 5$\times$5 & 0.775 & 0.899 & 0.730 & 0.696 & 0.767 & 0.402 & 0.728 & 0.691 & 0.770 & 0.388 \\
 & CNN 7$\times$7 & 0.718 & 0.836 & 0.707 & 0.680 & 0.736 & 0.281 & 0.694 & 0.706 & 0.681 & 0.406 \\
\midrule
+ TESSERA & XGBoost & 0.836 & 0.934 & 0.773 & 0.731 & 0.820 & 0.306 & 0.770 & 0.726 & 0.821 & 0.296 \\
 & CNN 3$\times$3 & 0.817 & 0.927 & 0.768 & 0.716 & 0.829 & 0.252 & 0.762 & 0.739 & 0.787 & 0.323 \\
 & CNN 5$\times$5 & 0.782 & 0.907 & 0.745 & 0.688 & 0.812 & 0.391 & 0.742 & 0.706 & 0.783 & 0.425 \\
 & CNN 7$\times$7 & 0.730 & 0.876 & 0.756 & 0.710 & 0.809 & 0.266 & 0.739 & 0.745 & 0.733 & 0.374 \\
\bottomrule
\end{tabular}
\end{threeparttable}%
}
\end{table}

% ============================================================================
% Supplementary figure - histograms
% ============================================================================

\begin{figure}[H]
    \centering
    \includegraphics[width=1\linewidth]{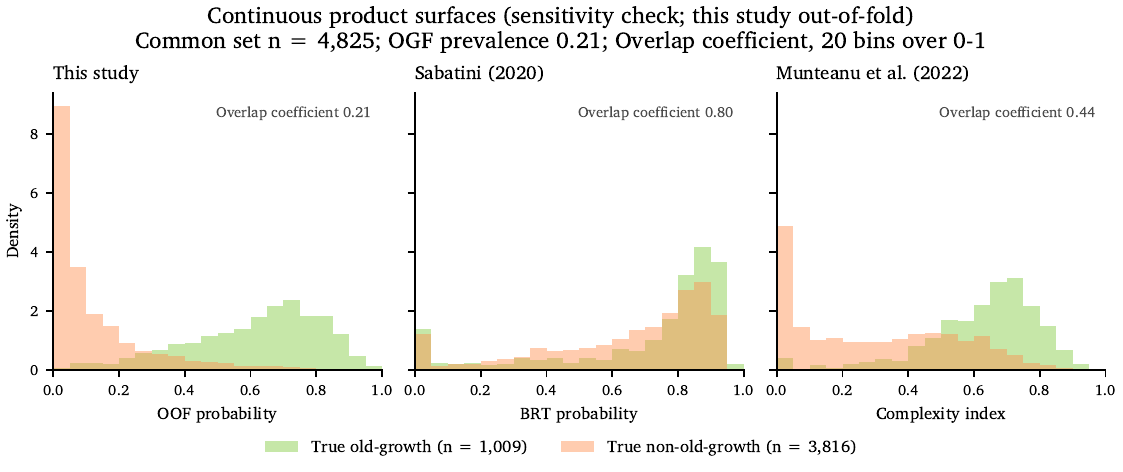}
    \caption{Overlap coefficients and histograms, calculated for the XGBoost model with \textit{baseline + TESSERA}, and the continuous primary forest \textcite{sabatini_spatially-explicit_2020} and forest structural complexity \textcite{munteanu_using_2022} surfaces.}
    \label{fig:histogram}
\end{figure}

% ============================================================================
% Supplementary table - Pooled parcel performance matrix
% label: tab:sup_performance_matrix
% ============================================================================
\begin{landscape}
\begingroup
\scriptsize
\setlength{\tabcolsep}{2pt}
\renewcommand{\arraystretch}{0.92}
\setlength{\aboverulesep}{1pt}
\setlength{\belowrulesep}{1pt}
\begin{longtable}{llrcccccccc}
\caption{Pooled out-of-fold parcel-level performance [95\% CI] for each feature set, architecture and training-exclusion buffer, at both the inner-fold and the pooled outer decision threshold.}\label{tab:sup_performance_matrix}\\
\toprule
 & & & \multicolumn{2}{c}{Threshold-free} & \multicolumn{3}{c}{Inner-fold threshold} & \multicolumn{3}{c}{Outer pooled threshold} \\
\cmidrule(lr){4-5}\cmidrule(lr){6-8}\cmidrule(lr){9-11}
Feature set & Architecture & Buffer (km) & PR-AUC & ROC-AUC & $F_1$ & Precision & Recall & $F_1$ & Precision & Recall \\
\midrule
\endfirsthead
\multicolumn{11}{l}{\footnotesize\textit{Table \ref{tab:sup_performance_matrix} continued from the previous page.}}\\
\toprule
 & & & \multicolumn{2}{c}{Threshold-free} & \multicolumn{3}{c}{Inner-fold threshold} & \multicolumn{3}{c}{Outer pooled threshold} \\
\cmidrule(lr){4-5}\cmidrule(lr){6-8}\cmidrule(lr){9-11}
Feature set & Architecture & Buffer (km) & PR-AUC & ROC-AUC & $F_1$ & Precision & Recall & $F_1$ & Precision & Recall \\
\midrule
\endhead
\midrule
\multicolumn{11}{r}{\footnotesize\textit{Continued on the next page.}}\\
\endfoot
\bottomrule
\endlastfoot
Baseline & XGBoost & 0 & 0.548 [0.441, 0.644] & 0.828 [0.790, 0.859] & 0.550 [0.440, 0.627] & 0.502 [0.391, 0.590] & 0.610 [0.463, 0.719] & 0.572 [0.481, 0.639] & 0.503 [0.401, 0.593] & 0.662 [0.555, 0.741] \\
 &  & 10 & 0.475 [0.381, 0.568] & 0.788 [0.748, 0.828] & 0.503 [0.410, 0.574] & 0.433 [0.333, 0.537] & 0.601 [0.459, 0.725] & 0.521 [0.428, 0.600] & 0.389 [0.297, 0.487] & 0.786 [0.695, 0.866] \\
 &  & 20 & 0.469 [0.356, 0.570] & 0.751 [0.692, 0.806] & 0.467 [0.358, 0.552] & 0.451 [0.343, 0.553] & 0.485 [0.335, 0.630] & 0.483 [0.380, 0.568] & 0.388 [0.288, 0.489] & 0.638 [0.498, 0.761] \\
 & CNN 3$\times$3 & 0 & 0.526 [0.418, 0.623] & 0.826 [0.782, 0.863] & 0.568 [0.480, 0.636] & 0.499 [0.400, 0.590] & 0.661 [0.557, 0.742] & 0.567 [0.481, 0.635] & 0.475 [0.376, 0.565] & 0.703 [0.618, 0.771] \\
 &  & 10 & 0.402 [0.311, 0.501] & 0.772 [0.718, 0.823] & 0.498 [0.404, 0.571] & 0.409 [0.312, 0.512] & 0.636 [0.505, 0.748] & 0.503 [0.413, 0.580] & 0.386 [0.294, 0.482] & 0.723 [0.626, 0.810] \\
 &  & 20 & 0.348 [0.254, 0.477] & 0.728 [0.666, 0.790] & 0.454 [0.362, 0.535] & 0.388 [0.286, 0.509] & 0.547 [0.426, 0.648] & 0.470 [0.374, 0.558] & 0.355 [0.260, 0.459] & 0.696 [0.588, 0.786] \\
 & CNN 5$\times$5 & 0 & 0.566 [0.444, 0.655] & 0.820 [0.772, 0.858] & 0.554 [0.467, 0.619] & 0.538 [0.436, 0.626] & 0.571 [0.465, 0.653] & 0.569 [0.486, 0.636] & 0.476 [0.380, 0.564] & 0.708 [0.621, 0.774] \\
 &  & 10 & 0.465 [0.350, 0.564] & 0.778 [0.722, 0.827] & 0.490 [0.399, 0.564] & 0.428 [0.327, 0.529] & 0.575 [0.446, 0.686] & 0.506 [0.413, 0.583] & 0.392 [0.299, 0.486] & 0.715 [0.605, 0.804] \\
 &  & 20 & 0.339 [0.246, 0.483] & 0.703 [0.646, 0.756] & 0.433 [0.349, 0.506] & 0.370 [0.272, 0.480] & 0.521 [0.417, 0.617] & 0.449 [0.367, 0.521] & 0.344 [0.256, 0.439] & 0.647 [0.568, 0.726] \\
 & CNN 7$\times$7 & 0 & 0.531 [0.418, 0.630] & 0.801 [0.746, 0.848] & 0.546 [0.458, 0.614] & 0.525 [0.421, 0.617] & 0.569 [0.467, 0.654] & 0.556 [0.473, 0.624] & 0.498 [0.398, 0.590] & 0.630 [0.536, 0.709] \\
 &  & 10 & 0.443 [0.342, 0.544] & 0.768 [0.712, 0.820] & 0.496 [0.410, 0.566] & 0.416 [0.321, 0.517] & 0.613 [0.489, 0.716] & 0.503 [0.418, 0.573] & 0.432 [0.335, 0.539] & 0.603 [0.481, 0.703] \\
 &  & 20 & 0.417 [0.311, 0.535] & 0.750 [0.703, 0.792] & 0.488 [0.409, 0.556] & 0.418 [0.321, 0.521] & 0.586 [0.491, 0.672] & 0.492 [0.415, 0.560] & 0.411 [0.315, 0.514] & 0.612 [0.528, 0.687] \\
\midrule
+ Conventional EO & XGBoost & 0 & 0.765 [0.691, 0.823] & 0.925 [0.902, 0.945] & 0.718 [0.641, 0.772] & 0.683 [0.612, 0.741] & 0.757 [0.631, 0.851] & 0.719 [0.644, 0.771] & 0.682 [0.609, 0.737] & 0.760 [0.638, 0.852] \\
 &  & 10 & 0.684 [0.590, 0.759] & 0.891 [0.855, 0.920] & 0.646 [0.551, 0.713] & 0.591 [0.493, 0.674] & 0.712 [0.557, 0.822] & 0.648 [0.556, 0.712] & 0.616 [0.520, 0.697] & 0.683 [0.531, 0.794] \\
 &  & 20 & 0.618 [0.536, 0.697] & 0.867 [0.830, 0.902] & 0.585 [0.494, 0.651] & 0.621 [0.529, 0.715] & 0.553 [0.418, 0.675] & 0.614 [0.527, 0.677] & 0.578 [0.485, 0.667] & 0.654 [0.517, 0.769] \\
 & CNN 3$\times$3 & 0 & 0.707 [0.631, 0.773] & 0.917 [0.891, 0.940] & 0.710 [0.635, 0.765] & 0.630 [0.548, 0.694] & 0.813 [0.704, 0.893] & 0.713 [0.646, 0.763] & 0.613 [0.529, 0.680] & 0.850 [0.772, 0.911] \\
 &  & 10 & 0.602 [0.515, 0.693] & 0.880 [0.841, 0.915] & 0.651 [0.564, 0.723] & 0.553 [0.451, 0.649] & 0.792 [0.665, 0.882] & 0.646 [0.560, 0.718] & 0.560 [0.457, 0.660] & 0.764 [0.638, 0.857] \\
 &  & 20 & 0.494 [0.385, 0.637] & 0.843 [0.785, 0.899] & 0.592 [0.493, 0.683] & 0.510 [0.390, 0.635] & 0.706 [0.583, 0.810] & 0.604 [0.509, 0.694] & 0.492 [0.379, 0.608] & 0.783 [0.680, 0.870] \\
 & CNN 5$\times$5 & 0 & 0.717 [0.638, 0.779] & 0.895 [0.850, 0.931] & 0.693 [0.614, 0.751] & 0.659 [0.587, 0.717] & 0.730 [0.612, 0.826] & 0.698 [0.631, 0.750] & 0.656 [0.582, 0.716] & 0.745 [0.648, 0.826] \\
 &  & 10 & 0.643 [0.541, 0.733] & 0.887 [0.850, 0.918] & 0.658 [0.575, 0.725] & 0.607 [0.504, 0.698] & 0.718 [0.613, 0.803] & 0.669 [0.589, 0.735] & 0.582 [0.471, 0.678] & 0.788 [0.726, 0.843] \\
 &  & 20 & 0.468 [0.344, 0.643] & 0.833 [0.769, 0.894] & 0.611 [0.512, 0.703] & 0.515 [0.390, 0.643] & 0.753 [0.672, 0.825] & 0.611 [0.512, 0.702] & 0.530 [0.398, 0.671] & 0.722 [0.647, 0.789] \\
 & CNN 7$\times$7 & 0 & 0.653 [0.564, 0.729] & 0.871 [0.819, 0.918] & 0.678 [0.585, 0.745] & 0.647 [0.571, 0.706] & 0.712 [0.581, 0.819] & 0.683 [0.596, 0.749] & 0.634 [0.553, 0.692] & 0.739 [0.618, 0.839] \\
 &  & 10 & 0.585 [0.486, 0.679] & 0.846 [0.789, 0.896] & 0.638 [0.540, 0.712] & 0.588 [0.478, 0.679] & 0.699 [0.563, 0.797] & 0.639 [0.542, 0.713] & 0.595 [0.484, 0.688] & 0.690 [0.560, 0.787] \\
 &  & 20 & 0.503 [0.403, 0.622] & 0.801 [0.733, 0.870] & 0.567 [0.476, 0.648] & 0.510 [0.410, 0.611] & 0.637 [0.502, 0.757] & 0.563 [0.477, 0.643] & 0.503 [0.403, 0.609] & 0.640 [0.511, 0.758] \\
\midrule
+ AlphaEarth & XGBoost & 0 & 0.765 [0.689, 0.827] & 0.933 [0.911, 0.953] & 0.738 [0.664, 0.790] & 0.702 [0.626, 0.761] & 0.779 [0.660, 0.866] & 0.743 [0.666, 0.797] & 0.729 [0.653, 0.789] & 0.758 [0.636, 0.848] \\
 &  & 10 & 0.701 [0.616, 0.773] & 0.904 [0.872, 0.933] & 0.678 [0.591, 0.745] & 0.618 [0.518, 0.708] & 0.750 [0.615, 0.855] & 0.678 [0.591, 0.748] & 0.623 [0.519, 0.718] & 0.744 [0.609, 0.851] \\
 &  & 20 & 0.657 [0.580, 0.732] & 0.888 [0.854, 0.922] & 0.623 [0.532, 0.692] & 0.661 [0.571, 0.742] & 0.590 [0.464, 0.697] & 0.652 [0.569, 0.718] & 0.600 [0.511, 0.680] & 0.714 [0.591, 0.815] \\
 & CNN 3$\times$3 & 0 & 0.698 [0.623, 0.773] & 0.917 [0.895, 0.936] & 0.705 [0.632, 0.752] & 0.636 [0.558, 0.699] & 0.792 [0.675, 0.876] & 0.705 [0.629, 0.753] & 0.663 [0.588, 0.726] & 0.753 [0.632, 0.844] \\
 &  & 10 & 0.626 [0.549, 0.705] & 0.883 [0.852, 0.912] & 0.632 [0.550, 0.698] & 0.531 [0.431, 0.632] & 0.782 [0.665, 0.876] & 0.639 [0.560, 0.701] & 0.557 [0.457, 0.657] & 0.749 [0.625, 0.853] \\
 &  & 20 & 0.603 [0.543, 0.675] & 0.868 [0.834, 0.900] & 0.629 [0.556, 0.690] & 0.514 [0.419, 0.618] & 0.812 [0.735, 0.875] & 0.626 [0.554, 0.683] & 0.518 [0.425, 0.620] & 0.790 [0.702, 0.858] \\
 & CNN 5$\times$5 & 0 & 0.671 [0.595, 0.756] & 0.910 [0.884, 0.934] & 0.703 [0.635, 0.754] & 0.634 [0.550, 0.700] & 0.790 [0.689, 0.866] & 0.708 [0.631, 0.763] & 0.670 [0.591, 0.735] & 0.751 [0.631, 0.841] \\
 &  & 10 & 0.582 [0.503, 0.670] & 0.859 [0.818, 0.899] & 0.613 [0.532, 0.682] & 0.506 [0.405, 0.605] & 0.778 [0.664, 0.873] & 0.617 [0.527, 0.689] & 0.527 [0.419, 0.635] & 0.743 [0.611, 0.850] \\
 &  & 20 & 0.568 [0.481, 0.665] & 0.856 [0.808, 0.900] & 0.600 [0.513, 0.671] & 0.471 [0.371, 0.571] & 0.825 [0.733, 0.901] & 0.608 [0.523, 0.682] & 0.511 [0.403, 0.624] & 0.750 [0.637, 0.843] \\
 & CNN 7$\times$7 & 0 & 0.641 [0.554, 0.719] & 0.878 [0.824, 0.918] & 0.675 [0.583, 0.739] & 0.673 [0.592, 0.740] & 0.676 [0.545, 0.775] & 0.689 [0.607, 0.747] & 0.637 [0.555, 0.704] & 0.750 [0.622, 0.842] \\
 &  & 10 & 0.528 [0.432, 0.625] & 0.834 [0.770, 0.892] & 0.593 [0.498, 0.669] & 0.528 [0.419, 0.641] & 0.675 [0.525, 0.794] & 0.599 [0.507, 0.673] & 0.514 [0.408, 0.620] & 0.717 [0.566, 0.833] \\
 &  & 20 & 0.522 [0.405, 0.640] & 0.840 [0.777, 0.897] & 0.593 [0.491, 0.675] & 0.508 [0.388, 0.629] & 0.713 [0.563, 0.828] & 0.607 [0.509, 0.685] & 0.498 [0.386, 0.607] & 0.777 [0.640, 0.876] \\
\midrule
+ TESSERA & XGBoost & 0 & 0.840 [0.793, 0.878] & 0.954 [0.940, 0.966] & 0.770 [0.713, 0.812] & 0.743 [0.686, 0.795] & 0.800 [0.689, 0.883] & 0.789 [0.746, 0.823] & 0.725 [0.664, 0.779] & 0.864 [0.795, 0.918] \\
 &  & 10 & 0.727 [0.608, 0.832] & 0.913 [0.876, 0.946] & 0.701 [0.602, 0.770] & 0.656 [0.546, 0.744] & 0.751 [0.597, 0.863] & 0.706 [0.609, 0.776] & 0.665 [0.552, 0.756] & 0.753 [0.603, 0.859] \\
 &  & 20 & 0.710 [0.628, 0.774] & 0.897 [0.863, 0.930] & 0.666 [0.564, 0.734] & 0.724 [0.651, 0.775] & 0.617 [0.471, 0.736] & 0.673 [0.583, 0.736] & 0.654 [0.574, 0.722] & 0.694 [0.547, 0.813] \\
 & CNN 3$\times$3 & 0 & 0.782 [0.716, 0.838] & 0.941 [0.920, 0.957] & 0.746 [0.687, 0.789] & 0.691 [0.618, 0.749] & 0.812 [0.733, 0.873] & 0.754 [0.702, 0.792] & 0.659 [0.586, 0.718] & 0.881 [0.826, 0.924] \\
 &  & 10 & 0.695 [0.579, 0.798] & 0.901 [0.861, 0.936] & 0.680 [0.586, 0.748] & 0.599 [0.500, 0.687] & 0.786 [0.625, 0.895] & 0.683 [0.576, 0.752] & 0.691 [0.589, 0.778] & 0.675 [0.509, 0.788] \\
 &  & 20 & 0.648 [0.545, 0.744] & 0.876 [0.830, 0.920] & 0.628 [0.531, 0.710] & 0.525 [0.417, 0.633] & 0.782 [0.633, 0.891] & 0.626 [0.525, 0.705] & 0.574 [0.456, 0.691] & 0.689 [0.539, 0.803] \\
 & CNN 5$\times$5 & 0 & 0.749 [0.682, 0.802] & 0.930 [0.909, 0.950] & 0.716 [0.647, 0.763] & 0.670 [0.599, 0.727] & 0.768 [0.650, 0.855] & 0.742 [0.689, 0.782] & 0.658 [0.588, 0.716] & 0.851 [0.771, 0.909] \\
 &  & 10 & 0.652 [0.541, 0.744] & 0.876 [0.836, 0.911] & 0.630 [0.529, 0.713] & 0.536 [0.427, 0.651] & 0.765 [0.597, 0.881] & 0.630 [0.521, 0.700] & 0.632 [0.517, 0.729] & 0.627 [0.463, 0.747] \\
 &  & 20 & 0.634 [0.526, 0.725] & 0.881 [0.844, 0.915] & 0.610 [0.519, 0.679] & 0.565 [0.485, 0.643] & 0.663 [0.522, 0.778] & 0.638 [0.569, 0.699] & 0.540 [0.461, 0.620] & 0.779 [0.689, 0.847] \\
 & CNN 7$\times$7 & 0 & 0.709 [0.631, 0.776] & 0.908 [0.873, 0.937] & 0.720 [0.647, 0.775] & 0.688 [0.607, 0.752] & 0.755 [0.651, 0.835] & 0.741 [0.674, 0.787] & 0.672 [0.591, 0.736] & 0.825 [0.761, 0.877] \\
 &  & 10 & 0.542 [0.448, 0.666] & 0.857 [0.809, 0.902] & 0.614 [0.518, 0.698] & 0.522 [0.414, 0.650] & 0.744 [0.594, 0.851] & 0.619 [0.524, 0.703] & 0.526 [0.417, 0.657] & 0.751 [0.608, 0.852] \\
 &  & 20 & 0.612 [0.520, 0.706] & 0.880 [0.841, 0.916] & 0.624 [0.531, 0.701] & 0.524 [0.415, 0.632] & 0.773 [0.650, 0.862] & 0.626 [0.534, 0.701] & 0.503 [0.396, 0.609] & 0.829 [0.734, 0.897] \\
\midrule
Coordinate-only (x, y) & XGBoost & 0 & 0.347 [0.222, 0.450] & 0.637 [0.572, 0.688] & 0.353 [0.261, 0.424] & 0.248 [0.179, 0.312] & 0.613 [0.431, 0.766] & 0.397 [0.302, 0.464] & 0.295 [0.222, 0.359] & 0.606 [0.428, 0.748] \\
 &  & 10 & 0.233 [0.162, 0.293] & 0.529 [0.446, 0.601] & 0.298 [0.199, 0.365] & 0.233 [0.165, 0.290] & 0.412 [0.219, 0.591] & 0.348 [0.265, 0.424] & 0.211 [0.153, 0.270] & 0.986 [0.958, 1.000] \\
 &  & 20 & 0.234 [0.170, 0.302] & 0.534 [0.452, 0.612] & 0.207 [0.079, 0.296] & 0.217 [0.125, 0.286] & 0.198 [0.054, 0.364] & 0.359 [0.272, 0.438] & 0.228 [0.165, 0.292] & 0.843 [0.685, 0.965] \\
 & CNN 3$\times$3 & 0 & 0.279 [0.175, 0.416] & 0.603 [0.515, 0.684] & 0.373 [0.267, 0.457] & 0.265 [0.184, 0.349] & 0.624 [0.413, 0.800] & 0.378 [0.275, 0.461] & 0.268 [0.188, 0.350] & 0.639 [0.421, 0.818] \\
 &  & 10 & 0.319 [0.208, 0.407] & 0.584 [0.493, 0.668] & 0.276 [0.170, 0.351] & 0.225 [0.143, 0.317] & 0.358 [0.173, 0.553] & 0.378 [0.294, 0.452] & 0.240 [0.178, 0.301] & 0.885 [0.743, 0.972] \\
 &  & 20 & 0.239 [0.180, 0.293] & 0.533 [0.449, 0.618] & 0.303 [0.225, 0.360] & 0.220 [0.164, 0.277] & 0.488 [0.296, 0.677] & 0.376 [0.289, 0.450] & 0.234 [0.171, 0.294] & 0.952 [0.864, 1.000] \\
 & CNN 5$\times$5 & 0 & 0.319 [0.206, 0.472] & 0.650 [0.558, 0.725] & 0.395 [0.294, 0.480] & 0.269 [0.190, 0.346] & 0.748 [0.574, 0.876] & 0.413 [0.297, 0.501] & 0.305 [0.209, 0.398] & 0.640 [0.440, 0.778] \\
 &  & 10 & 0.223 [0.164, 0.293] & 0.528 [0.430, 0.620] & 0.241 [0.142, 0.308] & 0.236 [0.168, 0.309] & 0.247 [0.115, 0.401] & 0.366 [0.279, 0.444] & 0.226 [0.164, 0.289] & 0.956 [0.876, 1.000] \\
 &  & 20 & 0.208 [0.153, 0.269] & 0.508 [0.416, 0.599] & 0.247 [0.162, 0.312] & 0.189 [0.123, 0.258] & 0.356 [0.195, 0.538] & 0.376 [0.289, 0.450] & 0.234 [0.171, 0.295] & 0.947 [0.860, 0.997] \\
 & CNN 7$\times$7 & 0 & 0.222 [0.143, 0.367] & 0.566 [0.454, 0.669] & 0.341 [0.227, 0.452] & 0.257 [0.157, 0.384] & 0.504 [0.332, 0.691] & 0.383 [0.276, 0.482] & 0.268 [0.178, 0.369] & 0.672 [0.521, 0.804] \\
 &  & 10 & 0.244 [0.183, 0.343] & 0.546 [0.464, 0.622] & 0.236 [0.132, 0.302] & 0.258 [0.194, 0.351] & 0.217 [0.094, 0.367] & 0.364 [0.276, 0.443] & 0.225 [0.162, 0.289] & 0.949 [0.874, 0.993] \\
 &  & 20 & 0.222 [0.161, 0.308] & 0.535 [0.445, 0.629] & 0.205 [0.106, 0.281] & 0.245 [0.138, 0.391] & 0.175 [0.075, 0.299] & 0.377 [0.290, 0.452] & 0.235 [0.171, 0.296] & 0.953 [0.870, 1.000] \\
\end{longtable}
\endgroup

% ============================================================================
% Supplementary table - Feature-set contrasts vs baseline and coordinate control
% label: tab:sup_feature_set_contrasts
% ============================================================================
\begin{table}[H]
\centering
\caption{Paired feature-set contrasts [95\% CI] against the \textit{baseline} predictor set and the coordinate-only control (XGBoost), by training-exclusion buffer.}
\label{tab:sup_feature_set_contrasts}
\scriptsize
\setlength{\tabcolsep}{6pt}
\renewcommand{\arraystretch}{0.92}
\begin{threeparttable}
\begin{tabular}{lrcccccc}
\toprule
 &  & \multicolumn{2}{c}{Threshold-free} & \multicolumn{3}{c}{Inner-fold threshold} & \\
\cmidrule(lr){3-4}\cmidrule(lr){5-7}
Contrast & Buffer & PR-AUC & ROC-AUC & $F_1$ & Precision & Recall & $p$ \\
 & (km) &  &  &  &  &  & \\
\midrule
+ Conventional EO $-$ Baseline & 0 & \textbf{\phantom{$-$}0.217 [\phantom{$-$}0.152, \phantom{$-$}0.290]} & \phantom{$-$}0.097 [\phantom{$-$}0.075, \phantom{$-$}0.125] & \phantom{$-$}0.168 [\phantom{$-$}0.128, \phantom{$-$}0.224] & \phantom{$-$}0.182 [\phantom{$-$}0.127, \phantom{$-$}0.254] & \phantom{$-$}0.148 [\phantom{$-$}0.102, \phantom{$-$}0.208] & \textbf{0.0004} \\
 & 10 & \textbf{\phantom{$-$}0.208 [\phantom{$-$}0.152, \phantom{$-$}0.261]} & \phantom{$-$}0.103 [\phantom{$-$}0.078, \phantom{$-$}0.128] & \phantom{$-$}0.143 [\phantom{$-$}0.108, \phantom{$-$}0.182] & \phantom{$-$}0.159 [\phantom{$-$}0.115, \phantom{$-$}0.206] & \phantom{$-$}0.111 [\phantom{$-$}0.045, \phantom{$-$}0.177] & \textbf{0.0004} \\
 & 20 & \textbf{\phantom{$-$}0.149 [\phantom{$-$}0.086, \phantom{$-$}0.222]} & \phantom{$-$}0.116 [\phantom{$-$}0.081, \phantom{$-$}0.158] & \phantom{$-$}0.118 [\phantom{$-$}0.071, \phantom{$-$}0.176] & \phantom{$-$}0.171 [\phantom{$-$}0.120, \phantom{$-$}0.231] & \phantom{$-$}0.068 [\phantom{$-$}0.004, \phantom{$-$}0.145] & \textbf{0.0004} \\
\midrule
+ AlphaEarth $-$ Baseline & 0 & \textbf{\phantom{$-$}0.217 [\phantom{$-$}0.150, \phantom{$-$}0.289]} & \phantom{$-$}0.105 [\phantom{$-$}0.082, \phantom{$-$}0.136] & \phantom{$-$}0.188 [\phantom{$-$}0.148, \phantom{$-$}0.247] & \phantom{$-$}0.200 [\phantom{$-$}0.142, \phantom{$-$}0.278] & \phantom{$-$}0.169 [\phantom{$-$}0.115, \phantom{$-$}0.244] & \textbf{0.0004} \\
 & 10 & \textbf{\phantom{$-$}0.226 [\phantom{$-$}0.170, \phantom{$-$}0.283]} & \phantom{$-$}0.116 [\phantom{$-$}0.090, \phantom{$-$}0.143] & \phantom{$-$}0.175 [\phantom{$-$}0.141, \phantom{$-$}0.218] & \phantom{$-$}0.186 [\phantom{$-$}0.139, \phantom{$-$}0.243] & \phantom{$-$}0.150 [\phantom{$-$}0.083, \phantom{$-$}0.226] & \textbf{0.0004} \\
 & 20 & \textbf{\phantom{$-$}0.188 [\phantom{$-$}0.111, \phantom{$-$}0.273]} & \phantom{$-$}0.137 [\phantom{$-$}0.099, \phantom{$-$}0.183] & \phantom{$-$}0.156 [\phantom{$-$}0.090, \phantom{$-$}0.234] & \phantom{$-$}0.210 [\phantom{$-$}0.150, \phantom{$-$}0.283] & \phantom{$-$}0.105 [\phantom{$-$}0.004, \phantom{$-$}0.212] & \textbf{0.0004} \\
\midrule
+ TESSERA $-$ Baseline & 0 & \textbf{\phantom{$-$}0.292 [\phantom{$-$}0.222, \phantom{$-$}0.375]} & \phantom{$-$}0.126 [\phantom{$-$}0.101, \phantom{$-$}0.159] & \phantom{$-$}0.220 [\phantom{$-$}0.172, \phantom{$-$}0.291] & \phantom{$-$}0.241 [\phantom{$-$}0.172, \phantom{$-$}0.337] & \phantom{$-$}0.190 [\phantom{$-$}0.133, \phantom{$-$}0.260] & \textbf{0.0004} \\
 & 10 & \textbf{\phantom{$-$}0.252 [\phantom{$-$}0.177, \phantom{$-$}0.340]} & \phantom{$-$}0.125 [\phantom{$-$}0.095, \phantom{$-$}0.158] & \phantom{$-$}0.198 [\phantom{$-$}0.153, \phantom{$-$}0.250] & \phantom{$-$}0.224 [\phantom{$-$}0.144, \phantom{$-$}0.310] & \phantom{$-$}0.151 [\phantom{$-$}0.068, \phantom{$-$}0.232] & \textbf{0.0004} \\
 & 20 & \textbf{\phantom{$-$}0.242 [\phantom{$-$}0.169, \phantom{$-$}0.319]} & \phantom{$-$}0.146 [\phantom{$-$}0.107, \phantom{$-$}0.192] & \phantom{$-$}0.199 [\phantom{$-$}0.138, \phantom{$-$}0.269] & \phantom{$-$}0.273 [\phantom{$-$}0.204, \phantom{$-$}0.345] & \phantom{$-$}0.133 [\phantom{$-$}0.048, \phantom{$-$}0.222] & \textbf{0.0004} \\
\midrule
Baseline $-$ Coordinate-only & 0 & \textbf{\phantom{$-$}0.202 [\phantom{$-$}0.094, \phantom{$-$}0.313]} & \phantom{$-$}0.190 [\phantom{$-$}0.123, \phantom{$-$}0.265] & \phantom{$-$}0.198 [\phantom{$-$}0.092, \phantom{$-$}0.291] & \phantom{$-$}0.254 [\phantom{$-$}0.173, \phantom{$-$}0.326] & $-$0.004 [$-$0.229, \phantom{$-$}0.220] & \textbf{0.0008} \\
 & 10 & \textbf{\phantom{$-$}0.242 [\phantom{$-$}0.168, \phantom{$-$}0.325]} & \phantom{$-$}0.259 [\phantom{$-$}0.180, \phantom{$-$}0.348] & \phantom{$-$}0.205 [\phantom{$-$}0.132, \phantom{$-$}0.302] & \phantom{$-$}0.199 [\phantom{$-$}0.124, \phantom{$-$}0.293] & \phantom{$-$}0.188 [\phantom{$-$}0.021, \phantom{$-$}0.376] & \textbf{0.0004} \\
 & 20 & \textbf{\phantom{$-$}0.235 [\phantom{$-$}0.131, \phantom{$-$}0.326]} & \phantom{$-$}0.217 [\phantom{$-$}0.115, \phantom{$-$}0.327] & \phantom{$-$}0.260 [\phantom{$-$}0.151, \phantom{$-$}0.402] & \phantom{$-$}0.234 [\phantom{$-$}0.139, \phantom{$-$}0.367] & \phantom{$-$}0.286 [\phantom{$-$}0.141, \phantom{$-$}0.439] & \textbf{0.0004} \\
\bottomrule
\end{tabular}
\end{threeparttable}%
\end{table}

% ============================================================================
% Supplementary table - EO-representation contrasts
% label: tab:sup_embedding_contrasts
% ============================================================================
\begin{table}[H]
\centering
\caption{Paired contrasts [95\% CI] among the Earth observation feature sets (XGBoost), by training-exclusion buffer.}
\label{tab:sup_embedding_contrasts}
\scriptsize
\setlength{\tabcolsep}{6pt}
\renewcommand{\arraystretch}{0.92}
\begin{threeparttable}
\begin{tabular}{lrcccccc}
\toprule
 &  & \multicolumn{2}{c}{Threshold-free} & \multicolumn{3}{c}{Inner-fold threshold} & \\
\cmidrule(lr){3-4}\cmidrule(lr){5-7}
Contrast & Buffer & PR-AUC & ROC-AUC & $F_1$ & Precision & Recall & $p$ \\
 & (km) &  &  &  &  &  & \\
\midrule
+ AlphaEarth $-$ + Conventional EO & 0 & $-$0.000 [$-$0.046, \phantom{$-$}0.047] & \phantom{$-$}0.008 [$-$0.005, \phantom{$-$}0.020] & \phantom{$-$}0.020 [$-$0.011, \phantom{$-$}0.054] & \phantom{$-$}0.018 [$-$0.019, \phantom{$-$}0.055] & \phantom{$-$}0.022 [$-$0.030, \phantom{$-$}0.087] & 0.936 \\
 & 10 & \phantom{$-$}0.018 [$-$0.026, \phantom{$-$}0.064] & \phantom{$-$}0.013 [$-$0.003, \phantom{$-$}0.032] & \phantom{$-$}0.032 [$-$0.001, \phantom{$-$}0.071] & \phantom{$-$}0.027 [$-$0.011, \phantom{$-$}0.075] & \phantom{$-$}0.039 [$-$0.002, \phantom{$-$}0.080] & 0.364 \\
 & 20 & \textbf{\phantom{$-$}0.039 [\phantom{$-$}0.001, \phantom{$-$}0.078]} & \phantom{$-$}0.022 [\phantom{$-$}0.004, \phantom{$-$}0.040] & \phantom{$-$}0.038 [$-$0.015, \phantom{$-$}0.092] & \phantom{$-$}0.040 [$-$0.001, \phantom{$-$}0.076] & \phantom{$-$}0.037 [$-$0.040, \phantom{$-$}0.114] & \textbf{0.047} \\
\midrule
+ TESSERA $-$ + Conventional EO & 0 & \textbf{\phantom{$-$}0.075 [\phantom{$-$}0.046, \phantom{$-$}0.109]} & \phantom{$-$}0.029 [\phantom{$-$}0.018, \phantom{$-$}0.041] & \phantom{$-$}0.052 [\phantom{$-$}0.025, \phantom{$-$}0.089] & \phantom{$-$}0.060 [\phantom{$-$}0.032, \phantom{$-$}0.095] & \phantom{$-$}0.043 [\phantom{$-$}0.002, \phantom{$-$}0.091] & \textbf{0.0004} \\
 & 10 & \phantom{$-$}0.043 [$-$0.011, \phantom{$-$}0.106] & \phantom{$-$}0.023 [\phantom{$-$}0.010, \phantom{$-$}0.036] & \phantom{$-$}0.055 [\phantom{$-$}0.026, \phantom{$-$}0.090] & \phantom{$-$}0.065 [\phantom{$-$}0.018, \phantom{$-$}0.124] & \phantom{$-$}0.040 [$-$0.001, \phantom{$-$}0.093] & 0.127 \\
 & 20 & \textbf{\phantom{$-$}0.092 [\phantom{$-$}0.054, \phantom{$-$}0.123]} & \phantom{$-$}0.030 [\phantom{$-$}0.015, \phantom{$-$}0.045] & \phantom{$-$}0.081 [\phantom{$-$}0.042, \phantom{$-$}0.117] & \phantom{$-$}0.102 [\phantom{$-$}0.046, \phantom{$-$}0.149] & \phantom{$-$}0.064 [\phantom{$-$}0.010, \phantom{$-$}0.116] & \textbf{0.0004} \\
\midrule
+ TESSERA $-$ + AlphaEarth & 0 & \textbf{\phantom{$-$}0.075 [\phantom{$-$}0.036, \phantom{$-$}0.122]} & \phantom{$-$}0.021 [\phantom{$-$}0.012, \phantom{$-$}0.032] & \phantom{$-$}0.032 [\phantom{$-$}0.011, \phantom{$-$}0.063] & \phantom{$-$}0.041 [$-$0.001, \phantom{$-$}0.096] & \phantom{$-$}0.021 [$-$0.010, \phantom{$-$}0.053] & \textbf{0.0004} \\
 & 10 & \phantom{$-$}0.026 [$-$0.041, \phantom{$-$}0.099] & \phantom{$-$}0.009 [$-$0.011, \phantom{$-$}0.029] & \phantom{$-$}0.023 [$-$0.019, \phantom{$-$}0.064] & \phantom{$-$}0.038 [$-$0.027, \phantom{$-$}0.112] & \phantom{$-$}0.001 [$-$0.047, \phantom{$-$}0.048] & 0.452 \\
 & 20 & \textbf{\phantom{$-$}0.053 [\phantom{$-$}0.002, \phantom{$-$}0.097]} & \phantom{$-$}0.009 [$-$0.011, \phantom{$-$}0.029] & \phantom{$-$}0.043 [$-$0.001, \phantom{$-$}0.083] & \phantom{$-$}0.062 [\phantom{$-$}0.005, \phantom{$-$}0.114] & \phantom{$-$}0.028 [$-$0.027, \phantom{$-$}0.079] & \textbf{0.046} \\
\bottomrule
\end{tabular}
\end{threeparttable}%
\end{table}

% ============================================================================
% Supplementary table - Architecture contrasts (CNN vs XGBoost)
% label: tab:sup_architecture_contrasts
% ============================================================================
\begin{table}[H]
\centering
\caption{Paired architecture contrasts [95\% CI]: each CNN receptive field versus XGBoost, by feature set and training-exclusion buffer.}
\label{tab:sup_architecture_contrasts}
\scriptsize
\setlength{\tabcolsep}{5pt}
\renewcommand{\arraystretch}{0.92}
\begin{threeparttable}
\begin{tabular}{llrcccccc}
\toprule
 &  &  & \multicolumn{2}{c}{Threshold-free} & \multicolumn{3}{c}{Inner-fold threshold} & \\
\cmidrule(lr){4-5}\cmidrule(lr){6-8}
Feature set & Architecture & Buffer & PR-AUC & ROC-AUC & $F_1$ & Precision & Recall & $p$ \\
 &  & (km) &  &  &  &  &  & \\
\midrule
Baseline & CNN 3$\times$3 & 0 & $-$0.022 [$-$0.058, \phantom{$-$}0.017] & $-$0.002 [$-$0.018, \phantom{$-$}0.012] & \phantom{$-$}0.018 [$-$0.010, \phantom{$-$}0.059] & $-$0.003 [$-$0.035, \phantom{$-$}0.036] & \phantom{$-$}0.052 [\phantom{$-$}0.012, \phantom{$-$}0.103] & 0.232 \\
 &  & 10 & \textbf{$-$0.073 [$-$0.112, $-$0.026]} & $-$0.016 [$-$0.043, \phantom{$-$}0.007] & $-$0.005 [$-$0.034, \phantom{$-$}0.030] & $-$0.023 [$-$0.055, \phantom{$-$}0.011] & \phantom{$-$}0.036 [$-$0.010, \phantom{$-$}0.080] & \textbf{0.0064} \\
 &  & 20 & \textbf{$-$0.120 [$-$0.176, $-$0.050]} & $-$0.023 [$-$0.056, \phantom{$-$}0.006] & $-$0.013 [$-$0.055, \phantom{$-$}0.037] & $-$0.062 [$-$0.114, $-$0.004] & \phantom{$-$}0.062 [$-$0.025, \phantom{$-$}0.160] & \textbf{0.0012} \\
 & CNN 5$\times$5 & 0 & \phantom{$-$}0.018 [$-$0.019, \phantom{$-$}0.047] & $-$0.008 [$-$0.026, \phantom{$-$}0.007] & \phantom{$-$}0.004 [$-$0.019, \phantom{$-$}0.035] & \phantom{$-$}0.036 [\phantom{$-$}0.012, \phantom{$-$}0.068] & $-$0.039 [$-$0.084, \phantom{$-$}0.020] & 0.419 \\
 &  & 10 & $-$0.010 [$-$0.061, \phantom{$-$}0.030] & $-$0.010 [$-$0.040, \phantom{$-$}0.016] & $-$0.012 [$-$0.045, \phantom{$-$}0.023] & $-$0.005 [$-$0.049, \phantom{$-$}0.038] & $-$0.026 [$-$0.076, \phantom{$-$}0.028] & 0.516 \\
 &  & 20 & \textbf{$-$0.129 [$-$0.189, $-$0.044]} & $-$0.048 [$-$0.091, $-$0.004] & $-$0.034 [$-$0.087, \phantom{$-$}0.026] & $-$0.081 [$-$0.146, $-$0.012] & \phantom{$-$}0.037 [$-$0.065, \phantom{$-$}0.145] & \textbf{0.0024} \\
 & CNN 7$\times$7 & 0 & $-$0.017 [$-$0.054, \phantom{$-$}0.024] & $-$0.027 [$-$0.053, $-$0.003] & $-$0.004 [$-$0.031, \phantom{$-$}0.032] & \phantom{$-$}0.023 [$-$0.003, \phantom{$-$}0.052] & $-$0.041 [$-$0.085, \phantom{$-$}0.020] & 0.348 \\
 &  & 10 & $-$0.032 [$-$0.083, \phantom{$-$}0.026] & $-$0.020 [$-$0.046, \phantom{$-$}0.006] & $-$0.007 [$-$0.031, \phantom{$-$}0.023] & $-$0.017 [$-$0.049, \phantom{$-$}0.013] & \phantom{$-$}0.013 [$-$0.026, \phantom{$-$}0.058] & 0.231 \\
 &  & 20 & $-$0.052 [$-$0.106, \phantom{$-$}0.007] & $-$0.001 [$-$0.033, \phantom{$-$}0.034] & \phantom{$-$}0.021 [$-$0.026, \phantom{$-$}0.081] & $-$0.033 [$-$0.081, \phantom{$-$}0.015] & \phantom{$-$}0.101 [$-$0.008, \phantom{$-$}0.215] & 0.093 \\
\midrule
+ Conventional EO & CNN 3$\times$3 & 0 & \textbf{$-$0.058 [$-$0.085, $-$0.027]} & $-$0.008 [$-$0.016, $-$0.001] & $-$0.009 [$-$0.029, \phantom{$-$}0.014] & $-$0.054 [$-$0.080, $-$0.032] & \phantom{$-$}0.056 [\phantom{$-$}0.026, \phantom{$-$}0.098] & \textbf{0.0016} \\
 &  & 10 & \textbf{$-$0.081 [$-$0.125, $-$0.028]} & $-$0.011 [$-$0.030, \phantom{$-$}0.006] & \phantom{$-$}0.005 [$-$0.030, \phantom{$-$}0.052] & $-$0.038 [$-$0.063, $-$0.006] & \phantom{$-$}0.080 [\phantom{$-$}0.035, \phantom{$-$}0.135] & \textbf{0.0052} \\
 &  & 20 & \textbf{$-$0.123 [$-$0.187, $-$0.033]} & $-$0.024 [$-$0.059, \phantom{$-$}0.010] & \phantom{$-$}0.007 [$-$0.049, \phantom{$-$}0.067] & $-$0.111 [$-$0.183, $-$0.041] & \phantom{$-$}0.153 [\phantom{$-$}0.090, \phantom{$-$}0.212] & \textbf{0.0092} \\
 & CNN 5$\times$5 & 0 & \textbf{$-$0.048 [$-$0.089, $-$0.011]} & $-$0.030 [$-$0.063, $-$0.005] & $-$0.026 [$-$0.050, $-$0.005] & $-$0.025 [$-$0.057, \phantom{$-$}0.003] & $-$0.027 [$-$0.060, \phantom{$-$}0.010] & \textbf{0.0076} \\
 &  & 10 & $-$0.041 [$-$0.096, \phantom{$-$}0.013] & $-$0.004 [$-$0.034, \phantom{$-$}0.022] & \phantom{$-$}0.012 [$-$0.028, \phantom{$-$}0.063] & \phantom{$-$}0.016 [$-$0.011, \phantom{$-$}0.043] & \phantom{$-$}0.006 [$-$0.074, \phantom{$-$}0.098] & 0.136 \\
 &  & 20 & \textbf{$-$0.150 [$-$0.232, $-$0.022]} & $-$0.034 [$-$0.081, \phantom{$-$}0.011] & \phantom{$-$}0.026 [$-$0.049, \phantom{$-$}0.111] & $-$0.107 [$-$0.193, $-$0.015] & \phantom{$-$}0.200 [\phantom{$-$}0.099, \phantom{$-$}0.299] & \textbf{0.018} \\
 & CNN 7$\times$7 & 0 & \textbf{$-$0.112 [$-$0.166, $-$0.067]} & $-$0.054 [$-$0.087, $-$0.023] & $-$0.041 [$-$0.079, $-$0.012] & $-$0.037 [$-$0.073, $-$0.008] & $-$0.046 [$-$0.097, $-$0.001] & \textbf{0.0004} \\
 &  & 10 & \textbf{$-$0.098 [$-$0.146, $-$0.041]} & $-$0.045 [$-$0.078, $-$0.013] & $-$0.008 [$-$0.043, \phantom{$-$}0.029] & $-$0.004 [$-$0.033, \phantom{$-$}0.024] & $-$0.013 [$-$0.080, \phantom{$-$}0.058] & \textbf{0.0020} \\
 &  & 20 & \textbf{$-$0.115 [$-$0.159, $-$0.056]} & $-$0.065 [$-$0.110, $-$0.021] & $-$0.019 [$-$0.059, \phantom{$-$}0.026] & $-$0.111 [$-$0.177, $-$0.046] & \phantom{$-$}0.084 [\phantom{$-$}0.006, \phantom{$-$}0.161] & \textbf{0.0008} \\
\midrule
+ AlphaEarth & CNN 3$\times$3 & 0 & \textbf{$-$0.067 [$-$0.097, $-$0.029]} & $-$0.016 [$-$0.030, $-$0.004] & $-$0.033 [$-$0.060, $-$0.008] & $-$0.066 [$-$0.090, $-$0.046] & \phantom{$-$}0.013 [$-$0.037, \phantom{$-$}0.054] & \textbf{0.0008} \\
 &  & 10 & \textbf{$-$0.076 [$-$0.108, $-$0.035]} & $-$0.022 [$-$0.038, $-$0.004] & $-$0.046 [$-$0.088, \phantom{$-$}0.002] & $-$0.088 [$-$0.111, $-$0.054] & \phantom{$-$}0.032 [$-$0.043, \phantom{$-$}0.093] & \textbf{0.0004} \\
 &  & 20 & $-$0.054 [$-$0.112, \phantom{$-$}0.008] & $-$0.020 [$-$0.044, \phantom{$-$}0.007] & \phantom{$-$}0.006 [$-$0.064, \phantom{$-$}0.093] & $-$0.147 [$-$0.196, $-$0.085] & \phantom{$-$}0.222 [\phantom{$-$}0.120, \phantom{$-$}0.326] & 0.094 \\
 & CNN 5$\times$5 & 0 & \textbf{$-$0.094 [$-$0.128, $-$0.045]} & $-$0.023 [$-$0.037, $-$0.011] & $-$0.035 [$-$0.060, $-$0.011] & $-$0.068 [$-$0.104, $-$0.043] & \phantom{$-$}0.011 [$-$0.017, \phantom{$-$}0.044] & \textbf{0.0008} \\
 &  & 10 & \textbf{$-$0.119 [$-$0.154, $-$0.076]} & $-$0.045 [$-$0.068, $-$0.022] & $-$0.065 [$-$0.102, $-$0.020] & $-$0.113 [$-$0.140, $-$0.079] & \phantom{$-$}0.028 [$-$0.025, \phantom{$-$}0.083] & \textbf{0.0004} \\
 &  & 20 & \textbf{$-$0.089 [$-$0.155, $-$0.011]} & $-$0.032 [$-$0.070, $-$0.000] & $-$0.024 [$-$0.090, \phantom{$-$}0.056] & $-$0.190 [$-$0.256, $-$0.120] & \phantom{$-$}0.235 [\phantom{$-$}0.150, \phantom{$-$}0.327] & \textbf{0.027} \\
 & CNN 7$\times$7 & 0 & \textbf{$-$0.125 [$-$0.167, $-$0.087]} & $-$0.055 [$-$0.095, $-$0.028] & $-$0.064 [$-$0.103, $-$0.032] & $-$0.029 [$-$0.072, \phantom{$-$}0.005] & $-$0.103 [$-$0.157, $-$0.053] & \textbf{0.0004} \\
 &  & 10 & \textbf{$-$0.173 [$-$0.234, $-$0.107]} & $-$0.070 [$-$0.113, $-$0.032] & $-$0.085 [$-$0.131, $-$0.040] & $-$0.090 [$-$0.135, $-$0.036] & $-$0.075 [$-$0.142, $-$0.019] & \textbf{0.0004} \\
 &  & 20 & \textbf{$-$0.135 [$-$0.224, $-$0.048]} & $-$0.048 [$-$0.093, $-$0.012] & $-$0.030 [$-$0.078, \phantom{$-$}0.014] & $-$0.153 [$-$0.233, $-$0.076] & \phantom{$-$}0.123 [\phantom{$-$}0.057, \phantom{$-$}0.193] & \textbf{0.0004} \\
\midrule
+ TESSERA & CNN 3$\times$3 & 0 & \textbf{$-$0.059 [$-$0.088, $-$0.030]} & $-$0.013 [$-$0.023, $-$0.005] & $-$0.024 [$-$0.059, \phantom{$-$}0.006] & $-$0.053 [$-$0.105, $-$0.012] & \phantom{$-$}0.012 [$-$0.024, \phantom{$-$}0.051] & \textbf{0.0008} \\
 &  & 10 & \textbf{$-$0.032 [$-$0.062, $-$0.005]} & $-$0.012 [$-$0.025, $-$0.001] & $-$0.020 [$-$0.049, \phantom{$-$}0.007] & $-$0.057 [$-$0.102, $-$0.017] & \phantom{$-$}0.035 [$-$0.001, \phantom{$-$}0.067] & \textbf{0.025} \\
 &  & 20 & \textbf{$-$0.062 [$-$0.109, $-$0.008]} & $-$0.021 [$-$0.043, \phantom{$-$}0.002] & $-$0.038 [$-$0.100, \phantom{$-$}0.025] & $-$0.199 [$-$0.261, $-$0.116] & \phantom{$-$}0.165 [\phantom{$-$}0.105, \phantom{$-$}0.224] & \textbf{0.024} \\
 & CNN 5$\times$5 & 0 & \textbf{$-$0.091 [$-$0.129, $-$0.061]} & $-$0.024 [$-$0.036, $-$0.013] & $-$0.054 [$-$0.082, $-$0.031] & $-$0.073 [$-$0.132, $-$0.023] & $-$0.032 [$-$0.071, \phantom{$-$}0.008] & \textbf{0.0004} \\
 &  & 10 & \textbf{$-$0.076 [$-$0.118, $-$0.043]} & $-$0.038 [$-$0.055, $-$0.021] & $-$0.070 [$-$0.120, $-$0.022] & $-$0.121 [$-$0.188, $-$0.053] & \phantom{$-$}0.014 [$-$0.044, \phantom{$-$}0.060] & \textbf{0.0004} \\
 &  & 20 & \textbf{$-$0.076 [$-$0.143, $-$0.004]} & $-$0.016 [$-$0.047, \phantom{$-$}0.016] & $-$0.056 [$-$0.123, \phantom{$-$}0.034] & $-$0.159 [$-$0.206, $-$0.091] & \phantom{$-$}0.046 [$-$0.063, \phantom{$-$}0.165] & \textbf{0.039} \\
 & CNN 7$\times$7 & 0 & \textbf{$-$0.132 [$-$0.182, $-$0.085]} & $-$0.046 [$-$0.077, $-$0.024] & $-$0.050 [$-$0.098, $-$0.017] & $-$0.055 [$-$0.117, $-$0.007] & $-$0.045 [$-$0.103, \phantom{$-$}0.008] & \textbf{0.0004} \\
 &  & 10 & \textbf{$-$0.186 [$-$0.285, $-$0.094]} & $-$0.057 [$-$0.099, $-$0.023] & $-$0.087 [$-$0.153, $-$0.023] & $-$0.134 [$-$0.220, $-$0.045] & $-$0.007 [$-$0.071, \phantom{$-$}0.045] & \textbf{0.0004} \\
 &  & 20 & \textbf{$-$0.098 [$-$0.148, $-$0.035]} & $-$0.017 [$-$0.046, \phantom{$-$}0.011] & $-$0.042 [$-$0.116, \phantom{$-$}0.036] & $-$0.200 [$-$0.271, $-$0.113] & \phantom{$-$}0.156 [\phantom{$-$}0.080, \phantom{$-$}0.231] & \textbf{0.0060} \\
\bottomrule
\end{tabular}
\end{threeparttable}
\end{table}
\end{landscape}

% ============================================================================
% Supplementary table - Per-fold parcel PR-AUC
% label: tab:sup_per_fold_pr_auc
% ============================================================================
\begin{table}[H]
\centering
\caption{Per-fold out-of-fold parcel PR-AUC for every feature set, architecture and training-exclusion buffer, including the coordinate-only control.}
\label{tab:sup_per_fold_pr_auc}
\footnotesize
\setlength{\tabcolsep}{4pt}
\resizebox{\textwidth}{!}{%
\begin{threeparttable}
\begin{tabular}{llrrrrrrrrrr}
\toprule
 & & & \multicolumn{6}{c}{Outer fold} & & & \\
\cmidrule(lr){4-9}
Feature set & Architecture & Buffer & 1 & 2 & 3 & 4 & 5 & 6 & Mean & SD & Range \\
& & (km) & & & & & & & & & \\
\midrule
Baseline & XGBoost & 0 & 0.586 & 0.561 & 0.508 & 0.656 & 0.504 & 0.654 & 0.578 & 0.067 & 0.152 \\
 &  & 10 & 0.595 & 0.562 & 0.443 & 0.550 & 0.608 & 0.657 & 0.569 & 0.072 & 0.214 \\
 &  & 20 & 0.625 & 0.591 & 0.348 & 0.615 & 0.597 & 0.619 & 0.566 & 0.107 & 0.277 \\
 & CNN 3$\times$3 & 0 & 0.587 & 0.534 & 0.419 & 0.726 & 0.445 & 0.667 & 0.563 & 0.121 & 0.307 \\
 &  & 10 & 0.601 & 0.533 & 0.267 & 0.462 & 0.413 & 0.657 & 0.489 & 0.140 & 0.390 \\
 &  & 20 & 0.513 & 0.525 & 0.186 & 0.324 & 0.319 & 0.686 & 0.426 & 0.182 & 0.500 \\
 & CNN 5$\times$5 & 0 & 0.582 & 0.666 & 0.411 & 0.713 & 0.398 & 0.649 & 0.570 & 0.135 & 0.315 \\
 &  & 10 & 0.597 & 0.600 & 0.236 & 0.505 & 0.400 & 0.623 & 0.493 & 0.151 & 0.386 \\
 &  & 20 & 0.523 & 0.566 & 0.186 & 0.314 & 0.378 & 0.440 & 0.401 & 0.140 & 0.379 \\
 & CNN 7$\times$7 & 0 & 0.513 & 0.600 & 0.357 & 0.714 & 0.385 & 0.649 & 0.536 & 0.144 & 0.358 \\
 &  & 10 & 0.585 & 0.556 & 0.238 & 0.523 & 0.376 & 0.616 & 0.482 & 0.146 & 0.379 \\
 &  & 20 & 0.566 & 0.587 & 0.268 & 0.423 & 0.353 & 0.621 & 0.470 & 0.143 & 0.353 \\
\midrule
+ Conventional EO & XGBoost & 0 & 0.850 & 0.751 & 0.648 & 0.870 & 0.721 & 0.857 & 0.783 & 0.090 & 0.222 \\
 &  & 10 & 0.869 & 0.683 & 0.460 & 0.757 & 0.738 & 0.829 & 0.722 & 0.145 & 0.409 \\
 &  & 20 & 0.865 & 0.676 & 0.546 & 0.751 & 0.740 & 0.715 & 0.716 & 0.105 & 0.320 \\
 & CNN 3$\times$3 & 0 & 0.852 & 0.673 & 0.571 & 0.849 & 0.682 & 0.829 & 0.743 & 0.117 & 0.281 \\
 &  & 10 & 0.899 & 0.648 & 0.473 & 0.637 & 0.605 & 0.799 & 0.677 & 0.151 & 0.426 \\
 &  & 20 & 0.886 & 0.644 & 0.442 & 0.382 & 0.487 & 0.768 & 0.601 & 0.198 & 0.504 \\
 & CNN 5$\times$5 & 0 & 0.847 & 0.718 & 0.536 & 0.855 & 0.576 & 0.816 & 0.725 & 0.140 & 0.319 \\
 &  & 10 & 0.905 & 0.675 & 0.507 & 0.631 & 0.620 & 0.786 & 0.687 & 0.140 & 0.398 \\
 &  & 20 & 0.885 & 0.659 & 0.433 & 0.353 & 0.472 & 0.771 & 0.595 & 0.209 & 0.532 \\
 & CNN 7$\times$7 & 0 & 0.845 & 0.668 & 0.514 & 0.871 & 0.459 & 0.811 & 0.695 & 0.177 & 0.412 \\
 &  & 10 & 0.913 & 0.591 & 0.425 & 0.679 & 0.530 & 0.780 & 0.653 & 0.176 & 0.488 \\
 &  & 20 & 0.911 & 0.640 & 0.328 & 0.433 & 0.458 & 0.761 & 0.589 & 0.222 & 0.583 \\
\midrule
+ AlphaEarth & XGBoost & 0 & 0.905 & 0.747 & 0.746 & 0.891 & 0.755 & 0.849 & 0.816 & 0.075 & 0.159 \\
 &  & 10 & 0.900 & 0.692 & 0.645 & 0.785 & 0.808 & 0.821 & 0.775 & 0.092 & 0.255 \\
 &  & 20 & 0.894 & 0.710 & 0.642 & 0.739 & 0.731 & 0.760 & 0.746 & 0.083 & 0.252 \\
 & CNN 3$\times$3 & 0 & 0.895 & 0.668 & 0.692 & 0.864 & 0.706 & 0.718 & 0.757 & 0.096 & 0.226 \\
 &  & 10 & 0.936 & 0.628 & 0.634 & 0.716 & 0.659 & 0.730 & 0.717 & 0.115 & 0.308 \\
 &  & 20 & 0.933 & 0.623 & 0.616 & 0.681 & 0.661 & 0.765 & 0.713 & 0.120 & 0.317 \\
 & CNN 5$\times$5 & 0 & 0.806 & 0.633 & 0.682 & 0.821 & 0.683 & 0.814 & 0.740 & 0.083 & 0.189 \\
 &  & 10 & 0.826 & 0.628 & 0.589 & 0.599 & 0.662 & 0.810 & 0.686 & 0.106 & 0.237 \\
 &  & 20 & 0.818 & 0.606 & 0.592 & 0.511 & 0.641 & 0.813 & 0.663 & 0.125 & 0.307 \\
 & CNN 7$\times$7 & 0 & 0.831 & 0.649 & 0.491 & 0.784 & 0.603 & 0.715 & 0.679 & 0.124 & 0.340 \\
 &  & 10 & 0.876 & 0.616 & 0.423 & 0.530 & 0.508 & 0.704 & 0.609 & 0.162 & 0.453 \\
 &  & 20 & 0.856 & 0.613 & 0.492 & 0.380 & 0.686 & 0.755 & 0.630 & 0.174 & 0.476 \\
\midrule
+ TESSERA & XGBoost & 0 & 0.932 & 0.860 & 0.790 & 0.915 & 0.809 & 0.883 & 0.865 & 0.057 & 0.142 \\
 &  & 10 & 0.925 & 0.801 & 0.427 & 0.772 & 0.844 & 0.885 & 0.776 & 0.179 & 0.498 \\
 &  & 20 & 0.919 & 0.790 & 0.444 & 0.724 & 0.839 & 0.839 & 0.759 & 0.167 & 0.475 \\
 & CNN 3$\times$3 & 0 & 0.876 & 0.766 & 0.759 & 0.905 & 0.716 & 0.848 & 0.812 & 0.075 & 0.189 \\
 &  & 10 & 0.939 & 0.728 & 0.478 & 0.746 & 0.785 & 0.867 & 0.757 & 0.158 & 0.461 \\
 &  & 20 & 0.952 & 0.746 & 0.436 & 0.507 & 0.817 & 0.855 & 0.719 & 0.204 & 0.515 \\
 & CNN 5$\times$5 & 0 & 0.898 & 0.764 & 0.763 & 0.884 & 0.683 & 0.752 & 0.791 & 0.083 & 0.215 \\
 &  & 10 & 0.948 & 0.718 & 0.446 & 0.758 & 0.726 & 0.765 & 0.727 & 0.161 & 0.501 \\
 &  & 20 & 0.965 & 0.719 & 0.423 & 0.464 & 0.798 & 0.752 & 0.687 & 0.207 & 0.541 \\
 & CNN 7$\times$7 & 0 & 0.860 & 0.709 & 0.731 & 0.889 & 0.638 & 0.767 & 0.765 & 0.095 & 0.251 \\
 &  & 10 & 0.914 & 0.668 & 0.479 & 0.696 & 0.400 & 0.801 & 0.660 & 0.193 & 0.514 \\
 &  & 20 & 0.924 & 0.705 & 0.427 & 0.530 & 0.631 & 0.806 & 0.670 & 0.182 & 0.498 \\
\midrule
Coordinate-only (x, y) & XGBoost & 0 & 0.580 & 0.446 & 0.202 & 0.295 & 0.419 & 0.294 & 0.373 & 0.135 & 0.378 \\
 &  & 10 & 0.276 & 0.333 & 0.163 & 0.222 & 0.200 & 0.459 & 0.275 & 0.108 & 0.296 \\
 &  & 20 & 0.596 & 0.280 & 0.143 & 0.268 & 0.262 & 0.175 & 0.287 & 0.161 & 0.453 \\
 & CNN 3$\times$3 & 0 & 0.654 & 0.350 & 0.231 & 0.326 & 0.253 & 0.266 & 0.347 & 0.157 & 0.423 \\
 & CNN 5$\times$5 & 0 & 0.370 & 0.404 & 0.187 & 0.498 & 0.357 & 0.334 & 0.358 & 0.102 & 0.311 \\
 & CNN 7$\times$7 & 0 & 0.317 & 0.439 & 0.179 & 0.361 & 0.288 & 0.135 & 0.287 & 0.113 & 0.304 \\
\bottomrule
\end{tabular}
\end{threeparttable}%
}
\end{table}

% ============================================================================
% Supplementary table - Per-fold parcel ROC-AUC
% label: tab:sup_per_fold_roc_auc
% ============================================================================
\begin{table}[H]
\centering
\caption{Per-fold out-of-fold parcel ROC-AUC for every feature set, architecture and training-exclusion buffer, including the coordinate-only control.}
\label{tab:sup_per_fold_roc_auc}
\footnotesize
\setlength{\tabcolsep}{4pt}
\resizebox{\textwidth}{!}{%
\begin{threeparttable}
\begin{tabular}{llrrrrrrrrrr}
\toprule
 & & & \multicolumn{6}{c}{Outer fold} & & & \\
\cmidrule(lr){4-9}
Feature set & Architecture & Buffer & 1 & 2 & 3 & 4 & 5 & 6 & Mean & SD & Range \\
& & (km) & & & & & & & & & \\
\midrule
Baseline & XGBoost & 0 & 0.817 & 0.825 & 0.866 & 0.844 & 0.783 & 0.876 & 0.835 & 0.034 & 0.093 \\
 &  & 10 & 0.812 & 0.825 & 0.841 & 0.803 & 0.850 & 0.881 & 0.835 & 0.028 & 0.077 \\
 &  & 20 & 0.814 & 0.835 & 0.773 & 0.771 & 0.839 & 0.884 & 0.819 & 0.043 & 0.113 \\
 & CNN 3$\times$3 & 0 & 0.818 & 0.814 & 0.833 & 0.847 & 0.763 & 0.882 & 0.826 & 0.040 & 0.119 \\
 &  & 10 & 0.817 & 0.814 & 0.784 & 0.755 & 0.777 & 0.879 & 0.804 & 0.043 & 0.124 \\
 &  & 20 & 0.791 & 0.820 & 0.659 & 0.636 & 0.688 & 0.888 & 0.747 & 0.101 & 0.252 \\
 & CNN 5$\times$5 & 0 & 0.788 & 0.848 & 0.812 & 0.845 & 0.742 & 0.875 & 0.818 & 0.048 & 0.133 \\
 &  & 10 & 0.787 & 0.823 & 0.728 & 0.764 & 0.762 & 0.872 & 0.789 & 0.051 & 0.144 \\
 &  & 20 & 0.789 & 0.752 & 0.639 & 0.629 & 0.747 & 0.785 & 0.723 & 0.071 & 0.160 \\
 & CNN 7$\times$7 & 0 & 0.742 & 0.828 & 0.785 & 0.840 & 0.725 & 0.888 & 0.802 & 0.062 & 0.163 \\
 &  & 10 & 0.797 & 0.823 & 0.721 & 0.762 & 0.742 & 0.875 & 0.787 & 0.057 & 0.155 \\
 &  & 20 & 0.802 & 0.811 & 0.725 & 0.726 & 0.706 & 0.858 & 0.771 & 0.061 & 0.152 \\
\midrule
+ Conventional EO & XGBoost & 0 & 0.960 & 0.901 & 0.931 & 0.941 & 0.894 & 0.959 & 0.931 & 0.028 & 0.066 \\
 &  & 10 & 0.957 & 0.876 & 0.873 & 0.901 & 0.915 & 0.953 & 0.912 & 0.037 & 0.085 \\
 &  & 20 & 0.949 & 0.879 & 0.895 & 0.863 & 0.917 & 0.932 & 0.906 & 0.033 & 0.086 \\
 & CNN 3$\times$3 & 0 & 0.963 & 0.871 & 0.907 & 0.941 & 0.880 & 0.959 & 0.920 & 0.040 & 0.092 \\
 &  & 10 & 0.967 & 0.858 & 0.880 & 0.867 & 0.864 & 0.954 & 0.898 & 0.049 & 0.108 \\
 &  & 20 & 0.956 & 0.869 & 0.865 & 0.687 & 0.822 & 0.950 & 0.858 & 0.099 & 0.269 \\
 & CNN 5$\times$5 & 0 & 0.957 & 0.868 & 0.884 & 0.943 & 0.793 & 0.961 & 0.901 & 0.066 & 0.168 \\
 &  & 10 & 0.963 & 0.878 & 0.882 & 0.862 & 0.851 & 0.956 & 0.899 & 0.049 & 0.112 \\
 &  & 20 & 0.952 & 0.864 & 0.877 & 0.674 & 0.820 & 0.952 & 0.856 & 0.103 & 0.279 \\
 & CNN 7$\times$7 & 0 & 0.963 & 0.851 & 0.872 & 0.945 & 0.765 & 0.958 & 0.892 & 0.078 & 0.198 \\
 &  & 10 & 0.968 & 0.831 & 0.839 & 0.869 & 0.832 & 0.953 & 0.882 & 0.063 & 0.137 \\
 &  & 20 & 0.965 & 0.827 & 0.803 & 0.702 & 0.813 & 0.947 & 0.843 & 0.099 & 0.263 \\
\midrule
+ AlphaEarth & XGBoost & 0 & 0.968 & 0.896 & 0.960 & 0.957 & 0.921 & 0.962 & 0.944 & 0.029 & 0.072 \\
 &  & 10 & 0.967 & 0.881 & 0.938 & 0.928 & 0.932 & 0.955 & 0.933 & 0.030 & 0.087 \\
 &  & 20 & 0.966 & 0.890 & 0.938 & 0.870 & 0.922 & 0.942 & 0.921 & 0.035 & 0.096 \\
 & CNN 3$\times$3 & 0 & 0.964 & 0.877 & 0.949 & 0.955 & 0.907 & 0.928 & 0.930 & 0.033 & 0.087 \\
 &  & 10 & 0.976 & 0.858 & 0.938 & 0.892 & 0.870 & 0.938 & 0.912 & 0.046 & 0.118 \\
 &  & 20 & 0.982 & 0.859 & 0.931 & 0.857 & 0.878 & 0.951 & 0.909 & 0.052 & 0.125 \\
 & CNN 5$\times$5 & 0 & 0.926 & 0.869 & 0.947 & 0.938 & 0.904 & 0.958 & 0.924 & 0.032 & 0.088 \\
 &  & 10 & 0.921 & 0.857 & 0.927 & 0.847 & 0.873 & 0.957 & 0.897 & 0.044 & 0.110 \\
 &  & 20 & 0.928 & 0.854 & 0.928 & 0.767 & 0.873 & 0.959 & 0.885 & 0.069 & 0.191 \\
 & CNN 7$\times$7 & 0 & 0.949 & 0.868 & 0.834 & 0.920 & 0.849 & 0.937 & 0.893 & 0.049 & 0.115 \\
 &  & 10 & 0.956 & 0.844 & 0.829 & 0.809 & 0.806 & 0.939 & 0.864 & 0.067 & 0.150 \\
 &  & 20 & 0.947 & 0.861 & 0.863 & 0.674 & 0.896 & 0.957 & 0.866 & 0.103 & 0.283 \\
\midrule
+ TESSERA & XGBoost & 0 & 0.975 & 0.939 & 0.970 & 0.964 & 0.948 & 0.970 & 0.961 & 0.014 & 0.036 \\
 &  & 10 & 0.965 & 0.922 & 0.879 & 0.919 & 0.952 & 0.969 & 0.934 & 0.034 & 0.090 \\
 &  & 20 & 0.969 & 0.925 & 0.880 & 0.871 & 0.951 & 0.962 & 0.926 & 0.042 & 0.098 \\
 & CNN 3$\times$3 & 0 & 0.961 & 0.902 & 0.967 & 0.959 & 0.916 & 0.965 & 0.945 & 0.028 & 0.065 \\
 &  & 10 & 0.977 & 0.893 & 0.883 & 0.917 & 0.930 & 0.969 & 0.928 & 0.038 & 0.093 \\
 &  & 20 & 0.984 & 0.912 & 0.871 & 0.729 & 0.944 & 0.964 & 0.901 & 0.093 & 0.255 \\
 & CNN 5$\times$5 & 0 & 0.971 & 0.903 & 0.967 & 0.960 & 0.908 & 0.944 & 0.942 & 0.030 & 0.068 \\
 &  & 10 & 0.982 & 0.889 & 0.873 & 0.912 & 0.902 & 0.948 & 0.918 & 0.040 & 0.109 \\
 &  & 20 & 0.988 & 0.909 & 0.864 & 0.748 & 0.944 & 0.951 & 0.901 & 0.086 & 0.240 \\
 & CNN 7$\times$7 & 0 & 0.945 & 0.885 & 0.957 & 0.952 & 0.876 & 0.950 & 0.927 & 0.037 & 0.082 \\
 &  & 10 & 0.973 & 0.879 & 0.889 & 0.887 & 0.763 & 0.960 & 0.892 & 0.075 & 0.211 \\
 &  & 20 & 0.972 & 0.909 & 0.871 & 0.756 & 0.896 & 0.957 & 0.894 & 0.077 & 0.216 \\
\midrule
Coordinate-only (x, y) & XGBoost & 0 & 0.705 & 0.604 & 0.675 & 0.530 & 0.743 & 0.678 & 0.656 & 0.077 & 0.213 \\
 &  & 10 & 0.553 & 0.514 & 0.483 & 0.377 & 0.538 & 0.705 & 0.528 & 0.107 & 0.328 \\
 &  & 20 & 0.685 & 0.467 & 0.505 & 0.462 & 0.613 & 0.547 & 0.546 & 0.088 & 0.223 \\
 & CNN 3$\times$3 & 0 & 0.781 & 0.487 & 0.549 & 0.637 & 0.628 & 0.626 & 0.618 & 0.099 & 0.293 \\
 & CNN 5$\times$5 & 0 & 0.662 & 0.648 & 0.622 & 0.691 & 0.604 & 0.699 & 0.654 & 0.037 & 0.094 \\
 & CNN 7$\times$7 & 0 & 0.605 & 0.680 & 0.559 & 0.657 & 0.630 & 0.436 & 0.594 & 0.088 & 0.244 \\
\bottomrule
\end{tabular}
\end{threeparttable}%
}
\end{table}

% ============================================================================
% Supplementary table - Per-fold parcel F1
% label: tab:sup_per_fold_f1
% ============================================================================
\begin{table}[H]
\centering
\caption{Per-fold out-of-fold parcel $F_1$ for every feature set, architecture and training-exclusion buffer, including the coordinate-only control.}
\label{tab:sup_per_fold_f1}
\footnotesize
\setlength{\tabcolsep}{4pt}
\resizebox{\textwidth}{!}{%
\begin{threeparttable}
\begin{tabular}{llrrrrrrrrrr}
\toprule
 & & & \multicolumn{6}{c}{Outer fold} & & & \\
\cmidrule(lr){4-9}
Feature set & Architecture & Buffer & 1 & 2 & 3 & 4 & 5 & 6 & Mean & SD & Range \\
& & (km) & & & & & & & & & \\
\midrule
Baseline & XGBoost & 0 & 0.479 & 0.642 & 0.273 & 0.649 & 0.477 & 0.557 & 0.513 & 0.140 & 0.376 \\
 &  & 10 & 0.428 & 0.572 & 0.280 & 0.579 & 0.456 & 0.558 & 0.479 & 0.116 & 0.299 \\
 &  & 20 & 0.440 & 0.400 & 0.121 & 0.524 & 0.489 & 0.568 & 0.424 & 0.160 & 0.447 \\
 & CNN 3$\times$3 & 0 & 0.564 & 0.657 & 0.413 & 0.616 & 0.467 & 0.606 & 0.554 & 0.095 & 0.244 \\
 &  & 10 & 0.498 & 0.635 & 0.244 & 0.534 & 0.433 & 0.580 & 0.487 & 0.138 & 0.391 \\
 &  & 20 & 0.515 & 0.487 & 0.048 & 0.501 & 0.412 & 0.587 & 0.425 & 0.193 & 0.539 \\
 & CNN 5$\times$5 & 0 & 0.335 & 0.668 & 0.377 & 0.654 & 0.463 & 0.591 & 0.515 & 0.143 & 0.333 \\
 &  & 10 & 0.327 & 0.584 & 0.215 & 0.544 & 0.467 & 0.579 & 0.453 & 0.151 & 0.370 \\
 &  & 20 & 0.362 & 0.471 & 0.167 & 0.496 & 0.441 & 0.496 & 0.405 & 0.127 & 0.329 \\
 & CNN 7$\times$7 & 0 & 0.360 & 0.644 & 0.340 & 0.647 & 0.444 & 0.604 & 0.507 & 0.142 & 0.307 \\
 &  & 10 & 0.415 & 0.644 & 0.198 & 0.532 & 0.444 & 0.553 & 0.464 & 0.154 & 0.446 \\
 &  & 20 & 0.484 & 0.566 & 0.248 & 0.521 & 0.426 & 0.555 & 0.467 & 0.119 & 0.318 \\
\midrule
+ Conventional EO & XGBoost & 0 & 0.802 & 0.727 & 0.472 & 0.800 & 0.638 & 0.772 & 0.702 & 0.128 & 0.329 \\
 &  & 10 & 0.616 & 0.714 & 0.269 & 0.704 & 0.601 & 0.759 & 0.611 & 0.178 & 0.489 \\
 &  & 20 & 0.670 & 0.575 & 0.197 & 0.582 & 0.611 & 0.680 & 0.552 & 0.179 & 0.482 \\
 & CNN 3$\times$3 & 0 & 0.829 & 0.715 & 0.506 & 0.797 & 0.615 & 0.756 & 0.703 & 0.122 & 0.322 \\
 &  & 10 & 0.833 & 0.698 & 0.413 & 0.687 & 0.533 & 0.754 & 0.653 & 0.154 & 0.421 \\
 &  & 20 & 0.843 & 0.644 & 0.293 & 0.513 & 0.525 & 0.742 & 0.593 & 0.194 & 0.549 \\
 & CNN 5$\times$5 & 0 & 0.805 & 0.684 & 0.427 & 0.814 & 0.565 & 0.784 & 0.680 & 0.156 & 0.387 \\
 &  & 10 & 0.830 & 0.705 & 0.440 & 0.665 & 0.556 & 0.761 & 0.659 & 0.142 & 0.390 \\
 &  & 20 & 0.813 & 0.679 & 0.528 & 0.514 & 0.545 & 0.732 & 0.635 & 0.124 & 0.299 \\
 & CNN 7$\times$7 & 0 & 0.807 & 0.688 & 0.351 & 0.812 & 0.478 & 0.775 & 0.652 & 0.194 & 0.461 \\
 &  & 10 & 0.782 & 0.648 & 0.281 & 0.668 & 0.569 & 0.753 & 0.617 & 0.181 & 0.501 \\
 &  & 20 & 0.793 & 0.611 & 0.194 & 0.505 & 0.527 & 0.712 & 0.557 & 0.209 & 0.599 \\
\midrule
+ AlphaEarth & XGBoost & 0 & 0.751 & 0.739 & 0.529 & 0.852 & 0.676 & 0.789 & 0.723 & 0.111 & 0.323 \\
 &  & 10 & 0.641 & 0.728 & 0.448 & 0.775 & 0.587 & 0.771 & 0.658 & 0.127 & 0.327 \\
 &  & 20 & 0.684 & 0.707 & 0.277 & 0.505 & 0.638 & 0.731 & 0.590 & 0.173 & 0.454 \\
 & CNN 3$\times$3 & 0 & 0.797 & 0.709 & 0.545 & 0.820 & 0.636 & 0.686 & 0.699 & 0.102 & 0.275 \\
 &  & 10 & 0.771 & 0.693 & 0.514 & 0.697 & 0.488 & 0.711 & 0.646 & 0.116 & 0.283 \\
 &  & 20 & 0.777 & 0.685 & 0.641 & 0.654 & 0.467 & 0.741 & 0.661 & 0.108 & 0.310 \\
 & CNN 5$\times$5 & 0 & 0.704 & 0.718 & 0.575 & 0.803 & 0.629 & 0.735 & 0.694 & 0.081 & 0.228 \\
 &  & 10 & 0.584 & 0.697 & 0.562 & 0.623 & 0.493 & 0.717 & 0.613 & 0.085 & 0.224 \\
 &  & 20 & 0.616 & 0.674 & 0.624 & 0.534 & 0.509 & 0.695 & 0.609 & 0.074 & 0.186 \\
 & CNN 7$\times$7 & 0 & 0.720 & 0.740 & 0.326 & 0.730 & 0.622 & 0.721 & 0.643 & 0.161 & 0.414 \\
 &  & 10 & 0.615 & 0.718 & 0.286 & 0.605 & 0.493 & 0.706 & 0.571 & 0.162 & 0.433 \\
 &  & 20 & 0.617 & 0.703 & 0.317 & 0.505 & 0.553 & 0.755 & 0.575 & 0.156 & 0.438 \\
\midrule
+ TESSERA & XGBoost & 0 & 0.783 & 0.777 & 0.547 & 0.828 & 0.758 & 0.815 & 0.751 & 0.103 & 0.280 \\
 &  & 10 & 0.688 & 0.748 & 0.273 & 0.722 & 0.739 & 0.804 & 0.662 & 0.195 & 0.532 \\
 &  & 20 & 0.688 & 0.766 & 0.127 & 0.557 & 0.744 & 0.775 & 0.609 & 0.250 & 0.648 \\
 & CNN 3$\times$3 & 0 & 0.797 & 0.732 & 0.670 & 0.837 & 0.645 & 0.787 & 0.745 & 0.076 & 0.192 \\
 &  & 10 & 0.794 & 0.736 & 0.299 & 0.701 & 0.615 & 0.798 & 0.657 & 0.188 & 0.498 \\
 &  & 20 & 0.835 & 0.734 & 0.322 & 0.518 & 0.569 & 0.782 & 0.627 & 0.193 & 0.513 \\
 & CNN 5$\times$5 & 0 & 0.773 & 0.748 & 0.485 & 0.821 & 0.640 & 0.737 & 0.700 & 0.121 & 0.336 \\
 &  & 10 & 0.758 & 0.732 & 0.307 & 0.669 & 0.506 & 0.759 & 0.622 & 0.181 & 0.452 \\
 &  & 20 & 0.802 & 0.721 & 0.472 & 0.206 & 0.609 & 0.728 & 0.590 & 0.220 & 0.596 \\
 & CNN 7$\times$7 & 0 & 0.775 & 0.741 & 0.588 & 0.799 & 0.625 & 0.748 & 0.713 & 0.085 & 0.211 \\
 &  & 10 & 0.670 & 0.729 & 0.400 & 0.676 & 0.434 & 0.776 & 0.614 & 0.158 & 0.376 \\
 &  & 20 & 0.751 & 0.727 & 0.429 & 0.527 & 0.571 & 0.759 & 0.627 & 0.138 & 0.330 \\
\midrule
Coordinate-only (x, y) & XGBoost & 0 & 0.456 & 0.424 & 0.334 & 0.178 & 0.326 & 0.338 & 0.343 & 0.097 & 0.277 \\
 &  & 10 & 0.080 & 0.414 & 0.129 & 0.152 & 0.334 & 0.134 & 0.207 & 0.134 & 0.334 \\
 &  & 20 & 0.000 & 0.000 & 0.000 & 0.211 & 0.343 & 0.000 & 0.092 & 0.149 & 0.343 \\
 & CNN 3$\times$3 & 0 & 0.395 & 0.449 & 0.000 & 0.429 & 0.034 & 0.328 & 0.272 & 0.202 & 0.449 \\
 & CNN 5$\times$5 & 0 & 0.400 & 0.512 & 0.287 & 0.422 & 0.260 & 0.379 & 0.377 & 0.092 & 0.252 \\
 & CNN 7$\times$7 & 0 & 0.387 & 0.419 & 0.179 & 0.506 & 0.059 & 0.303 & 0.309 & 0.165 & 0.447 \\
\bottomrule
\end{tabular}
\end{threeparttable}%
}
\end{table}

% ============================================================================
% Supplementary table - Leave-one-fold-out PR-AUC
% label: tab:sup_lofo_pr_auc
% ============================================================================
\begin{table}[H]
\centering
\caption{Leave-one-fold-out sensitivity: pooled out-of-fold parcel PR-AUC per configuration when each fold's parcels are excluded in turn.}
\label{tab:sup_lofo_pr_auc}
\footnotesize
\setlength{\tabcolsep}{4pt}
\resizebox{\textwidth}{!}{%
\begin{threeparttable}
\begin{tabular}{lcccccccc}
\toprule
Configuration & All folds & Excl. fold 1 & Excl. fold 2 & Excl. fold 3 & Excl. fold 4 & Excl. fold 5 & Excl. fold 6 & Range \\
\midrule
OGF prevalence of subset & 0.209 & 0.206 & 0.195 & 0.225 & 0.198 & 0.212 & 0.220 & 0.031 \\
Baseline (0 km) & 0.548 & 0.550 & 0.540 & 0.560 & 0.516 & 0.588 & 0.545 & 0.072 \\
Baseline (10 km) & 0.475 & 0.478 & 0.476 & 0.486 & 0.458 & 0.527 & 0.470 & 0.068 \\
Baseline + EO (0 km) & 0.765 & 0.763 & 0.782 & 0.785 & 0.745 & 0.775 & 0.748 & 0.040 \\
Baseline + EO (10 km) & 0.684 & 0.689 & 0.684 & 0.712 & 0.665 & 0.702 & 0.658 & 0.053 \\
Baseline + TESSERA (0 km) & 0.840 & 0.837 & 0.857 & 0.854 & 0.824 & 0.847 & 0.833 & 0.033 \\
Baseline + TESSERA (10 km) & 0.727 & 0.732 & 0.709 & 0.770 & 0.740 & 0.718 & 0.705 & 0.065 \\
Baseline + AlphaEarth (0 km) & 0.765 & 0.764 & 0.781 & 0.778 & 0.740 & 0.785 & 0.755 & 0.044 \\
Baseline + AlphaEarth (10 km) & 0.701 & 0.710 & 0.702 & 0.719 & 0.682 & 0.721 & 0.690 & 0.040 \\
Coordinate-only control (x, y) (0 km) & 0.347 & 0.317 & 0.322 & 0.372 & 0.371 & 0.339 & 0.365 & 0.054 \\
Coordinate-only control (x, y) (10 km) & 0.233 & 0.237 & 0.199 & 0.247 & 0.247 & 0.244 & 0.228 & 0.049 \\
\bottomrule
\end{tabular}
\end{threeparttable}%
}
\end{table}

% ============================================================================
% Supplementary table - Leave-one-fold-out contrasts
% label: tab:sup_lofo_contrasts
% ============================================================================

% ============================================================================
% Supplementary table - Leave-one-fold-out contrasts
% label: tab:sup_lofo_contrasts
% ============================================================================
\begin{table}[H]
\centering
\caption{Leave-one-fold-out sensitivity: paired PR-AUC contrasts recomputed on the identical five-fold subsets.}
\label{tab:sup_lofo_contrasts}
\footnotesize
\setlength{\tabcolsep}{4pt}
\resizebox{\textwidth}{!}{%
\begin{threeparttable}
\begin{tabular}{lcccccccc}
\toprule
Contrast & All folds & Excl. fold 1 & Excl. fold 2 & Excl. fold 3 & Excl. fold 4 & Excl. fold 5 & Excl. fold 6 & Range \\
\midrule
OGF prevalence of subset & 0.209 & 0.206 & 0.195 & 0.225 & 0.198 & 0.212 & 0.220 & 0.031 \\
{[Baseline + EO] $-$ [Baseline]} (0 km) & 0.217 & 0.212 & 0.242 & 0.225 & 0.230 & 0.187 & 0.203 & 0.055 \\
{[Baseline + EO] $-$ [Baseline]} (10 km) & 0.208 & 0.211 & 0.209 & 0.226 & 0.206 & 0.176 & 0.188 & 0.051 \\
{[Baseline + AlphaEarth] $-$ [Baseline]} (0 km) & 0.217 & 0.213 & 0.241 & 0.218 & 0.225 & 0.197 & 0.211 & 0.045 \\
{[Baseline + AlphaEarth] $-$ [Baseline]} (10 km) & 0.226 & 0.231 & 0.227 & 0.234 & 0.223 & 0.195 & 0.220 & 0.039 \\
{[Baseline + TESSERA] $-$ [Baseline]} (0 km) & 0.292 & 0.287 & 0.317 & 0.294 & 0.308 & 0.259 & 0.288 & 0.058 \\
{[Baseline + TESSERA] $-$ [Baseline]} (10 km) & 0.252 & 0.253 & 0.234 & 0.284 & 0.281 & 0.191 & 0.235 & 0.093 \\
{[Baseline + AlphaEarth] $-$ [Baseline + EO]} (0 km) & $-$0.000 & 0.001 & $-$0.001 & $-$0.007 & $-$0.005 & 0.010 & 0.007 & 0.017 \\
{[Baseline + AlphaEarth] $-$ [Baseline + EO]} (10 km) & 0.018 & 0.020 & 0.018 & 0.008 & 0.017 & 0.019 & 0.031 & 0.024 \\
{[Baseline + TESSERA] $-$ [Baseline + EO]} (0 km) & 0.075 & 0.075 & 0.075 & 0.069 & 0.078 & 0.072 & 0.085 & 0.017 \\
{[Baseline + TESSERA] $-$ [Baseline + EO]} (10 km) & 0.043 & 0.042 & 0.025 & 0.058 & 0.075 & 0.016 & 0.047 & 0.059 \\
{[Baseline + TESSERA] $-$ [Baseline + AlphaEarth]} (0 km) & 0.075 & 0.074 & 0.076 & 0.076 & 0.083 & 0.063 & 0.078 & 0.021 \\
{[Baseline + TESSERA] $-$ [Baseline + AlphaEarth]} (10 km) & 0.026 & 0.022 & 0.007 & 0.051 & 0.058 & $-$0.003 & 0.015 & 0.061 \\
\bottomrule
\end{tabular}
\end{threeparttable}%
}
\end{table}

% ============================================================================
% Supplementary table - Seed-sweep configurations
% label: tab:sup_seed_sweep
% ============================================================================
\begin{table}[H]
\centering
\caption{Seed sweep: retraining variability of the pooled out-of-fold parcel PR-AUC (XGBoost, five training seeds at the tuned per-fold hyperparameters).}
\label{tab:sup_seed_sweep}
\footnotesize
\setlength{\tabcolsep}{4pt}
\begin{threeparttable}
\begin{tabular}{lccccc}
\toprule
Configuration & Paper seed & $n$ seeds & Seed mean & Seed SD & Seed min max \\
\midrule
Baseline (0 km) & 0.548 & 5 & 0.547 & 0.001 & 0.546$-$0.548 \\
Baseline (10 km) & 0.475 & 5 & 0.475 & 0.002 & 0.473$-$0.478 \\
Baseline + AlphaEarth (0 km) & 0.765 & 5 & 0.769 & 0.003 & 0.765$-$0.772 \\
Baseline + AlphaEarth (10 km) & 0.701 & 5 & 0.699 & 0.004 & 0.693$-$0.701 \\
Baseline + EO (0 km) & 0.765 & 5 & 0.766 & 0.003 & 0.764$-$0.770 \\
Baseline + EO (10 km) & 0.684 & 5 & 0.686 & 0.003 & 0.683$-$0.688 \\
Baseline + TESSERA (0 km) & 0.840 & 5 & 0.841 & 0.001 & 0.839$-$0.842 \\
Baseline + TESSERA (10 km) & 0.727 & 5 & 0.726 & 0.002 & 0.723$-$0.728 \\
\bottomrule
\end{tabular}
\end{threeparttable}
\end{table}

% ============================================================================
% Supplementary table - Seed-sweep contrasts
% label: tab:sup_seed_sweep_contrasts
% ============================================================================
\begin{table}[H]
\centering
\caption{Seed sweep: retraining variability of the pre-specified PR-AUC contrasts.}
\label{tab:sup_seed_sweep_contrasts}
\footnotesize
\setlength{\tabcolsep}{4pt}
\begin{threeparttable}
\begin{tabular}{lccccc}
\toprule
Contrast & Paper seed & $n$ seeds & Seed mean & Seed SD & Seed min max \\
\midrule
{[Baseline + AlphaEarth] $-$ [Baseline]} (0 km) & 0.217 & 5 & 0.222 & 0.003 & 0.217$-$0.225 \\
{[Baseline + AlphaEarth] $-$ [Baseline]} (10 km) & 0.226 & 5 & 0.224 & 0.004 & 0.219$-$0.228 \\
{[Baseline + AlphaEarth] $-$ [Baseline + EO]} (0 km) & $-$0.000 & 5 & 0.003 & 0.002 & $-$0.000$-$0.005 \\
{[Baseline + AlphaEarth] $-$ [Baseline + EO]} (10 km) & 0.018 & 5 & 0.013 & 0.006 & 0.005$-$0.018 \\
{[Baseline + EO] $-$ [Baseline]} (0 km) & 0.217 & 5 & 0.219 & 0.002 & 0.217$-$0.223 \\
{[Baseline + EO] $-$ [Baseline]} (10 km) & 0.208 & 5 & 0.211 & 0.003 & 0.208$-$0.214 \\
{[Baseline + TESSERA] $-$ [Baseline]} (0 km) & 0.292 & 5 & 0.294 & 0.001 & 0.292$-$0.296 \\
{[Baseline + TESSERA] $-$ [Baseline]} (10 km) & 0.252 & 5 & 0.251 & 0.003 & 0.246$-$0.254 \\
{[Baseline + TESSERA] $-$ [Baseline + AlphaEarth]} (0 km) & 0.075 & 5 & 0.072 & 0.003 & 0.069$-$0.075 \\
{[Baseline + TESSERA] $-$ [Baseline + AlphaEarth]} (10 km) & 0.026 & 5 & 0.027 & 0.005 & 0.022$-$0.035 \\
{[Baseline + TESSERA] $-$ [Baseline + EO]} (0 km) & 0.075 & 5 & 0.075 & 0.003 & 0.071$-$0.078 \\
{[Baseline + TESSERA] $-$ [Baseline + EO]} (10 km) & 0.043 & 5 & 0.040 & 0.003 & 0.036$-$0.043 \\
\bottomrule
\end{tabular}
\end{threeparttable}
\end{table}

% ============================================================================
% Supplementary table - Residual spatial dependence
% label: tab:sup_autocorrelation_residuals
% ============================================================================
\begingroup
\footnotesize
\setlength{\tabcolsep}{3pt}
\begin{longtable}{lrrrrlrrrrr}
\caption{Spatial dependence of the pooled out-of-fold parcel residuals of every configuration, including the coordinate-only control.}\label{tab:sup_autocorrelation_residuals}\\
\toprule
 & & \multicolumn{3}{c}{Effective range (km)} & & & \multicolumn{4}{c}{Moran's $I$ at distance band} \\
\cmidrule(lr){3-5}\cmidrule(lr){8-11}
Variable & $n$ & Sph. & Exp. & Gau. & Best & Nug.:sill & 5 km & 10 km & 15 km & 20 km \\
\midrule
\endfirsthead
\multicolumn{11}{l}{\footnotesize\textit{Table \ref{tab:sup_autocorrelation_residuals} continued from the previous page.}}\\
\toprule
 & & \multicolumn{3}{c}{Effective range (km)} & & & \multicolumn{4}{c}{Moran's $I$ at distance band} \\
\cmidrule(lr){3-5}\cmidrule(lr){8-11}
Variable & $n$ & Sph. & Exp. & Gau. & Best & Nug.:sill & 5 km & 10 km & 15 km & 20 km \\
\midrule
\endhead
\midrule
\multicolumn{11}{r}{\footnotesize\textit{Continued on the next page.}}\\
\endfoot
\bottomrule
\endlastfoot
\multicolumn{11}{l}{\textbf{Out-of-fold model residuals} ($k = 20$)}\\[1pt]
CNN 3$\times$3, Baseline & 4825 & 5.20 & 4.62 & 5.09 & Exp. & 0.19 & 0.119 & 0.061 & 0.026 & 0.009 \\
CNN 3$\times$3, AlphaEarth & 4825 & 8.57 & 8.38 & 8.26 & Exp. & 0.45 & 0.155 & 0.087 & 0.046 & 0.020 \\
CNN 3$\times$3, Conv.\ EO & 4825 & 5.46 & 4.59 & 5.29 & Sph. & 0.52 & 0.096 & 0.043 & 0.016 & 0.003 \\
CNN 3$\times$3, TESSERA & 4825 & 6.26 & 5.62 & 6.10 & Sph. & 0.54 & 0.103 & 0.052 & 0.025 & 0.007 \\
CNN 3$\times$3, Coordinate-only (x, y) & 4825 & 7.75 & 7.59 & 7.60 & Sph. & 0.30 & 0.254 & 0.109 & 0.027 & 0.008 \\
CNN 5$\times$5, Baseline & 4825 & 4.93 & 4.46 & 4.89 & Exp. & 0.22 & 0.104 & 0.056 & 0.029 & 0.011 \\
CNN 5$\times$5, AlphaEarth & 4825 & 7.52 & 6.97 & 7.35 & Exp. & 0.38 & 0.144 & 0.077 & 0.036 & 0.013 \\
CNN 5$\times$5, Conv.\ EO & 4825 & 5.09 & 4.26 & 5.02 & Sph. & 0.54 & 0.081 & 0.039 & 0.017 & 0.003 \\
CNN 5$\times$5, TESSERA & 4825 & 6.09 & 5.77 & 5.98 & Exp. & 0.35 & 0.119 & 0.067 & 0.042 & 0.020 \\
CNN 5$\times$5, Coordinate-only (x, y) & 4825 & 7.29 & 6.88 & 7.14 & Sph. & 0.31 & 0.232 & 0.098 & 0.023 & -0.004 \\
CNN 7$\times$7, Baseline & 4825 & 5.26 & 4.77 & 5.17 & Exp. & 0.20 & 0.119 & 0.062 & 0.030 & 0.012 \\
CNN 7$\times$7, AlphaEarth & 4825 & 6.10 & 5.00 & 5.88 & Sph. & 0.52 & 0.109 & 0.044 & 0.013 & -0.002 \\
CNN 7$\times$7, Conv.\ EO & 4825 & 5.41 & 4.53 & 5.31 & Sph. & 0.53 & 0.096 & 0.041 & 0.016 & 0.002 \\
CNN 7$\times$7, TESSERA & 4825 & 5.46 & 4.71 & 5.39 & Sph. & 0.52 & 0.087 & 0.041 & 0.016 & 0.003 \\
CNN 7$\times$7, Coordinate-only (x, y) & 4825 & -- & -- & 24.44 & Gau. & 0.41 & 0.499 & 0.376 & 0.264 & 0.172 \\
XGBoost, Baseline & 4825 & 5.31 & 4.87 & 5.20 & Exp. & 0.18 & 0.138 & 0.071 & 0.034 & 0.015 \\
XGBoost, AlphaEarth & 4825 & 6.54 & 6.01 & 6.37 & Exp. & 0.34 & 0.131 & 0.064 & 0.031 & 0.012 \\
XGBoost, Conv.\ EO & 4825 & 5.33 & 4.47 & 5.20 & Sph. & 0.49 & 0.112 & 0.052 & 0.019 & 0.003 \\
XGBoost, TESSERA & 4825 & 6.17 & 5.37 & 6.02 & Sph. & 0.56 & 0.105 & 0.049 & 0.021 & 0.003 \\
XGBoost, Coordinate-only (x, y) & 4825 & 5.58 & 5.11 & 5.56 & Sph. & 0.29 & 0.161 & 0.070 & 0.031 & 0.006 \\
\end{longtable}
\vspace{2pt}
\noindent\footnotesize A range is reported as ``--'' when the fit runs into a boundary of the analysed window (when the effective range reaches 90\% of the 40 km maximum lag or falls below the 1 km first lag class). \emph{Best} is the lowest-RMSE model with a resolved range (Sph. = spherical, Exp. = exponential, Gau. = Gaussian) and the nugget-to-sill ratio is of the selected model. All Moran's $I$ values are significant at $p < 0.05$ except those marked $\dagger$ (999 conditional permutations, so the attainable minimum is $p = 0.001$).\par
\endgroup

% ============================================================================
% Supplementary figures - spatial buffers
% ============================================================================

\begin{figure}[H]
    \centering
    \includegraphics[width=1\linewidth]{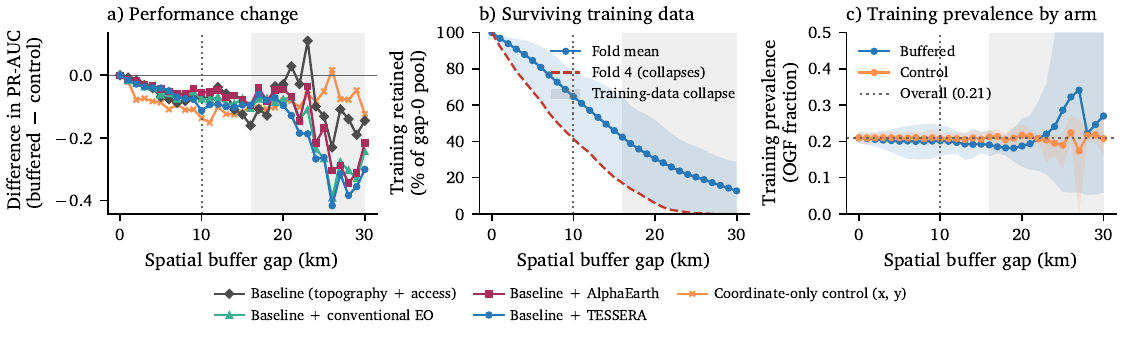}
    \caption{Spatial buffering analysis. (a) Relative performance difference between spatially gapped and parcel-matched control models; (b) Fall in training data with increasing spatial gap; (c) OGF prevalence with increasing spatial gap.}
    \label{fig:sup_buffer_performance}
\end{figure}

\begin{figure}[H]
    \centering
    \includegraphics[width=0.8\linewidth]{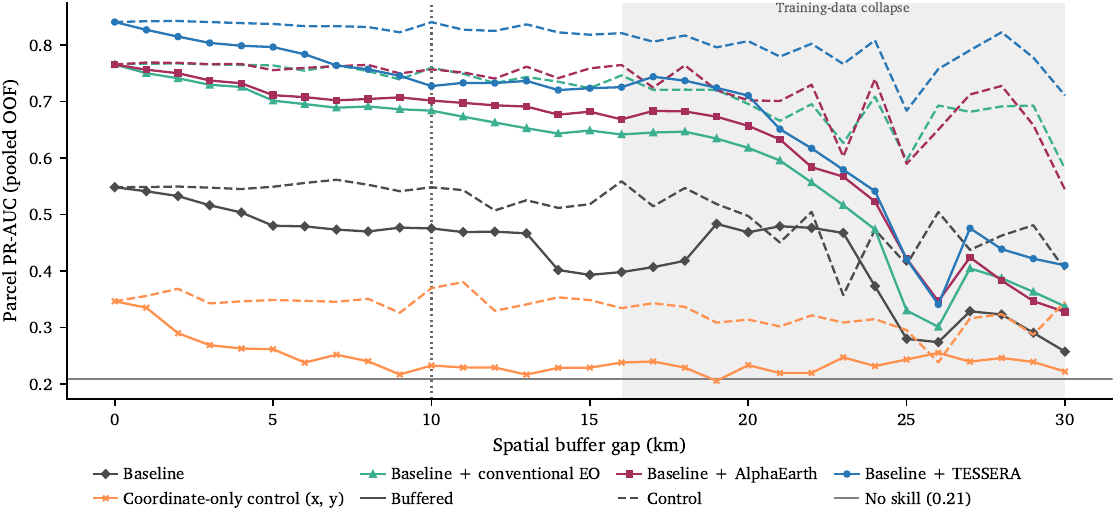}
    \caption{PR-AUC for spatially gapped and parcel-matched control models.}
    \label{fig:sup_abs_prauc_spatial_gap}
\end{figure}

% ============================================================================
% Supplementary table - Parcel vs native mapped areas
% label: tab:sup_area_parcel_vs_native
% ============================================================================
\begin{table}[H]
\centering
\caption{Mapped old-growth area at parcel resolution and at each product's native resolution. Percentage of full AOI (211,893\,ha)}
\label{tab:sup_area_parcel_vs_native}
\small
\begin{threeparttable}
\begin{tabular}{lrrrrrr}
\toprule
 & \multicolumn{2}{c}{Parcel} & \multicolumn{2}{c}{Native pixel-level} & \multicolumn{2}{c}{Difference} \\
\cmidrule(lr){2-3}\cmidrule(lr){4-5}\cmidrule(lr){6-7}
Product & (ha) & (\% AOI) & (ha) & (\% AOI) & (ha) & (\%) \\
\midrule
Sabatini (2020) & 20,865 & 9.8 & 23,644 & 11.2 & +2,779 & +13.3 \\
Munteanu et al. (2022) & 61,837 & 29.2 & 61,533 & 29.0 & -305 & -0.5 \\
Kathmann et al. (2017) & 64,146 & 30.3 & 64,547 & 30.5 & +400 & +0.6 \\
Schickhofer and Schwarz (2019) & 51,235 & 24.2 & 51,054 & 24.1 & -181 & -0.4 \\
\bottomrule
\end{tabular}
\end{threeparttable}
\end{table}

\begin{figure}[H]
    \centering
    \includegraphics[width=0.7\linewidth]{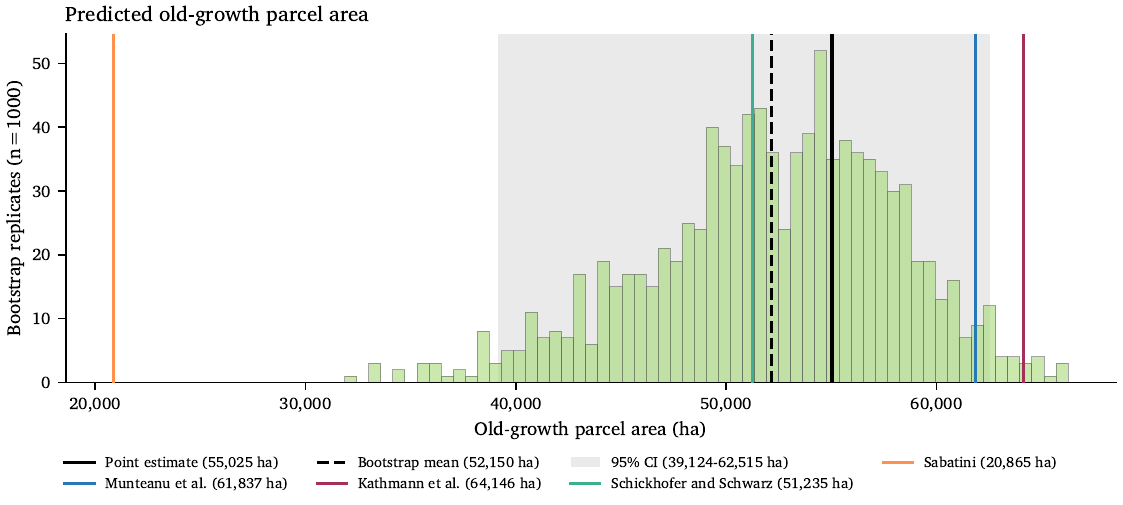}
    \caption{Histogram of OGF area estimates from 1{,}000 replicate area bootstrap using the XGBoost with \textit{baseline + TESSERA} model configuration, with point area estimate, bootstrap mean, 95\% CI, and existing study area estimates.}
    \label{fig:sup_area_bootstrap}
\end{figure}

% ============================================================================
% Supplementary table - Stratified ROC-AUC by forest type, altitude and slope
% label: tab:sup_stratified_roc_auc
% ============================================================================
\begin{table}[H]
\centering
\caption{ROC-AUC [95\% CI] of every product within strata of forest type, altitude and slope.}
\label{tab:sup_stratified_roc_auc}
\scriptsize
\setlength{\tabcolsep}{2pt}
\resizebox{\textwidth}{!}{%
\begin{threeparttable}
\begin{tabular}{lrrccccc}
\toprule
 & & & \multicolumn{5}{c}{ROC-AUC [95\% interval]} \\
\cmidrule(lr){4-8}
Stratum & $n$ & Prev. & Sabatini & Munteanu et al. & Kathmann et al. & Schickhofer & \textbf{This study} \\
 & & & (2020) & (2022) & (2017) & \& Schwarz (2019) & \\
\midrule
\multicolumn{8}{l}{\textit{Forest type}}\\
\quad broadleaf & 690 & 0.065 & 0.692 [0.554, 0.822] & 0.835 [0.709, 0.913] & 0.810 [0.664, 0.918] & 0.966 [0.944, 0.985] & 0.950 [0.906, 0.974] \\
\quad coniferous & 2{,}148 & 0.287 & 0.645 [0.562, 0.722] & 0.817 [0.776, 0.860] & 0.744 [0.701, 0.783] & 0.895 [0.848, 0.928] & 0.938 [0.911, 0.958] \\
\quad mixed & 1{,}922 & 0.177 & 0.670 [0.597, 0.755] & 0.901 [0.878, 0.923] & 0.795 [0.743, 0.843] & 0.915 [0.892, 0.932] & 0.967 [0.953, 0.980] \\
\midrule
\multicolumn{8}{l}{\textit{Altitude tertile (m)}}\\
\quad 535--1109 & 1{,}608 & 0.096 & 0.644 [0.546, 0.777] & 0.890 [0.830, 0.928] & 0.766 [0.692, 0.845] & 0.941 [0.917, 0.962] & 0.963 [0.942, 0.979] \\
\quad 1109--1406 & 1{,}608 & 0.228 & 0.714 [0.631, 0.771] & 0.877 [0.851, 0.903] & 0.833 [0.795, 0.864] & 0.920 [0.885, 0.944] & 0.967 [0.950, 0.980] \\
\quad 1406--1892 & 1{,}609 & 0.304 & 0.600 [0.523, 0.680] & 0.803 [0.757, 0.852] & 0.716 [0.663, 0.761] & 0.872 [0.821, 0.909] & 0.924 [0.893, 0.950] \\
\midrule
\multicolumn{8}{l}{\textit{Slope tertile ($^\circ$)}}\\
\quad 2--22 & 1{,}608 & 0.039 & 0.631 [0.533, 0.745] & 0.839 [0.793, 0.886] & 0.757 [0.667, 0.840] & 0.891 [0.839, 0.938] & 0.933 [0.892, 0.968] \\
\quad 22--29 & 1{,}608 & 0.175 & 0.614 [0.562, 0.680] & 0.856 [0.812, 0.894] & 0.773 [0.712, 0.824] & 0.901 [0.862, 0.936] & 0.947 [0.929, 0.964] \\
\quad 29--49 & 1{,}609 & 0.414 & 0.625 [0.538, 0.692] & 0.829 [0.789, 0.866] & 0.723 [0.687, 0.753] & 0.858 [0.815, 0.895] & 0.926 [0.900, 0.946] \\
\bottomrule
\end{tabular}
\end{threeparttable}%
}
\end{table}

% ============================================================================
% Supplementary table - Unlabelled-parcel agreement
% label: tab:sup_unlabelled_agreement
% ============================================================================
\begin{table}[H]
\centering
\caption{Agreement between the deployed model and Schickhofer and Schwarz (2019) by labelling stratum, with unlabelled parcels binned by recorded stand age. Area is the mapped 10 m pixel area within the unclipped parcel polygons, so the labelled OGF area exceeds the clipped figures reported in the manuscript.}
\label{tab:sup_unlabelled_agreement}
\footnotesize
\setlength{\tabcolsep}{4pt}
\resizebox{\textwidth}{!}{%
\begin{threeparttable}
\begin{tabular}{lccccccc}
\toprule
Stratum & Parcels & Area (ha) & \shortstack{This study\\OGF (\%)} & \shortstack{Schickhofer and\\Schwarz OGF (\%)} & \shortstack{Overall\\agreement (\%)} & \shortstack{OGF\\overlap (\%)} & \shortstack{Non-OGF\\overlap (\%)} \\
\midrule
Labelled OGF & 1,012 & 13,348 & 98.2 & 89.4 & 89.4 & 89.3 & 7.8 \\
Labelled non-OGF & 3,834 & 34,556 & 3.5 & 13.7 & 88.6 & 20.2 & 88.2 \\
Unlabelled, age 85-99 yr & 576 & 5,898 & 30.4 & 31.8 & 70.8 & 36.1 & 65.1 \\
Unlabelled, age 100-119 yr & 1,120 & 12,010 & 46.7 & 41.2 & 72.9 & 53.0 & 61.1 \\
Unlabelled, age 120-139 yr & 1,450 & 14,143 & 48.9 & 41.7 & 74.9 & 56.6 & 62.7 \\
Unlabelled, age 140-159 yr & 852 & 9,816 & 51.1 & 51.8 & 74.4 & 60.1 & 58.3 \\
Unlabelled, age 160+ yr & 766 & 10,888 & 54.3 & 49.1 & 67.4 & 52.0 & 49.5 \\
Unlabelled, no recorded age & 5,833 & 49,040 & 15.5 & 14.2 & 87.5 & 40.8 & 86.3 \\
All unlabelled & 10,597 & 101,794 & 29.9 & 27.3 & 80.8 & 49.8 & 76.3 \\
All parcels & 15,443 & 149,698 & 27.8 & 28.0 & 83.3 & 53.9 & 79.2 \\
\bottomrule
\end{tabular}
\end{threeparttable}%
}
\end{table}

% ============================================================================
% Supplementary figures - study agreement
% ============================================================================

\begin{figure}[H]
    \centering
    \includegraphics[width=0.9\linewidth]{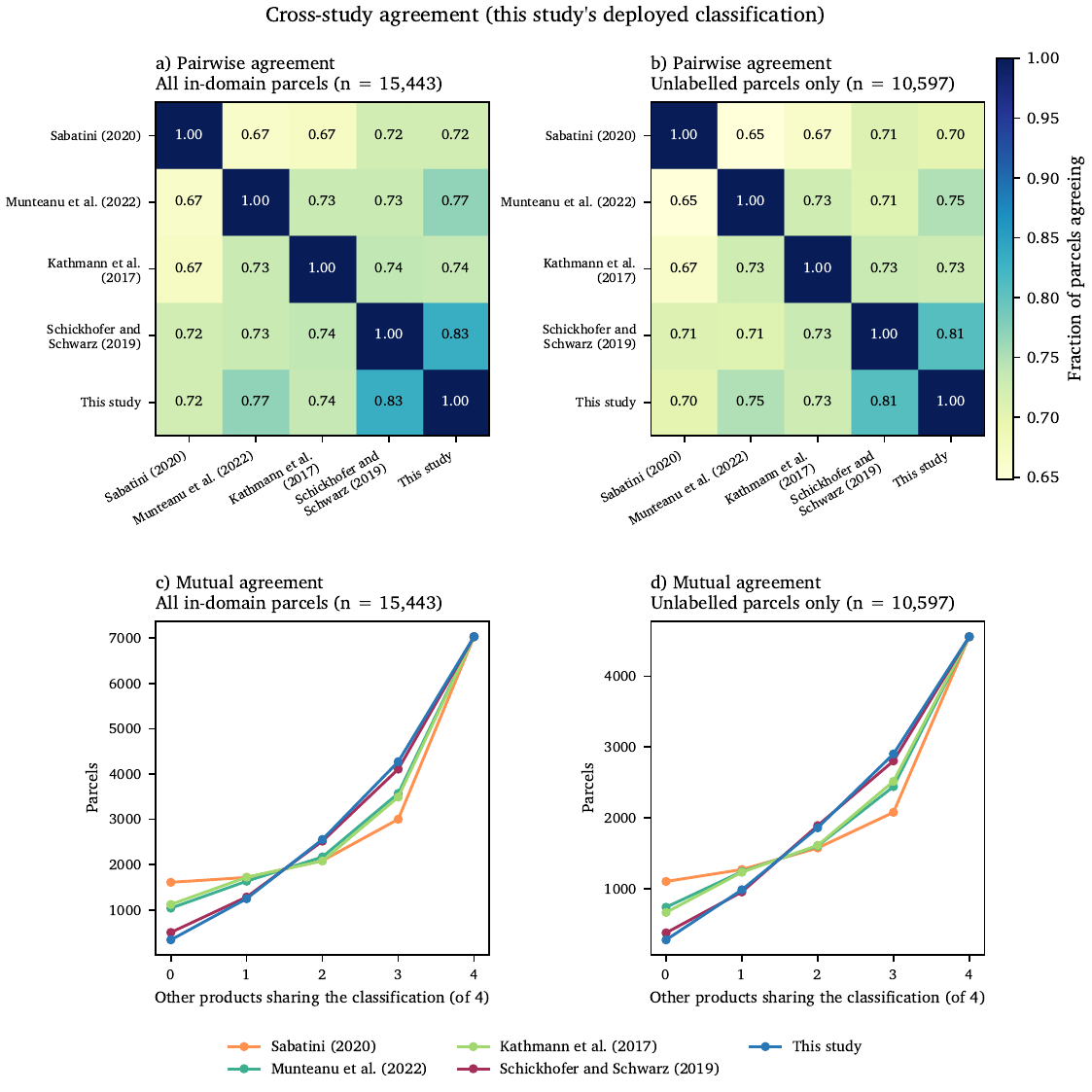}
    \caption{Cross-study parcel-level pairwise agreement. In-domain parcels are the 15,443 parcels with a model prediction (Section~\ref{sec:supp_ref_labels}).}
    \label{fig:cross_study_agreement}
\end{figure}

\end{document}